\documentclass[lettersize, journal]{IEEEtran}

\usepackage{amsmath, amsfonts}
\usepackage{algorithmic}
\usepackage{algorithm}
\usepackage{array}
\usepackage{color}
\usepackage[caption=false, font=normalsize, labelfont=sf, textfont=sf]{subfig}
\usepackage{textcomp}
\usepackage{stfloats}
\usepackage{url}
\usepackage{verbatim}
\usepackage{graphicx}
\usepackage{cite}
\usepackage{booktabs}
\usepackage{threeparttable}
\usepackage{multirow}
\usepackage[caption=false]{subfig}

\usepackage{tikz}
\usetikzlibrary{fadings, patterns}

\usepackage{pgfplots, pgfplotstable}
\usepgfplotslibrary{external, statistics, fillbetween, colormaps}

\pgfplotsset{
    glassybar/.style = {
        ybar,
        bar width = 12 pt,
        draw = #1!80!black,
        line width = 0.4 pt,
        fill = #1,
        fill opacity = 0.5,
        postaction = {
            fill = white,
            opacity = 0.5,
            path fading = north
        }
    }
}
\pgfplotsset{
    glassyarea/.style = {
        draw = #1!80!black,
        fill = #1,
        fill opacity = 0.05,
        postaction = {
            fill = white,
            opacity = 0.05,
            path fading = north
        }
    }
}
\pgfplotsset{
    hatchbar/.style 2 args = {
        ybar,
        bar width = 12pt,
        draw = #2!80!black,
        line width = 0.4pt,
        fill = none,
        pattern = #1,
        pattern color = #2!80!black
    }
}
\pgfplotsset{
    custom hatch legend/.style = {
        legend image code/.code = {
            \draw[black, fill = white] (0 pt, -4 pt) rectangle (16 pt, 4 pt);
            \fill[pattern = north east lines, pattern color = black] (0 pt, -4 pt) rectangle (16 pt, 4 pt);
        }
    }
}
\pgfplotsset{
    confusion matrix/.style={
        colormap={bluered}{
            color(0) = (blue!70); 
            color(1) = (cyan!70); 
            color(2) = (green!70);
            color(3) = (yellow!70); 
            color(4) = (red!70); 
            color(5) = (red!70!black!70)
        },
        enlargelimits = false,
        y dir = reverse,
        label style = {font = \scriptsize},
        xlabel style = {yshift = 12 pt},
        tick style = {draw = none},
        xticklabel style = {rotate = -30, anchor = north east, xshift = 2 pt, yshift = 10 pt, font = \scriptsize},
        xticklabel pos = upper,
        yticklabel style = {font = \scriptsize},
        nodes near coords,
        nodes near coords style = {
            font = \tiny,
            anchor = center,
            /pgf/number format/fixed,
            /pgf/number format/fixed zerofill,
            /pgf/number format/precision = 3,
            color = black
        }
    }
}

\definecolor{myblue}{rgb}{0.21, 0.49, 0.74}

\definecolor{OccRoad}{RGB}{128, 64, 128}
\definecolor{OccSidewalk}{RGB}{244, 35, 232}
\definecolor{OccBuilding}{RGB}{70, 70, 70}
\definecolor{OccWall}{RGB}{102, 102, 156}
\definecolor{OccFence}{RGB}{190, 153, 153}
\definecolor{OccPole}{RGB}{153, 153, 153}
\definecolor{OccTrafficLight}{RGB}{250, 170, 30}
\definecolor{OccTrafficSign}{RGB}{220, 220, 0}
\definecolor{OccVegetation}{RGB}{107, 142, 35}
\definecolor{OccTerrain}{RGB}{152, 251, 152}
\definecolor{OccSky}{RGB}{70, 130, 180}
\definecolor{OccPedestrian}{RGB}{220, 20, 60}
\definecolor{OccRider}{RGB}{255, 0, 0}
\definecolor{OccCar}{RGB}{0, 0, 142}
\definecolor{OccTruck}{RGB}{0, 0, 70}
\definecolor{OccBus}{RGB}{0, 60, 100}
\definecolor{OccTrain}{RGB}{0, 80, 100}
\definecolor{OccMotorcycle}{RGB}{0, 0, 230}
\definecolor{OccBicycle}{RGB}{119, 11, 32}
\definecolor{OccStatic}{RGB}{110, 190, 160}
\definecolor{OccDynamic}{RGB}{170, 120, 50}
\definecolor{OccOther}{RGB}{55, 90, 80}
\definecolor{OccWater}{RGB}{45, 60, 150}
\definecolor{OccRoadLine}{RGB}{227, 227, 227}
\definecolor{OccGround}{RGB}{81, 0, 81}
\definecolor{OccBridge}{RGB}{150, 100, 100}
\definecolor{OccRailTrack}{RGB}{230, 150, 140}
\definecolor{OccGuardRail}{RGB}{180, 165, 180}
\definecolor{OccRock}{RGB}{110, 110, 110}
\definecolor{OccTrafficCone}{RGB}{255, 165, 0}
\definecolor{OccBarrier}{RGB}{200, 128, 128}

\definecolor{Road}{RGB}{196, 80, 196}
\definecolor{Car}{RGB}{0, 128, 240}
\definecolor{Truck}{RGB}{128, 240, 64}
\definecolor{Bus}{RGB}{0, 144, 0}
\definecolor{Motorcycle}{RGB}{240, 240, 0}
\definecolor{Bicycle}{RGB}{0, 240, 240}
\definecolor{Rider}{RGB}{240, 144, 0}
\definecolor{Pedestrian}{RGB}{240, 0, 0}
\definecolor{TrafficLight}{RGB}{240, 160, 0}
\definecolor{TrafficSign}{RGB}{240, 0, 128}
\definecolor{TrafficCone}{RGB}{252, 180, 0}
\definecolor{Barrier}{RGB}{240, 128, 128}

\definecolor{MapRoad}{RGB}{196, 80, 196}
\definecolor{MapHazard}{RGB}{240, 80, 80}
\definecolor{MapRoadLine}{RGB}{160, 240, 40}
\definecolor{MapSidewalk}{RGB}{240, 196, 240}
\definecolor{MapCrosswalk}{RGB}{240, 196, 196}
\definecolor{MapTrafficCone}{RGB}{255, 165, 0}
\definecolor{MapBarrier}{RGB}{200, 128, 128}
\definecolor{MapCar}{RGB}{0, 128, 240}
\definecolor{MapTruck}{RGB}{80, 240, 80}
\definecolor{MapBus}{RGB}{0, 144, 0}
\definecolor{MapMotorcycle}{RGB}{240, 240, 0}
\definecolor{MapBicycle}{RGB}{0, 240, 240}
\definecolor{MapRider}{RGB}{240, 144, 0}
\definecolor{MapPedestrian}{RGB}{240, 0, 0}

\usepackage[pagebackref, breaklinks, colorlinks, allcolors=myblue]{hyperref}

\newcommand{\rb}[1]{\textbf{\color{red}#1}}
\newcommand{\gb}[1]{\textbf{\color{green}#1}}
\newcommand{\bb}[1]{\textbf{\color{blue}#1}}

\newcommand{\smallsim}{\smallsym{\mathrel}{\sim}}

\makeatletter
\newcommand{\smallsym}[2]{#1{\vcenter{\hbox{$\m@th\scriptscriptstyle#2$}}}}
\makeatother

\begin{document}

    \title{SimBEV2X: A Large-Scale Dataset and Data Generation Tool for Multi-Task Vehicle-to-Everything Cooperative Perception}

    \author{Goodarz Mehr, Sepideh Gohari, Montasir Abbas,~\IEEEmembership{Member,~IEEE}, Azim Eskandarian,~\IEEEmembership{Fellow,~IEEE}\thanks{Goodarz Mehr, Sepideh Gohari, and Azim Eskandarian are with the Autonomous Robots and Vehicles Laboratory (ARVL), College of Engineering, Virginia Commonwealth University (VCU), Richmond, Virginia, United States. E-mail: \{mehrg, goharis, eskandariana\}@vcu.edu (\textit{Corresponding author: Goodarz Mehr})\par Montasir Abbas is with the Charles E. Via, Jr. Department of Civil \& Environmental Engineering, Virginia Tech, Blacksburg, Virginia, United States. E-mail: abbas@vt.edu}}

    \markboth{IEEE Transactions on IEEE Transactions}
    {Mehr \MakeLowercase{\textit{et al.}}: SimBEV2X: A Large-Scale Dataset and Data Generation Tool for Multi-Task Vehicle-to-Everything Cooperative Perception}


    \maketitle

    \begin{abstract}
    Cooperative perception through vehicle-to-everything (V2X) communication can overcome the inherent physical limitations of individual autonomous vehicles, such as occlusions and limited sensor range. However, the development of robust V2X algorithms, particularly those relying on unified spatial representations like bird's-eye view (BEV) representation, is hampered by the lack of large-scale, multi-modal, multi-task datasets. Moreover, collecting and annotating a large set of synchronized, real-world multi-agent data is prohibitively expensive. This has resulted in a landscape where existing V2X datasets are notably limited in both size and scope. To overcome this, we introduce SimBEV2X, an advanced synthetic data generation tool built on the CARLA simulator. SimBEV2X automatically creates highly randomized driving scenarios to collect rich multi-modal sensor data alongside various types of ground truth including 3D object bounding boxes with unique track IDs, HD map information, BEV segmentation maps, and semantic occupancy voxel grids from both vehicles and road-side units (RSUs). We also present the SimBEV2X dataset, the largest V2X perception dataset to date. The dataset comprises 258 scenes, each involving up to 8 connected vehicles and up to 4 RSUs across a variety of road networks. The SimBEV2X dataset is an order of magnitude larger than existing V2X datasets and contains 102,200 frames, 588,520 lidar point clouds, more than 3 million images, over 27 million object bounding boxes, and a comprehensive set of other data annotations. Finally, we establish a strong baseline on the SimBEV2X dataset using CoopDet3D and propose CoBEVFusion, a novel architecture that combines CoopDet3D with fused axial attention (FAX) for context-aware multi-agent feature aggregation, resulting in superior performance. SimBEV2X, the SimBEV2X dataset, and CoBEVFusion are available at \href{https://simbev2x.org}{simbev2x.org} and \href{https://github.com/GoodarzMehr/SimBEV2X}{github.com/GoodarzMehr/SimBEV2X}. 
    \end{abstract}

    \begin{IEEEkeywords}
    Autonomous driving, bird's-eye view (BEV), cooperative perception, synthetic dataset, vehicle-to-everything (V2X), 3D semantic occupancy
    \vspace*{-4 pt}
    \end{IEEEkeywords}

    \section{Introduction} \label{sec:introduction}
    
    \IEEEPARstart{A}{s} the development of autonomous vehicle (AV) technology accelerates, the inherent limitations of single-agent perception systems become increasingly apparent. While individual vehicles equipped with multimodal sensors can achieve a robust understanding of their immediate surroundings \cite{zhao2023autonomous, wang2023multimodal, eskandarian2019research}, they remain fundamentally constrained by physical barriers such as occlusions, limited sensor range, and adverse weather conditions \cite{cui2024survey, bijelic2020seeing}. To mitigate these issues and ensure safe decision-making in complex environments, cooperative perception via vehicle-to-everything (V2X) communication has emerged as a promising paradigm \cite{ghorai2022state}. By enabling the exchange of sensory data and extracted features between multiple connected vehicles and road-side units (RSUs), V2X systems effectively expand the receptive field of each agent, drastically reducing uncertainty in dynamic driving scenarios.

    \begin{figure}[!t]
        \centering
        \includegraphics[width=0.64\columnwidth]{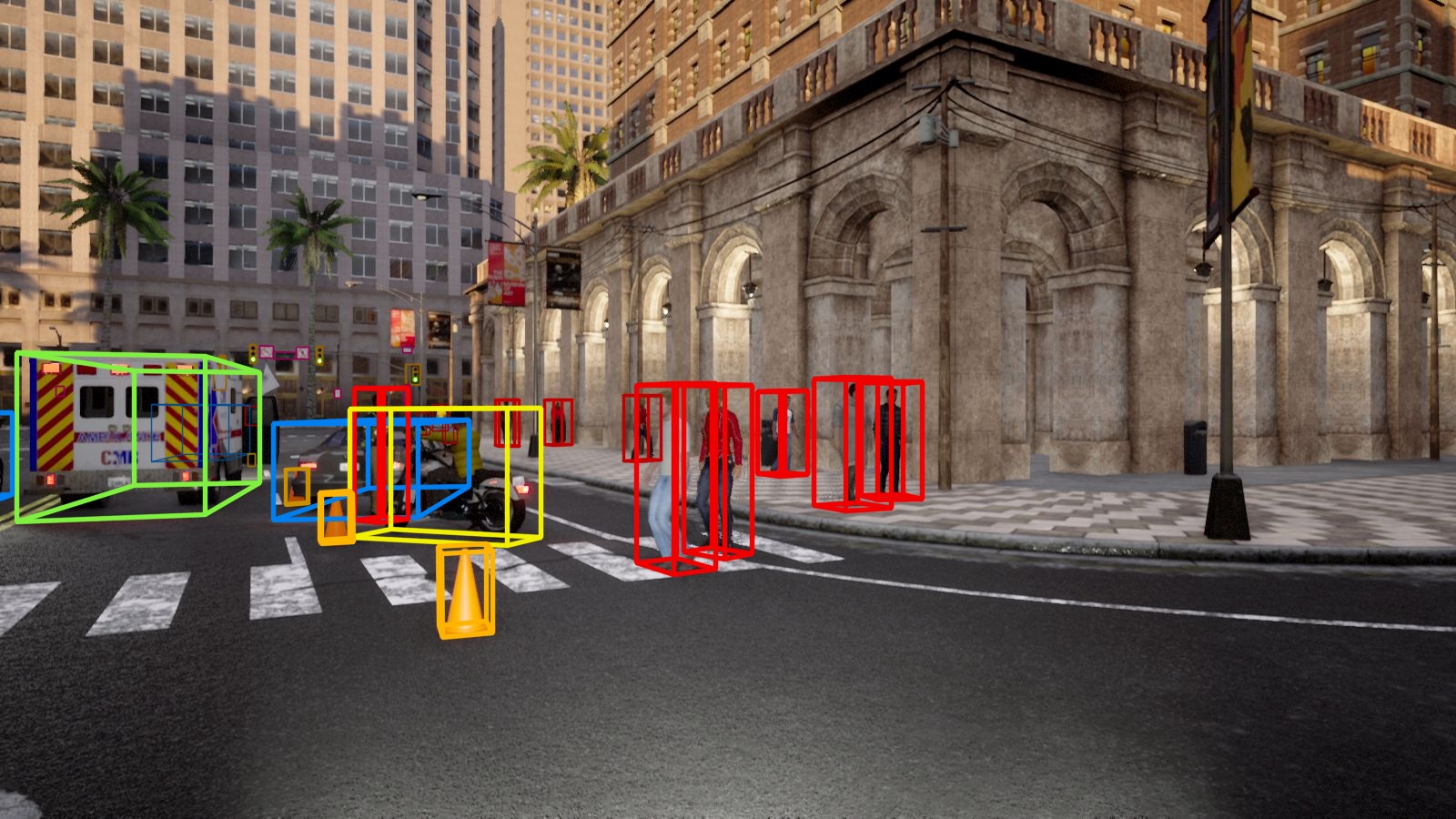}\hspace{-3.1pt}
        \includegraphics[width=0.36\columnwidth]{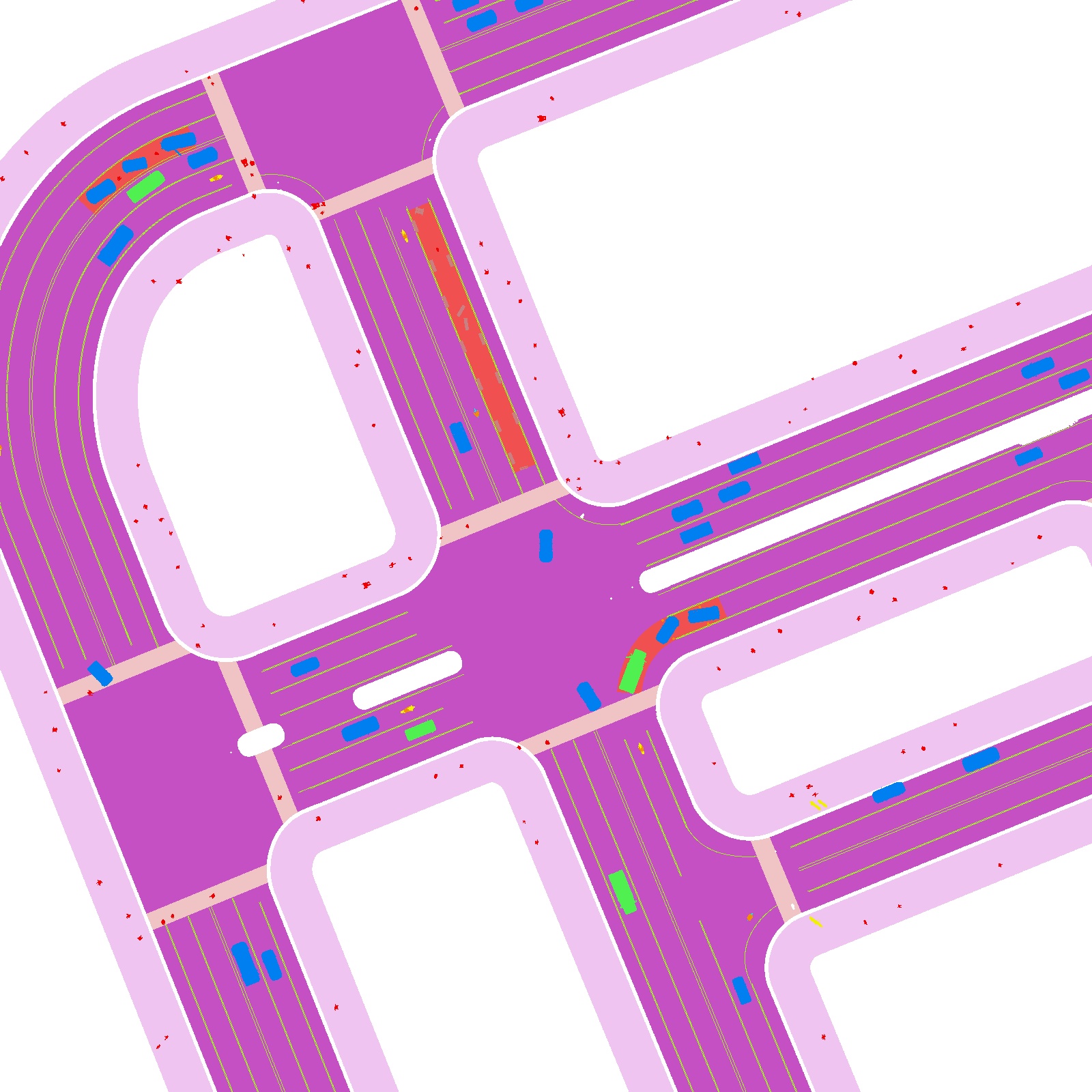}
        \includegraphics[width=\columnwidth]{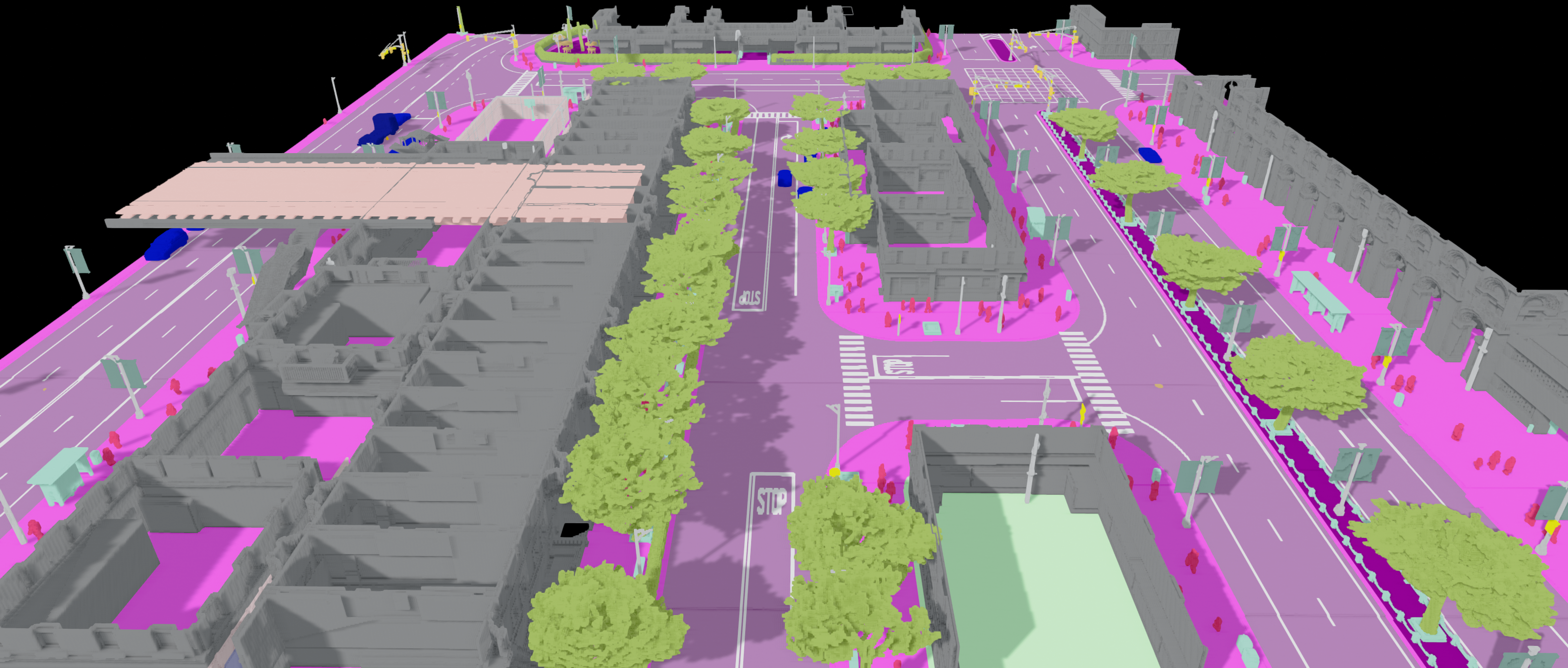}
        \setlength{\abovecaptionskip}{-12 pt}
        \caption{Three annotation types provided by SimBEV2X: 3D object bounding boxes (top left), BEV segmentation maps (top right), and 3D semantic occupancy voxel grids (bottom).} \label{fig:simbev2x-gt}
        \vspace*{-12 pt}
    \end{figure}

    A foundational requirement for effective V2X perception is a unified spatial representation that can seamlessly integrate information from diverse agents with varying perspectives and hardware configurations. Bird's-eye view (BEV) and 3D semantic occupancy representations have proven to be highly effective for this purpose \cite{ma2024vision}. BEV representation provides a concise and geometrically accurate overhead perspective that simplifies the alignment and fusion of spatial features from multiple agents. 3D semantic occupancy representation goes a step further by capturing the fine-grained, voxelized 3D structure of the environment, which is vital for downstream planning and control tasks. Consequently, algorithms that rely on these representations are rapidly becoming the standard for next-generation cooperative perception systems \cite{xu2023cobevt, ma2024macp, song2024collaborative, he2025mdnet}.

    Despite growing interest in advanced V2X perception models, their development has been bottlenecked by the lack of comprehensive large-scale datasets. Collecting real-world V2X data is much more challenging and resource-intensive than single-agent data collection. It requires precise spatial and temporal synchronization across multiple moving vehicles and stationary infrastructure elements, complicated by the extensive labor required for data annotation \cite{uricar2019challenges, liu2024survey}. Because of this, the existing V2X datasets are small and, as \cite{caesar2020nuscenes} suggests, may not be enough to evaluate the full potential of complex algorithms. Moreover, existing V2X datasets often lack environmental diversity (for example, collected at a single intersection) or rely on limited scenarios that fail to encapsulate the stochastic variability of real-world traffic patterns and environmental conditions. Finally, very few V2X datasets provide high-quality annotations for multiple tasks such as 3D object detection, BEV segmentation, and 3D semantic occupancy prediction.

    To overcome these barriers, we propose SimBEV2X, an advanced synthetic data generation tool built on the CARLA simulator \cite{dosovitskiy2017carla}. Based on our single-agent data generation tool SimBEV \cite{mehr2025simbev}, SimBEV2X is specifically engineered to generate rich, cooperative perception data by orchestrating complex, user-configurable multi-agent interactions. Rather than relying on rigid, pre-defined scenarios, SimBEV2X dynamically creates highly randomized environments, systematically varying weather and lighting conditions, background traffic density and behavior, road hazards, and initial agent positions to ensure a long-tail distribution of scenarios. For every generated scenario, SimBEV2X automatically collects rigorous annotations per frame, including HD map information, 3D object bounding boxes with unique track IDs, BEV segmentation maps, and semantic occupancy voxel grids for all connected agents, as shown in \autoref{fig:simbev2x-gt}.

    To demonstrate the scalability and utility of SimBEV2X, we use it to create the SimBEV2X dataset, the largest V2X perception dataset and benchmark to date, designed to push the boundaries of cooperative perception. The dataset comprises 258 distinct scenes, 102,200 frames, 588,520 lidar point clouds, more than 3 million images, and over 27 million object bounding boxes. It features scenes with complex road topologies populated by a variable set of up to 8 connected vehicles and up to 4 RSUs. By providing multi-modal multi-agent data alongside rich ground truths, the SimBEV2X dataset serves as a rigorous testing ground for state-of-the-art V2X algorithms.

    \renewcommand{\thefootnote}{\fnsymbol{footnote}}
    
    \begin{table*}[!t]
        \setlength{\abovecaptionskip}{0 pt}
        \caption{Cooperative Perception Datasets} \label{table:datasets}
        \centering
        \begin{threeparttable}
            \begin{tabular}{l | c c c c c c c c}
            \toprule
            \multirow{2}{*}{Dataset} & OPV2V & V2X-Sim & DAIR-V2X-C & V2V4Real & V2X-Seq (SPD) & TUMTraf-V2X & V2X-Real & \textbf{SimBEV2X} \\
             & \cite{xu2022opv2v} & \cite{li2022v2x} & \cite{yu2022dair} & \cite{xu2023v2v4real} & \cite{yu2023v2x} & \cite{zimmer2024tumtraf} & \cite{xiang2024v2x} & \textbf{(Ours)} \\
            \midrule
            Year & 2022 & 2022 & 2022 & 2023 & 2023 & 2024 & 2024 & \textbf{2026} \\
            Type & V2V & V2X & V2I & V2V & V2I & V2I & V2X & \textbf{V2X} \\
            Real/Sim & Sim & Sim & Real & Real & Real & Real & Real & \textbf{Sim} \\
            Location & CARLA & CARLA & China & USA & China & Germany & USA & \textbf{CARLA} \\
            Annotation range & 120 m & 70 m & 280 m & 200 m & 280 m & 200 m & 200 m & \textbf{200 m} \\
            Weather variation & No & No & Yes & No & Yes & Yes & Yes & \textbf{Yes\tnote{$\dagger$}} \\
            \# Object classes & 1 & 1 & 10 & 5 & 9 & 8 & 10 & \textbf{10} \\
            \# BEV classes & 3\tnote{$\ast$} & 7 & N/A & N/A & N/A & N/A & N/A & \textbf{14} \\
            \# 3D classes & 22\tnote{$\star$} & N/A & N/A & N/A & N/A & N/A & N/A & \textbf{31\tnote{$\ddagger$}} \\
            \# Frames & 11,464 & 10,000 & 39,000 & 10,000 & 15,000 & 1,000 & - & \textbf{102,200} \\
            \# Point clouds & 40,000 & 47,200 & 78,000 & 20,000 & 15,000 & 3,000 & 33,000 & \textbf{588,520} \\
            \# Images & 160,000 & - & 312,000 & 40,000 & 120,000 & 5,000 & 171,000 & \textbf{3,241,560} \\
            \# Bounding boxes & 232,913 & 26,600 & 464,000 & 240,000 & - & 29,380 & 1,200,000 & \textbf{27,337,066} \\
            \bottomrule
            \end{tabular}
            \begin{tablenotes}
            \item[$\ast$]{Introduced in \cite{xu2023cobevt}.\hspace*{60 pt}
            $^{\star}$ Introduced in \cite{wu2025synthetic}}.\hspace*{60 pt}
            $^{\dagger}${Dynamic weather variation (i.e. continuous weather shift \cite{sun2022shift}).}
            \item[$\ddagger$]{Labels for $sky$, $train$, and $rock$ are supported by SimBEV2X but not present in the dataset.}
            \end{tablenotes}
        \end{threeparttable}
        \vspace{-12 pt}
    \end{table*}

    Finally, to establish a strong baseline and highlight the dataset's potential, we benchmark it using CoopDet3D \cite{zimmer2024tumtraf}. We also introduce and benchmark CoBEVFusion, a novel cooperative perception model that adds fused axial attention (FAX) \cite{xu2023cobevt} to the max-pooling cooperative feature fuser of CoopDet3D to perform context-aware multi-agent feature aggregation and surpass CoopDet3D's performance.

    In summary, our main contributions are:

    \begin{itemize}
        \item We introduce SimBEV2X, a highly scalable and user-configurable V2X data generation tool built on the CARLA simulator. SimBEV2X uses domain randomization to generate diverse cooperative driving scenarios.

        \item We present the SimBEV2X dataset, the largest V2X dataset to date that comprises 258 scenes, each involving up to 8 connected vehicles and up to 4 RSUs. The dataset contains multi-modal sensor data and multiple forms of per-frame annotation that can benefit a wide variety of cooperative perception applications.

        \item We introduce a novel cooperative perception model, CoBEVFusion, and use it alongside CoopDet3D to benchmark the dataset and establish a baseline.
    \end{itemize}

    \section{Related Work} \label{sec:related-work}

    Autonomous driving datasets can be categorized based on the number of agents (vehicles and RSUs) involved. \autoref{table:datasets} highlights the main differences between the SimBEV2X dataset and the existing vehicle-to-vehicle (V2V), vehicle-to-infrastructure (V2I), and V2X datasets.

    \subsection{Single-Agent Datasets} \label{subsec:single-agent-datasets}

    Single-agent datasets are obtained from the viewpoint of a single entity, often an autonomous vehicle. Real-world datasets often target specific subsets of perception tasks, as the high cost of collecting and annotating data limits their scope. The KITTI dataset is one of the oldest and most prominent driving datasets and supports depth estimation and 2D/3D object detection and tracking \cite{geiger2013vision}. Other notable image-based datasets include Cityscapes \cite{cordts2016cityscapes} and Mapillary \cite{neuhold2017mapillary}, which focus on segmentation, while A*3D \cite{pham20203d} is geared towards 3D object detection. More recently, large-scale datasets such as BDD100K \cite{yu2020bdd100k}, Waymo Open \cite{sun2020scalability}, ApolloScape \cite{huang2018apolloscape}, Argoverse 2 \cite{wilson2023argoverse}, OmniHD-Scenes \cite{zheng2026omnihd}, and nuScenes \cite{caesar2020nuscenes} have emerged, offering multi-modal driving data and multi-task annotations but primarily focusing on object detection and tracking.

    Synthetic driving datasets are created using graphics engines and physics simulators. While such data may lack exact real-world fidelity, they require much less effort to generate and accurately annotate at scale. For example, SYNTHIA's dedicated simulator generates RGB and semantically segmented images \cite{ros2016synthia}. Video games have also served as a source of data. For instance, the GTA-V dataset consists of RGB and semantically segmented images extracted from GTA \cite{richter2016playing}, while ViPER expands on it by including optical flow images and discrete environmental labels \cite{richter2017playing}. Since 2016, CARLA's introduction has fostered the systemic generation of driving datasets \cite{dosovitskiy2017carla}. Examples include All-in-One Drive \cite{weng2023all} and SHIFT \cite{sun2022shift} -- a large-scale dataset for autonomous driving designed to simulate discrete and continuous changes in weather and traffic conditions to evaluate domain adaptation algorithms.

    \subsection{Multi-Agent Datasets} \label{subsec:multi-agent-datasets}

    Multi-agent datasets contain data from multiple viewpoints that can be used to improve each agent's understanding of its surrounding environment \cite{wang2025collaborative}. OPV2V \cite{xu2022opv2v} introduced the first V2V cooperative perception dataset and was created using CARLA and OpenCDA \cite{xu2021opencda}. Although initially limited to 3D object detection for a single class (car), the log files provided with the dataset have helped other researchers replay the existing scenes and extend OPV2V with annotations for BEV segmentation \cite{xu2023cobevt} and 3D semantic occupancy prediction \cite{wu2025synthetic}. V2X-Sim \cite{li2022v2x} further extends the use of CARLA by creating a V2X dataset that provides BEV annotations alongside 3D bounding boxes for cars.

    A number of real-world multi-agent datasets have also been introduced in the past few years. DAIR-V2X-C is the first such dataset for V2I cooperative perception \cite{yu2022dair}. Select sequences of this dataset are extended with track IDs and divided into a sequential perception dataset (SPD) and a trajectory forecasting dataset to form the V2X-Seq dataset \cite{yu2023v2x}. TUMTraf-V2X is another high-quality V2I dataset with 3D bounding boxes and unique track IDs for multiple object classes \cite{zimmer2024tumtraf}. In V2V4Real, two vehicles equipped with various sensors are used to collect data from a variety of scenarios, creating a V2V dataset \cite{xu2023v2v4real}. This approach is further extended through the use of two RSUs at an intersection to collect data for V2X-Real, the first real-world V2X dataset \cite{xiang2024v2x}.

    Collecting cooperative perception datasets, particularly in real-world settings, is both time- and resource-intensive. Consequently, the existing datasets face two main limitations. First, their relatively small scale can adversely affect model performance, as demonstrated by \cite{caesar2020nuscenes}. Second, most of them primarily target 3D object detection, limiting their applicability to a broader range of tasks such as cooperative 3D semantic occupancy prediction. These challenges motivated us to leverage the CARLA simulator and create the SimBEV2X dataset, which is similar in size to nuScenes \cite{caesar2020nuscenes}, offers a variety of annotations, and provides support for numerous sensor modalities, data from some of which can also serve as the ground truth (e.g., depth images).

    \vspace*{-2 pt}
    \subsection{Cooperative Perception Algorithms} \label{subsec:cooperative-perception-algorithms}

    Cooperative perception algorithms can be categorized as \textit{early fusion}, \textit{intermediate fusion}, and \textit{late fusion}, depending on the stage where cooperative data fusion takes place.
    
    \textit{Early fusion}, also known as raw fusion, was predominantly used during the early years of cooperative perception research. Studies using this approach primarily involve scenario-based experiments where lidar point clouds or camera images are merged to enhance the performance of perception algorithms \cite{kim2014multivehicle, sridhar2019cooperative, xiong2021toward, chen2019cooper, zhang2024incentivizing}. For lidar data, a common method is to consolidate point clouds obtained from multiple agents. Stitching camera images from different agents together to allow visibility past occlusions and utilizing images from multiple perspectives to detect and localize objects based on their shared visual features are also common approaches \cite{sridhar2019cooperative}. However, \textit{early fusion} has significant drawbacks. In particular, it requires the transmission of large data volumes that exceed the bandwidth of current V2X communication systems \cite{ma2024macp}.

    More recently, the focus has shifted towards \textit{intermediate fusion}, also known as feature fusion. In this approach, agents extract feature maps from their perception data and transmit them to other agents. Each agent then fuses these feature maps to enhance its perception \cite{xu2023cobevt, chen2019f, yang2021machine, li2021learning, hu2023collaboration, song2023fedbevt, xiang2023hm, xu2022v2x, luo2022complementarity, cui2022coopernaut, tian2022federated, ma2024macp, wang2020v2vnet, liu2023cooperative, xiang2023v2xp, qu2023sicp, xu2022opv2v, shi2025v2v, xin2025pnpda+, wang2025adaptive, cheng2025transformer, tang2025rocooper, liu2025coop, zha2025heterogeneous, song2025traf, li2025v2x, yuan2025sparsealign, xiang2025v2x, wang2025coopdetr, jin2025bandwidth, feng2025lcv2i, liu2025mmcooper, he2025mdnet, wang2025lightweight, meng2024agentalign}. While \textit{intermediate fusion} can address the V2X communication bandwidth challenges of \textit{early fusion} and has better performance compared to both \textit{early} and \textit{late fusion}, the downside is that it assumes that all participating agents employ an identical feature extraction backbone; otherwise, the extracted features belong to different statistical domains, leading to a large drop in performance \cite{xu2022bridging}. This can make \textit{intermediate fusion} impractical for heterogeneous deployments. Only \cite{xu2022bridging} addresses this issue, though the solution, training a separate domain adaptation model for every combination of perception models, is not very practical or scalable.

    The third and final approach to cooperative perception is \textit{late fusion}. In \textit{late fusion}, predictions of perception models - often object bounding boxes - are communicated between agents and fused \cite{bodla2017soft, xu2023model, zhou2022aicp, rockl2008v2v, xu2018pointfusion, zhang2021distributed, justolff, xiang2025mtrcp}. \textit{Late fusion} methods do not require the transmission of large volumes of data or rely on a specific perception model or sensing modality.

    \vspace*{-2 pt}
    \section{SimBEV2X} \label{sec:simbev2x}
    
    SimBEV2X uses a custom version of CARLA 0.9.16 to simulate the environment, perception sensors, and traffic behavior. Initialized by a configuration file, SimBEV2X automatically manages the entire data collection process by manipulating simulation elements through CARLA's Python API to produce the user's desired outcome. It simplifies user interactions with CARLA, equipping the user with CARLA's customizability when necessary and taking control when not. This flexibility and ease of use allows researchers to quickly create custom datasets that suit their needs.

    \begin{figure*}[!t]
        \centering
        \includegraphics[width=\textwidth]{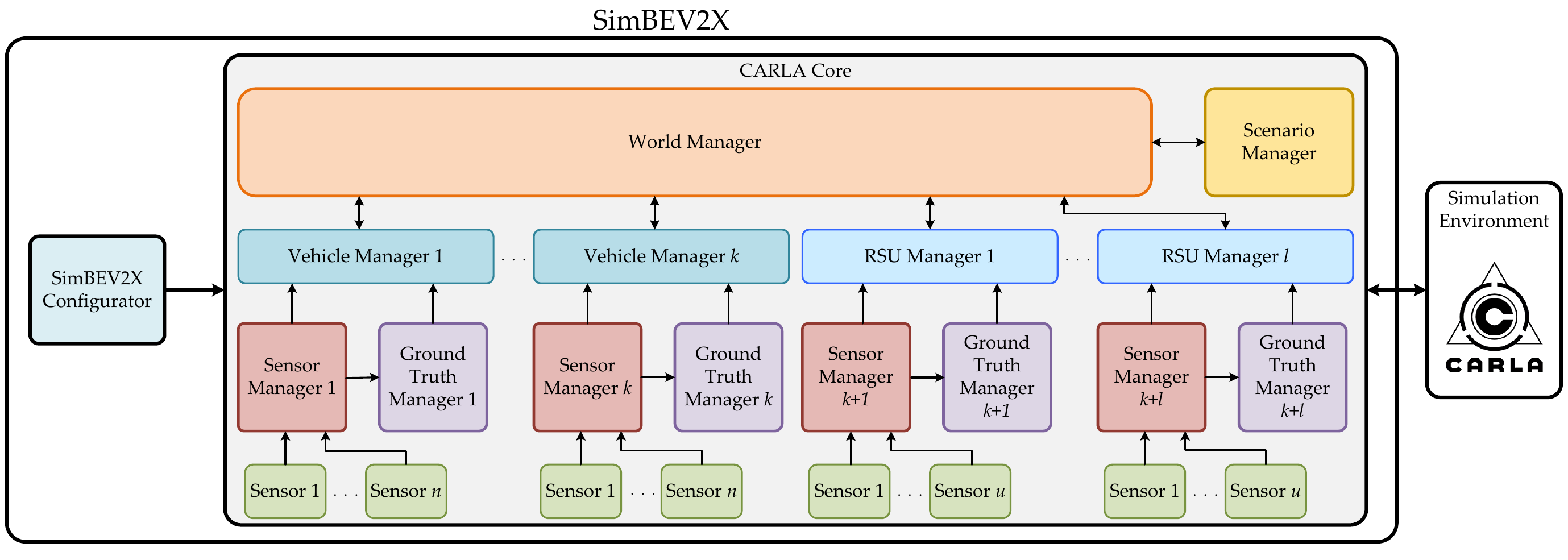}
        \setlength{\abovecaptionskip}{-20 pt}
        \caption{Overview of SimBEV2X's modular design. The \textit{Carla Core} module launches the CARLA server, creates a client, and then passes that client to the \textit{World Manager} module. \textit{World Manager} is responsible for creating scenarios, collecting data, and overseeing the other modules. The \textit{Scenario Manager} module is responsible for setting up each scenario by configuring the weather, lights, road hazards, and background traffic. Each data collection vehicle/RSU is managed by a \textit{Vehicle/RSU Manager} that controls its behavior and properties. Each \textit{Vehicle/RSU Manager} also controls a \textit{Sensor Manager}, which collects and organizes data from the sensors attached to the respective vehicle/RSU, and a \textit{Ground Truth Manager} that collects 3D object bounding boxes and calculates the BEV and HD map ground truths for that vehicle/RSU.}
        \label{fig:simbev2x-modules}
        \vspace*{-16 pt}
    \end{figure*}

    We made many changes to CARLA 0.9.16 to make it suitable for our simulations. Some changes were bug fixes that addressed core issues with the accuracy of data obtained from CARLA, and many were contributed to the main CARLA repository as well. Others were enhancements to different CARLA modules, such as the Traffic Manager. We also heavily customized the CARLA content library to improve the visual accuracy and realism of the simulations. More information about these changes is provided in Appendix A.I.
    
    \subsection{Design} \label{subsec:simbev2x-overview}
    
    SimBEV2X creates a statistically diverse set of scenarios by randomizing (within realistic bounds) as many aspects of the simulation as possible. To create a dataset, SimBEV2X generates and collects data from consecutive episodes, or scenes, each with a unique configuration. SimBEV2X can also replace individual scenes or replay them from the log files to collect additional data.

    SimBEV2X has a modular design, as shown in \autoref{fig:simbev2x-modules}. The \textit{Carla Core} module launches the CARLA server, creates a client, and then passes that client to the \textit{World Manager} module. \textit{World Manager} is SimBEV2X's primary module, responsible for creating scenarios, collecting data, and overseeing the \textit{Scenario Manager}, \textit{Vehicle Manager}, and \textit{RSU Manager} modules. The \textit{Scenario Manager} module is responsible for setting up each scenario by configuring the weather, lights, road hazards, and background traffic (vehicles and pedestrians). Each data collection vehicle/RSU is managed by a \textit{Vehicle/RSU Manager} that controls its behavior and properties. Each \textit{Vehicle/RSU Manager} also controls a \textit{Sensor Manager}, which collects and organizes data from the sensors attached to the respective vehicle/RSU, and a \textit{Ground Truth Manager} that collects 3D object bounding boxes and calculates the BEV and HD map ground truths for that vehicle/RSU.

\begin{figure*}[!t]
        \centering
        \includegraphics[width=\textwidth]{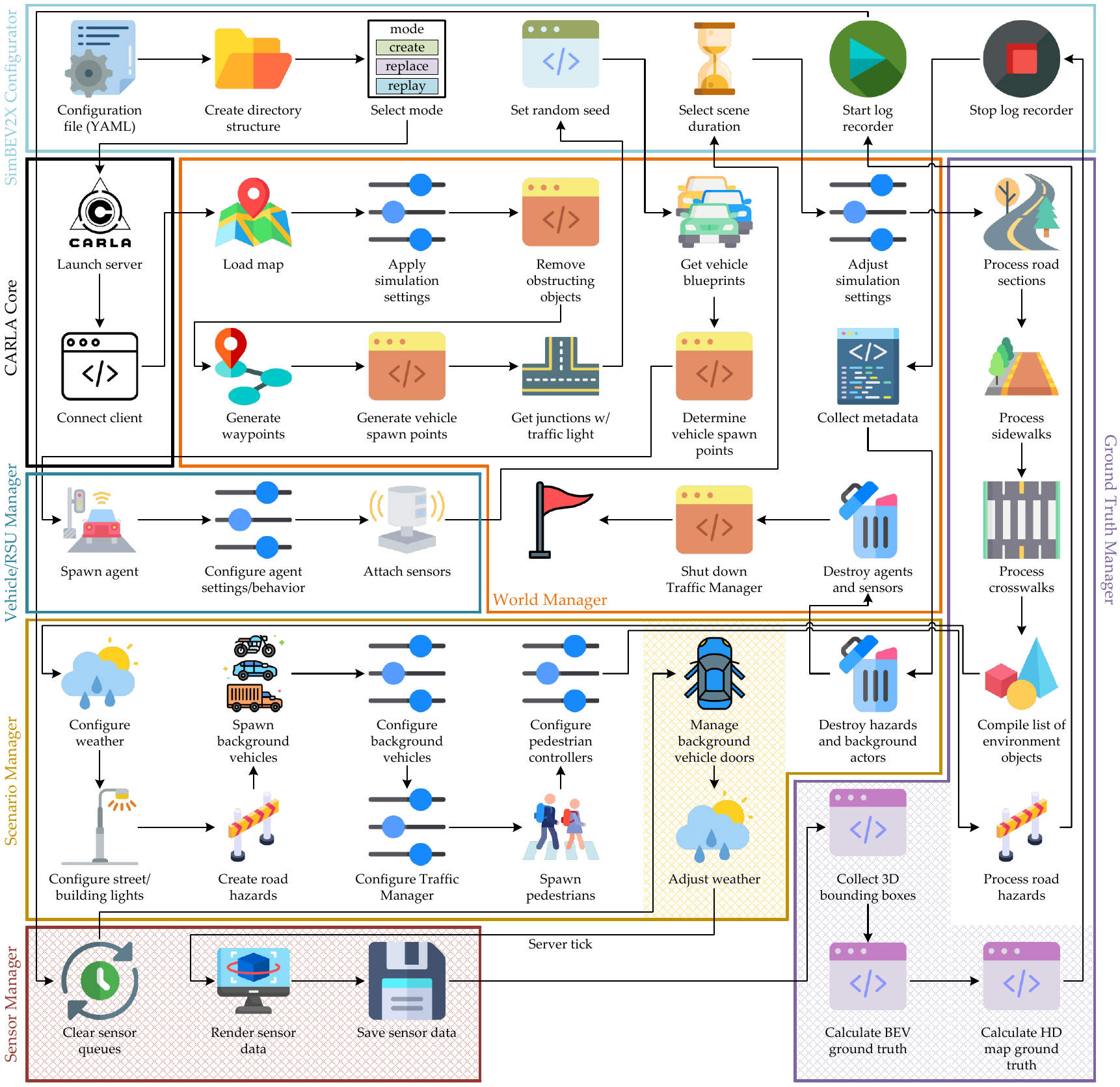}
        \setlength{\abovecaptionskip}{-16 pt}
        \caption{Overview of SimBEV2X's logic. Steps for Vehicle/RSU Manager, Sensor Manager, and Ground Truth Manager are repeated for each agent. Shaded steps are repeated for each time step until the end of the scene. Icons: Flaticon.com.}
        \label{fig:simbev2x-logic}
        \vspace*{-2 pt}
    \end{figure*}

    SimBEV2X's logic flow is shown in \autoref{fig:simbev2x-logic}. To create a new dataset, the user configures the desired number of scenes for each map (which can be an existing CARLA map or a custom one) for the training, validation, and test sets. If a SimBEV2X dataset already exists, for each map, SimBEV2X subtracts the number of existing scenes in that dataset from the number of desired scenes to determine how many additional scenes it needs to create. This feature allows the user to expand an existing dataset created by SimBEV2X or continue the data collection process in the event of a crash.

    At the start of each scene, SimBEV2X generates a set of potential spawn points that are a user-specified distance apart from each other and are uniformly distributed across the map's roads. Then it randomly selects a number of neighboring spawn points and spawns the data collection vehicles and their attached sensors, as well as a number of RSUs at nearby signalized intersections (if any), although the user has the option to specify a set of coordinates that SimBEV2X must use as spawn points for data collection vehicles instead. The number of vehicles and RSUs for each scene is randomly chosen within the user-specified limits. SimBEV2X also randomly chooses the scene's duration within the user-specified bounds.

    In the next step, SimBEV2X configures the weather and street/building lights and creates a random number of hazard areas (traffic accidents and road work areas). Then, it creates the background traffic. The user can specify the number of background vehicles and pedestrians, or allow SimBEV2X to randomly select each. CARLA's Traffic Manager controls the behavior of all vehicles and pedestrians throughout the scene.

    Because all vehicles and pedestrians start from rest, SimBEV2X runs the simulation for a user-specified warm-up period before collecting data to start from a more realistic state. At each time step during the data collection period, it collects data from the desired sensors, gathers 3D object bounding boxes, and calculates the BEV, HD map, and 3D semantic occupancy ground truths. Subsequently, it processes meta information about the collected data and a log of the scene, destroys all vehicles, RSUs, pedestrians, and sensors, and moves on to the next scene.
    
    \subsection{Sensors} \label{subsec:simbev-sensors}
    
\begin{figure}[!t]
        \centering
        \includegraphics[width=\columnwidth]{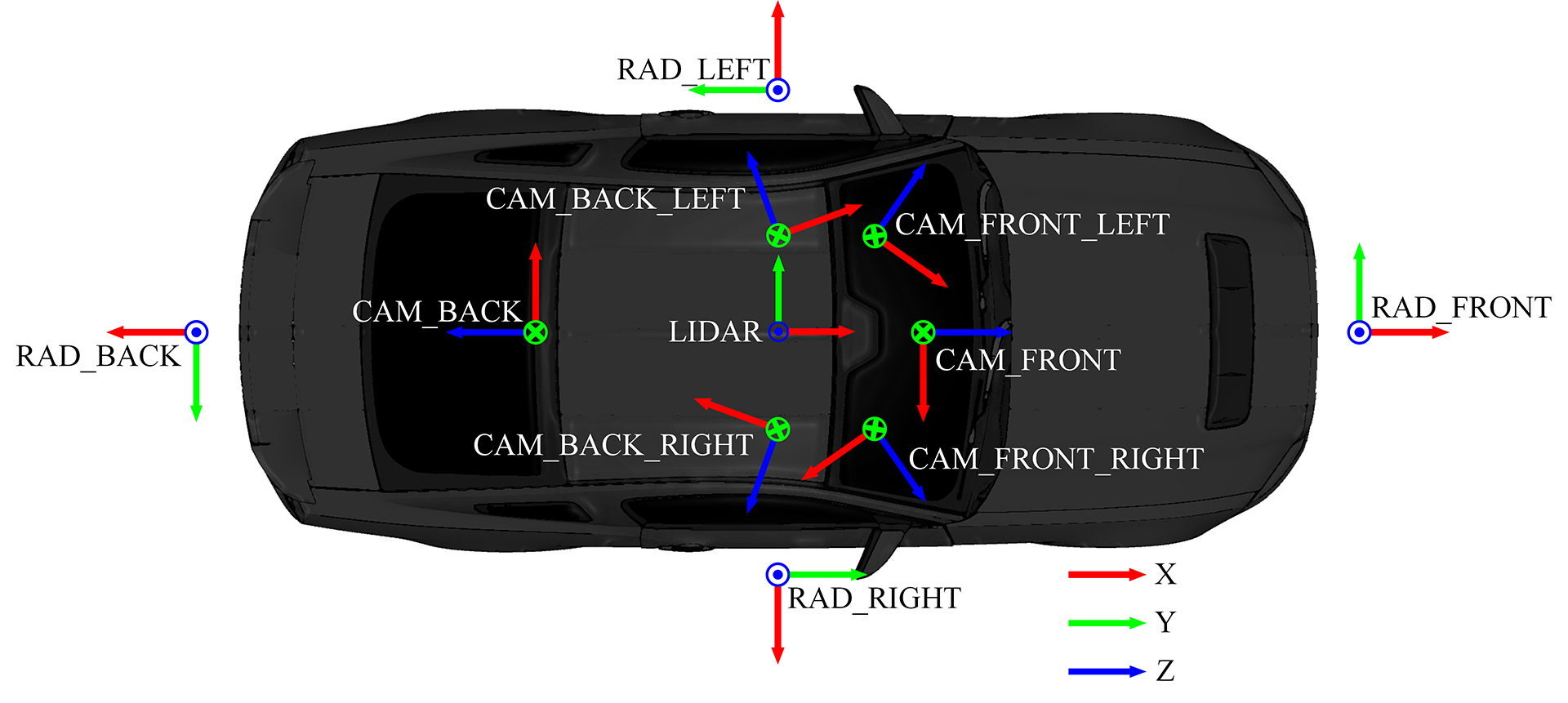}
        \setlength{\abovecaptionskip}{-8 pt}
        \caption{Coordinate frames of the perception sensors attached to data collection vehicles in SimBEV2X.}
        \label{fig:vehicle-sensors}
        \vspace*{-4 pt}
    \end{figure}

    SimBEV2X supports a variety of CARLA sensors, including five camera types (RGB, semantic segmentation, instance segmentation, depth, and optical flow), two lidar types (regular and semantic), radar, GNSS, and IMU. The user has full control over each sensor's parameters (e.g. a camera's resolution or field of view (FoV)), but the placement of the sensors is fixed, as shown in \autoref{fig:vehicle-sensors}, \autoref{table:vehicle-sensors}, and \autoref{table:rsu-sensors}. As in \cite{caesar2020nuscenes}, six cameras are placed above each data collection vehicle to offer a 360-degree view of the surroundings, while a radar is placed on each of the four sides of the vehicle. The GNSS and IMU are located at the center of the vehicle (the origin of the vehicle's coordinate system), and the lidar is placed atop the vehicle above that center. For each RSU, which is represented as a pole in the simulation, a radar is mounted at the bottom and a lidar at the top. Three cameras are placed in the middle, with the middle camera facing the center of the intersection and the other two cameras angled to the sides.

    \subsection{Scene Configuration} \label{subsec:simbev-config}
    
    In the following sections, we discuss how different aspects of the simulation are randomized for each scene to ensure that the generated scenes are as unique and diverse as possible.

    \paragraph{Spawning agents} \label{par:spawn-agents}

    SimBEV2X offers two different methods, namely \textit{cluster selection} and \textit{chain selection}, for selecting the spawn points of data collection vehicles from the set of available spawn points. The user can use either method or both. For both methods, the user has to specify a distance, denoted by $d_{cl}$ and $d_{ch}$, respectively, and the first spawn point is selected randomly. In \textit{cluster selection}, the remaining spawn points are selected randomly from the set of spawn points that fall within a radius $r$ of the first spawn point, where
    \begin{equation} \label{eq:clustered-spawn}
    r = \frac{d_{cl}}{\sin(\frac{\pi}{N})}
    \end{equation}
    and $N$ is the number of data collection vehicles. Intuitively, $r$ is the largest radius for a circle with $N$ points on its perimeter where it is always possible to find at least two points whose distance is at most $2d_{cl}$. In \textit{chain selection}, each remaining spawn point is selected from the set of available spawn points that are within a $d_{ch}$ radius of at least one of the previously selected spawn points.

    To spawn RSUs, SimBEV2X looks at the signalized intersections close to the data collection vehicle spawn points. For \textit{cluster selection}, these are intersections that fall within a radius $r$ of the first spawn point. For \textit{chain selection}, these are intersections that are within $d_{ch}$ of at least one previously selected spawn point. SimBEV2X aggregates the location of traffic lights at such intersections and then randomly selects RSU spawn points from this set.
    
    \paragraph{Weather} \label{par:weather}

    Weather in CARLA is controlled through a set of parameters such as cloudiness, sun altitude angle, fog density, etc. By default, SimBEV2X randomly chooses these parameters for each scene (subject to some constraints to ensure that the weather is realistic) to create a diverse set of weather conditions. The user also has the option of setting any subset of these parameters to fixed values. For instance, setting the sun altitude angle to anything less than zero creates night-time scenes.

    SimBEV2X also supports dynamic weather (continuous weather shift \cite{sun2022shift}). If this feature is enabled, SimBEV2X randomly chooses the initial and final values of weather parameters for each scene and adjusts the weather at each time step to transition from the initial weather conditions at the start of a scene to the final weather conditions at the end of it, as demonstrated in \autoref{fig:dynamic-weather}. As before, the user has the option to set any subset of initial or final weather parameters to fixed values.
    
    \paragraph{Traffic} \label{par:traffic}
    
    SimBEV2X randomly chooses background vehicles from CARLA's vehicle library, which contains a variety of sedans, vans, trucks, bicycles, and motorcycles. The vehicle library also includes a few emergency vehicles whose emergency lights are turned on in each scene according to a user-specified probability. Vehicle colors are randomly selected from a large set of available colors (there are a few exceptions, for example, the firetruck is always red). For vehicles that have articulated doors, SimBEV2X randomly opens one or all door(s) according to a user-specified probability when those vehicles are stopped (for example, at a traffic light).

    \begin{table}[t]
        \centering
        \setlength{\abovecaptionskip}{0 pt}
        \caption{Vehicle Sensor Arrangements} \label{table:vehicle-sensors}
        \begin{tabular}{l c c c c}
            \toprule
            \textbf{Sensor} & \textbf{$x$ (m)} & \textbf{$y$ (m)}& \textbf{$z$ (m)} & \textbf{$\gamma$ (deg)} \\
            \toprule
            Front left camera & 0.4 & 0.4 & 1.6 & 55 \\
            Front camera & 0.6 & 0.0 & 1.6 & 0\\
            Front right camera & 0.4 & -0.4 & 1.6 & -55 \\
            Back left camera & 0.0 & 0.4 & 1.6 & 110 \\
            Back camera & -1.0 & 0.0 & 1.6 & 180 \\
            Back right camera & 0.0 & -0.4 & 1.6 & -110 \\
            Left radar & 0.0 & 1.0 & 0.6 & 90 \\
            Front radar & 2.4 & 0.0 & 0.6 & 0 \\
            Right radar & 0.0 & -1.0 & 0.6 & -90 \\
            Back radar & -2.4 & 0.0 & 0.6 & 180 \\
            Lidar & 0.0 & 0.0 & 1.8 & 0 \\
            \bottomrule
        \end{tabular}
        \vspace*{-4 pt}
    \end{table}
    \begin{table}[t]
        \centering
        \setlength{\abovecaptionskip}{0 pt}
        \caption{RSU Sensor Arrangements} \label{table:rsu-sensors}
        \begin{tabular}{l c c c c}
            \toprule
            \textbf{Sensor} & \textbf{$x$ (m)} & \textbf{$y$ (m)}& \textbf{$z$ (m)} & \textbf{$\gamma$ (deg)} \\
            \toprule
            Front left camera & 0.0 & 0.0 & 1.6 & 55 \\
            Front camera & 0.0 & 0.0 & 1.6 & 0\\
            Front right camera & 0.0 & 0.0 & 1.6 & -55 \\
            Radar & 0.0 & 0.0 & 1.2 & 0 \\
            Lidar & 0.0 & 0.0 & 3.2 & 0 \\
            \bottomrule
        \end{tabular}
        \vspace*{-4 pt}
    \end{table}
    
    SimBEV2X randomly chooses each vehicle's maximum speed (relative to the speed limit, for example, 10\% over/under) and how close vehicles can get to each other when they come to a stop. SimBEV2X also randomly chooses how long each traffic light stays green, though the user has the option to set any of these parameters to a fixed value.
    
    SimBEV2X randomly chooses pedestrians from CARLA's walker library, which contains models of different ages, genders, races, and body types. Then it randomly sets their walking speed and gives each a random destination to go to. CARLA 0.9.16 added support for wheelchaired pedestrians, so the user can specify what percentage of the pedestrians should use a wheelchair.
    
    \paragraph{Lights} \label{par:lights}
    
    SimBEV2X automatically manages street and building lights whenever the sun is below the horizon, whether that is at the start of a scene or occurs in the middle when dynamic weather is enabled. The user has the option to turn off all street and/or building lights, to randomize the color of building lights, to change the intensity of all street lights by a fixed (if desired, random) value to make them look brighter or dimmer, and to randomly turn off individual street lights to resemble broken ones in the real world, according to a user-specified probability.
    
    \paragraph{Reckless driving and jaywalking} \label{par:reckless}
    
    In SimBEV2X, the user can specify the percentage of vehicles (including data collection vehicles) that are distracted or reckless, creating interesting edge cases. Distracted vehicles ignore traffic lights and signs. Reckless vehicles ignore other vehicles and pedestrians as well. The user also controls the percentage of pedestrians that are allowed to cross the road at any point, not just at crosswalks.

    \begin{figure*}[!t]
        \centering
        \includegraphics[width=0.2\textwidth]{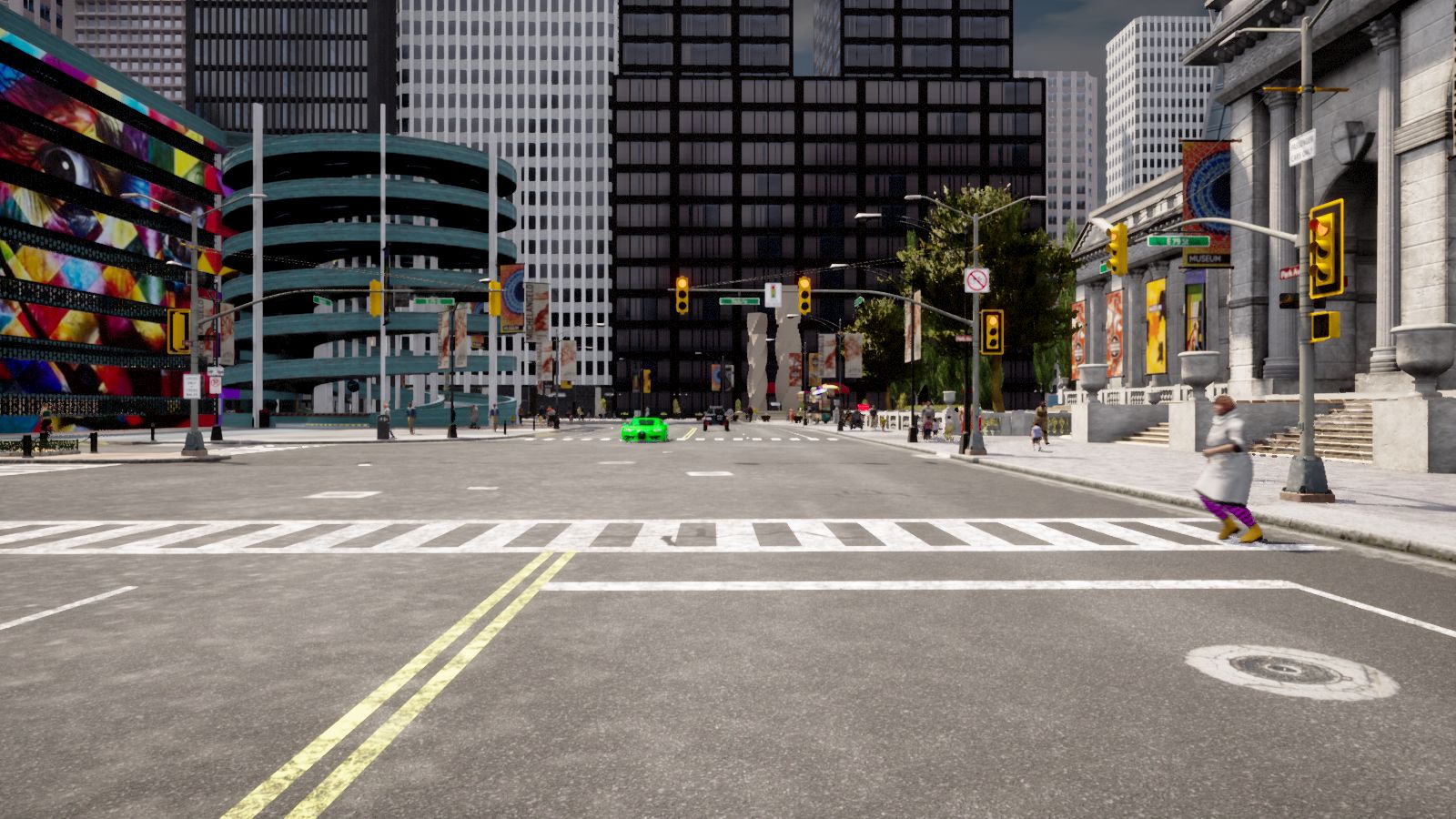}\hspace{-3.5pt}
        \vspace*{2 pt}
        \includegraphics[width=0.2\textwidth]{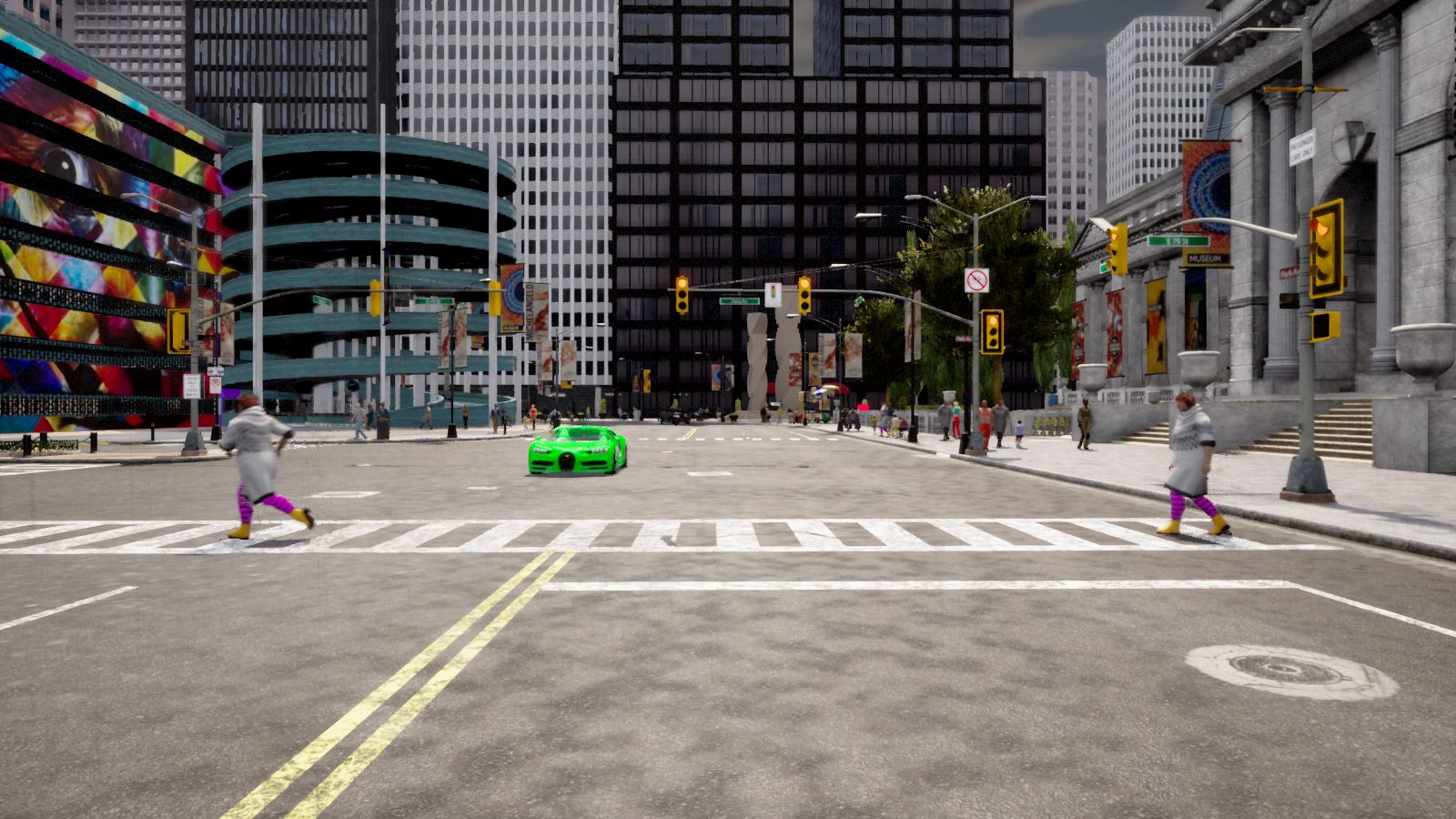}\hspace{-3.5pt}
        \includegraphics[width=0.2\textwidth]{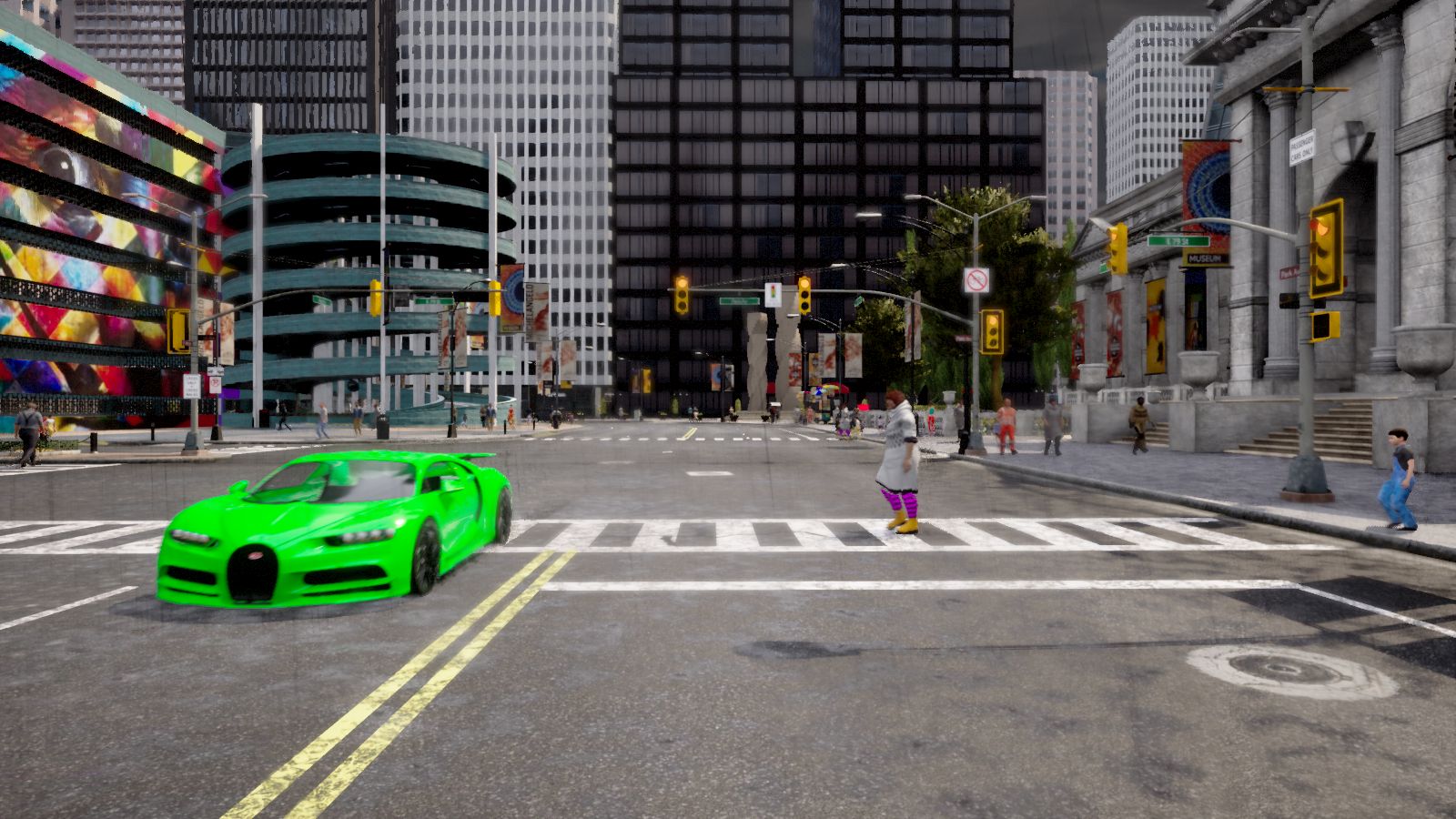}\hspace{-3.5pt}
        \includegraphics[width=0.2\textwidth]{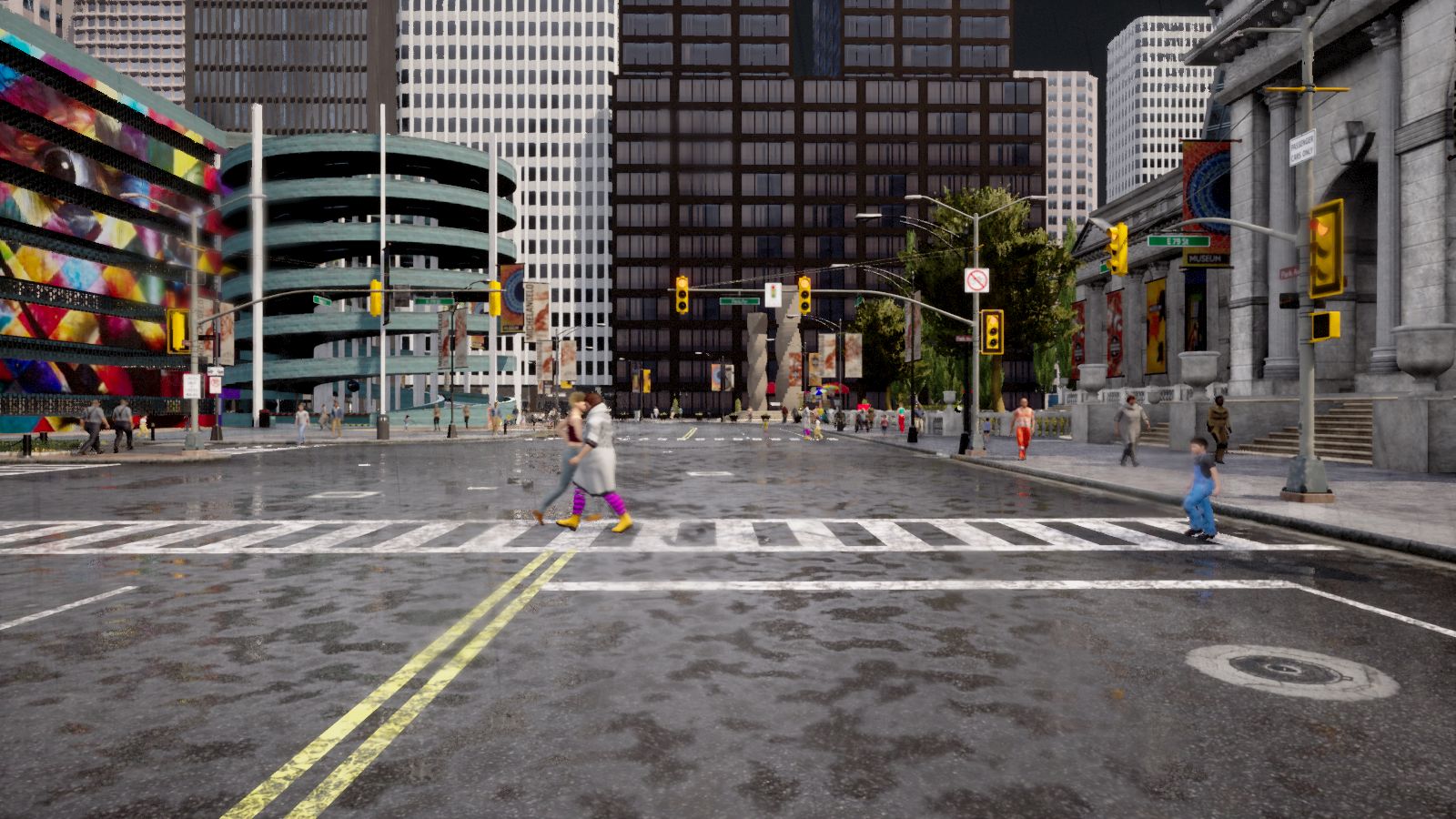}\hspace{-3.5pt}
        \includegraphics[width=0.2\textwidth]{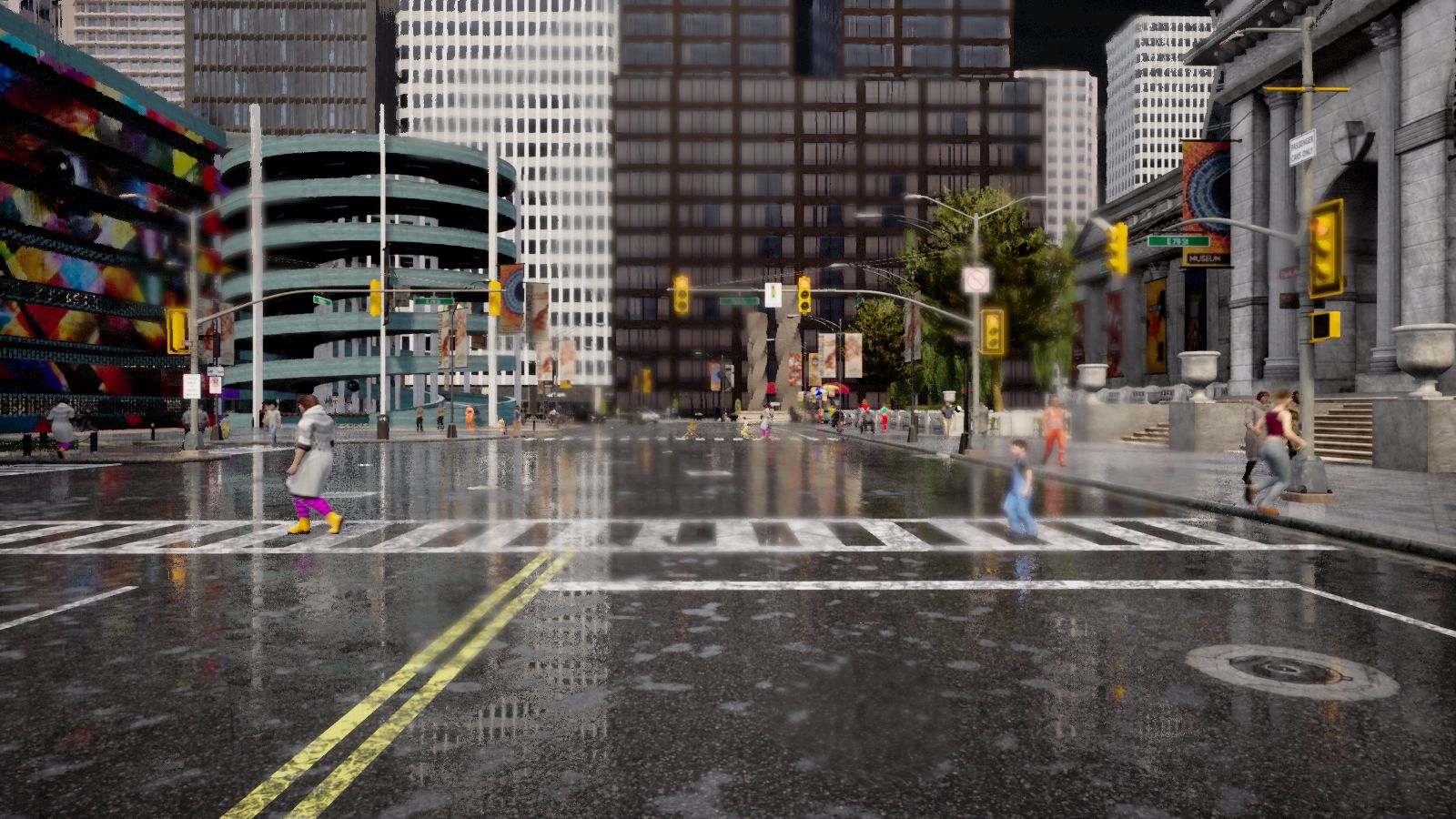}\hspace{-3.5pt}
        \includegraphics[width=0.2\textwidth]{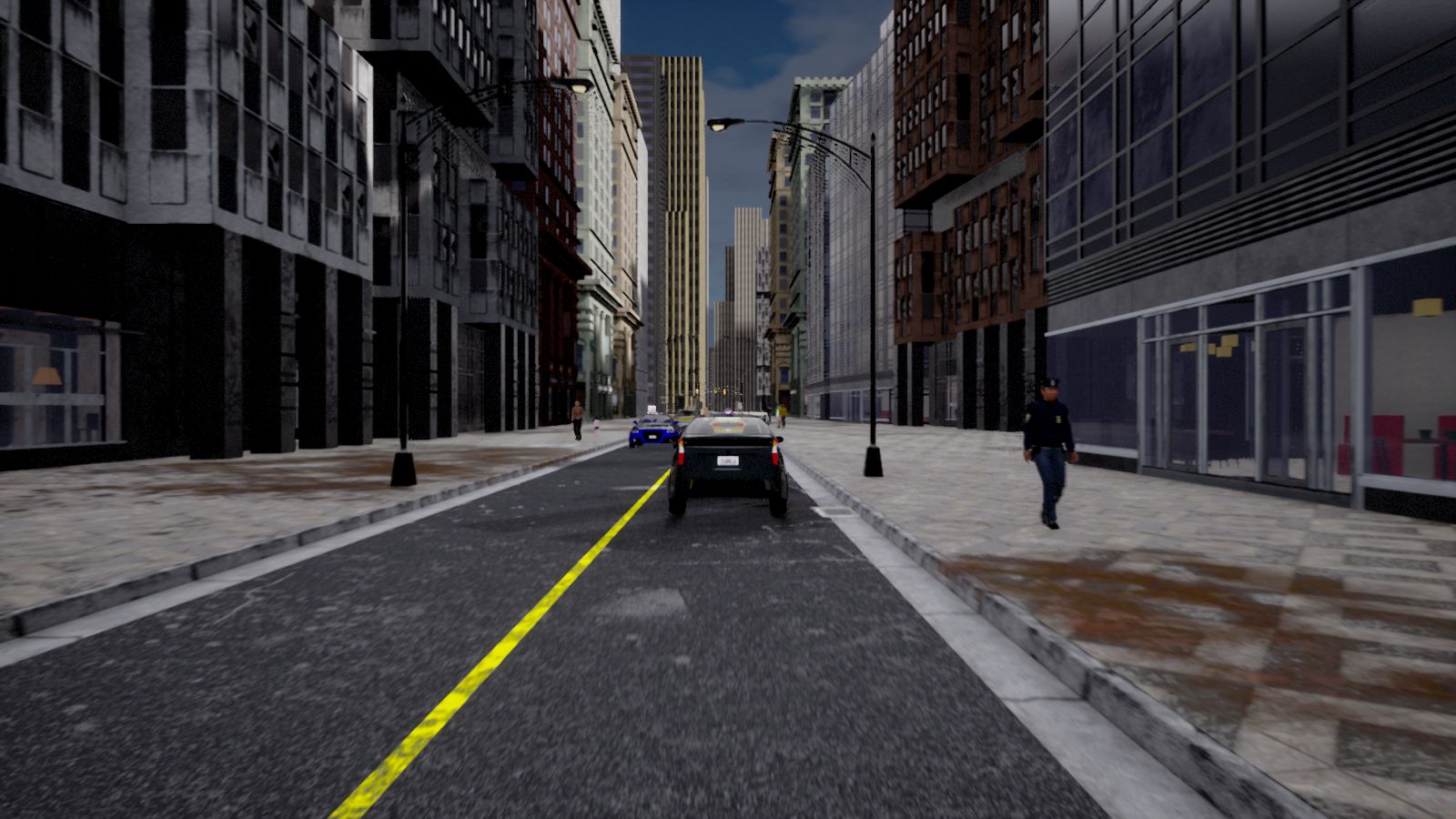}\hspace{-3.4pt}
        \vspace*{2 pt}
        \includegraphics[width=0.2\textwidth]{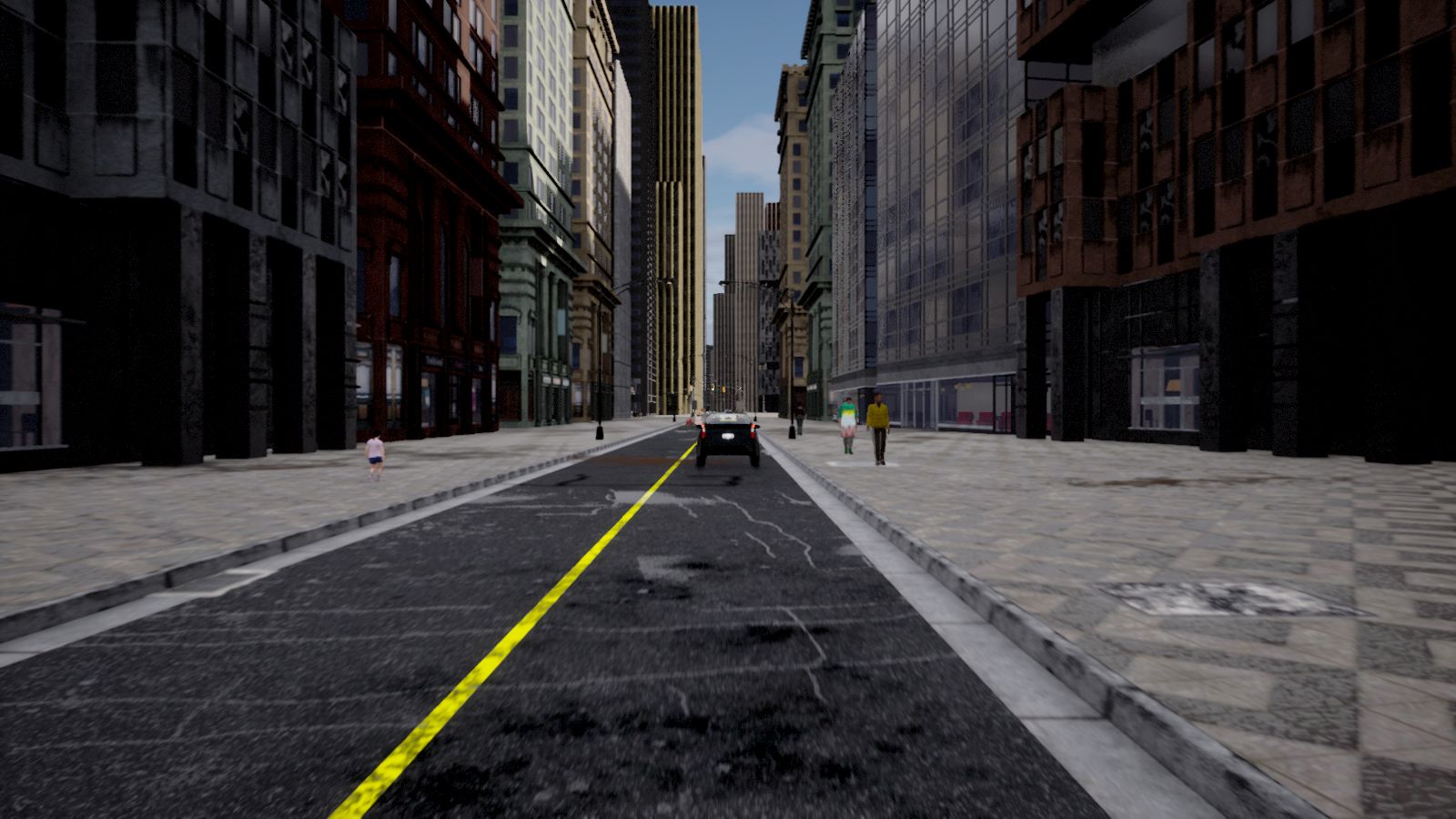}\hspace{-3.4pt}
        \includegraphics[width=0.2\textwidth]{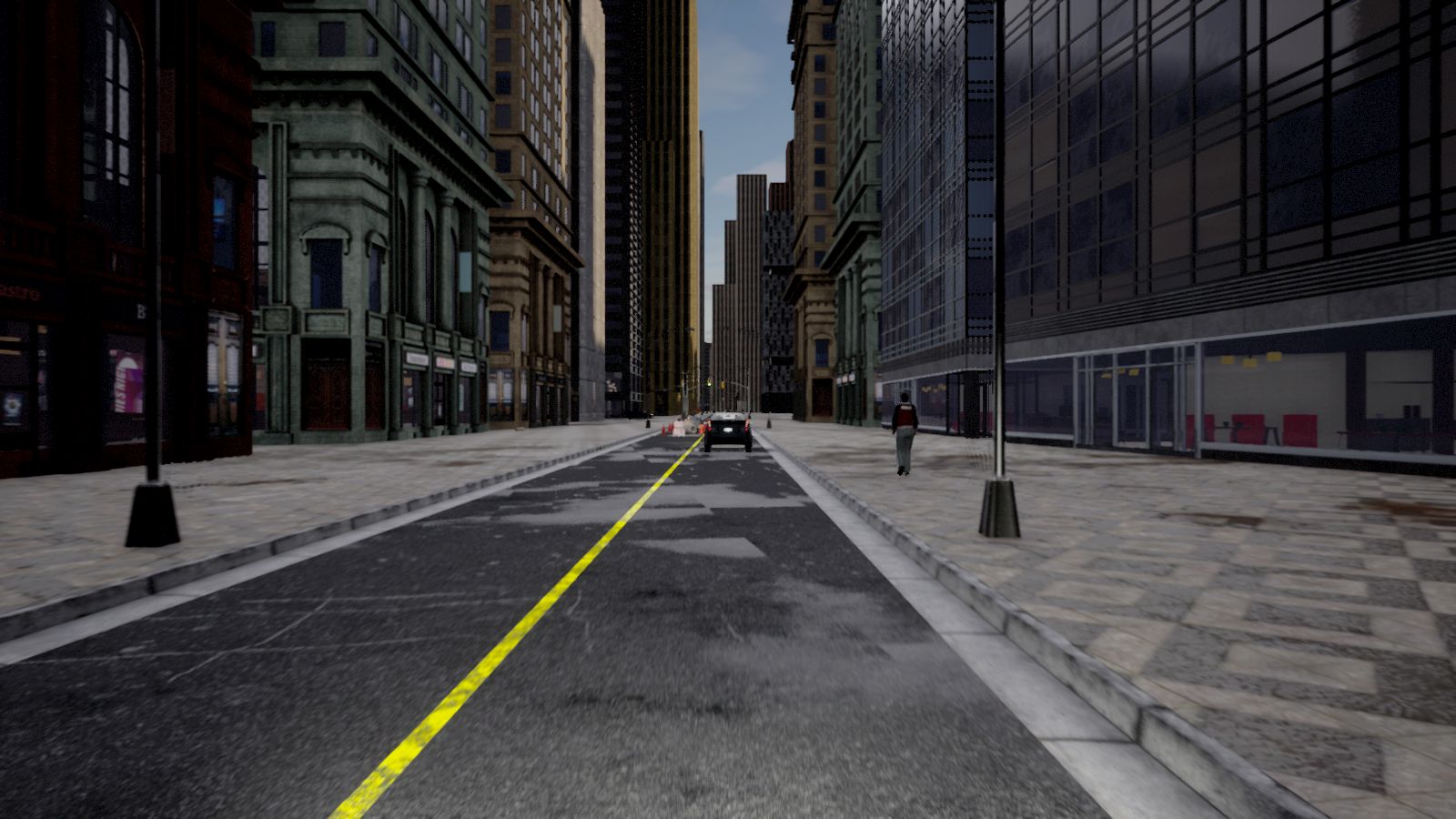}\hspace{-3.4pt}
        \includegraphics[width=0.2\textwidth]{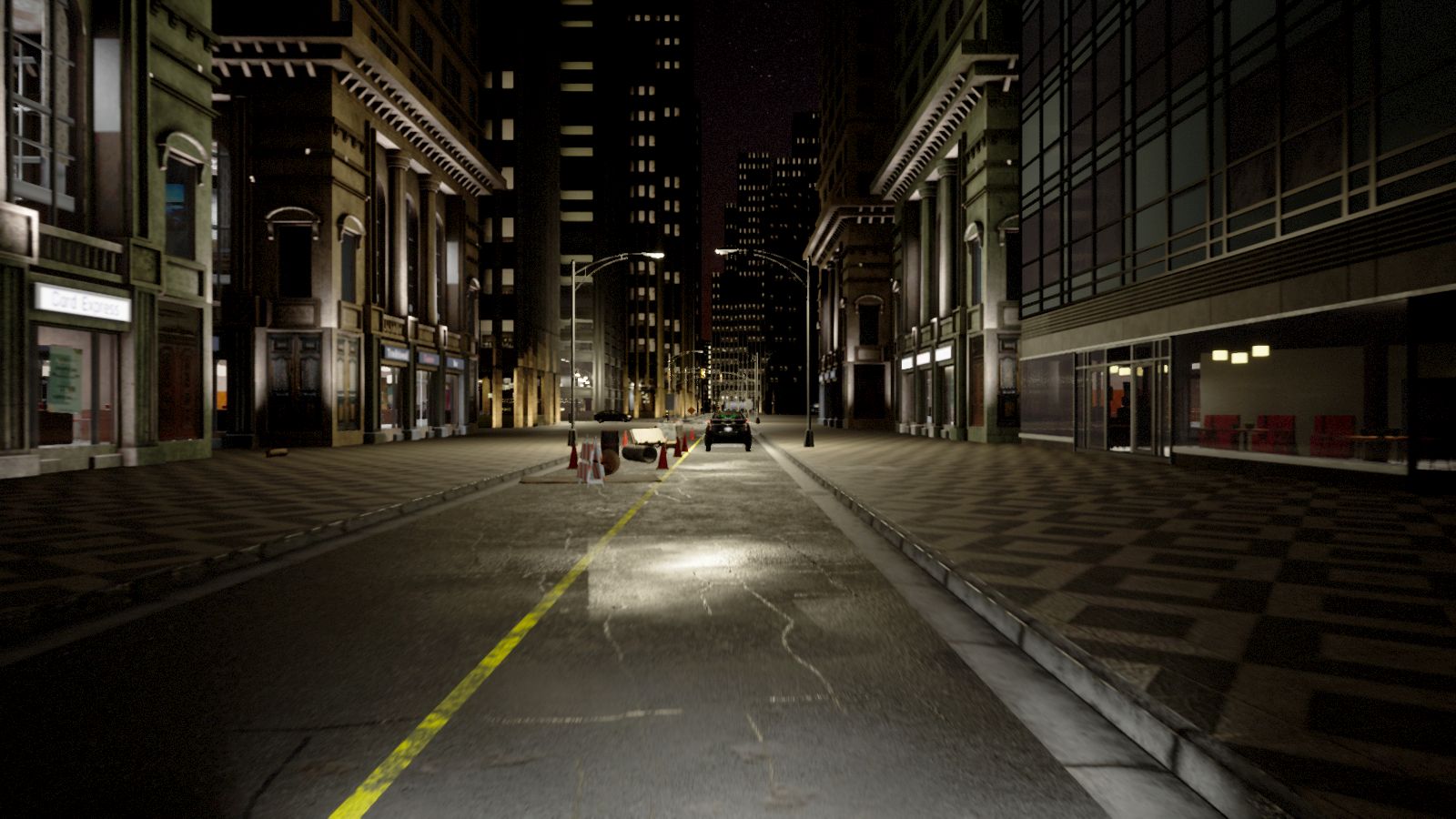}\hspace{-3.4pt}
        \includegraphics[width=0.2\textwidth]{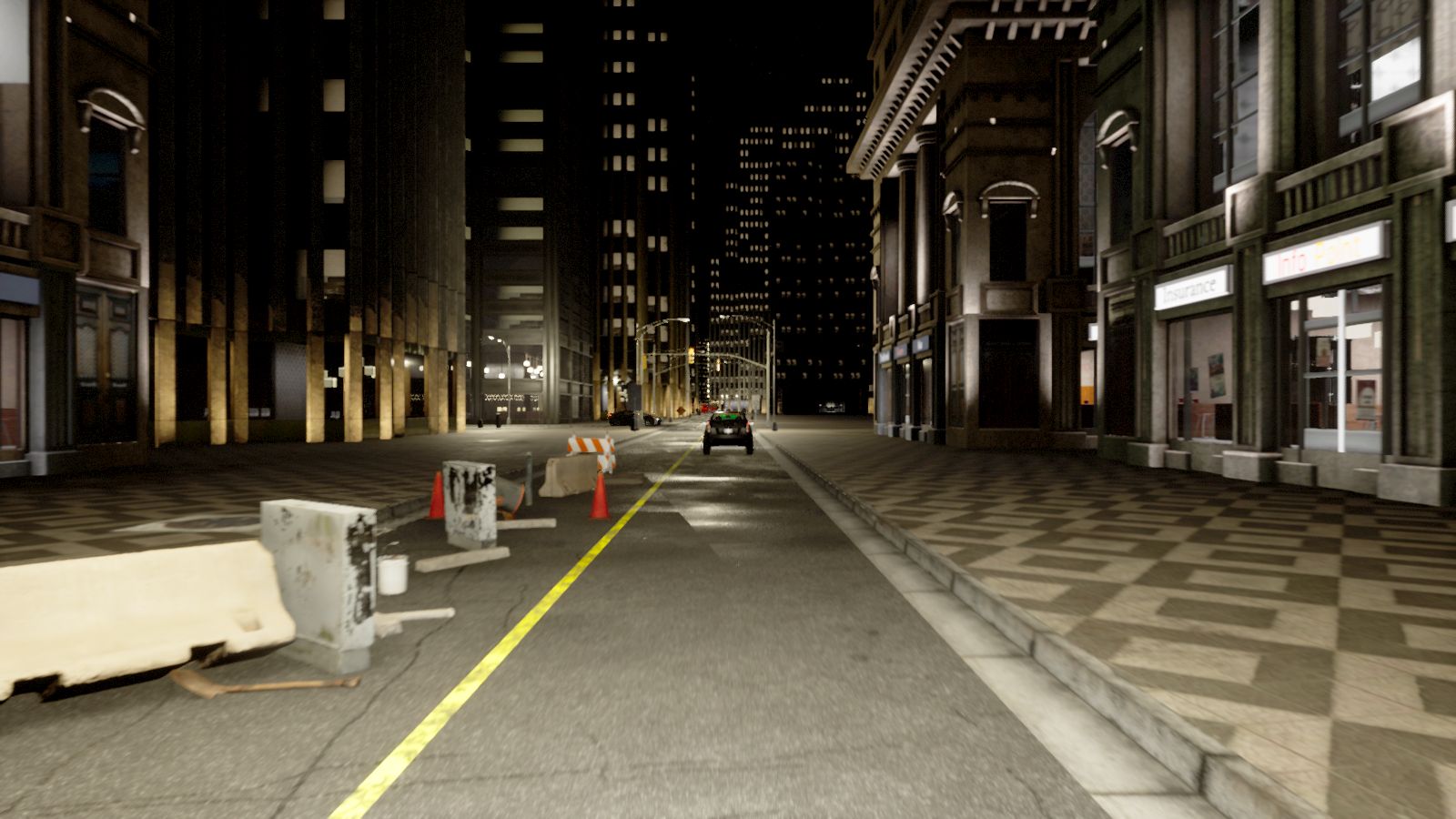}\hspace{-3.4pt}
        \includegraphics[width=0.2\textwidth]{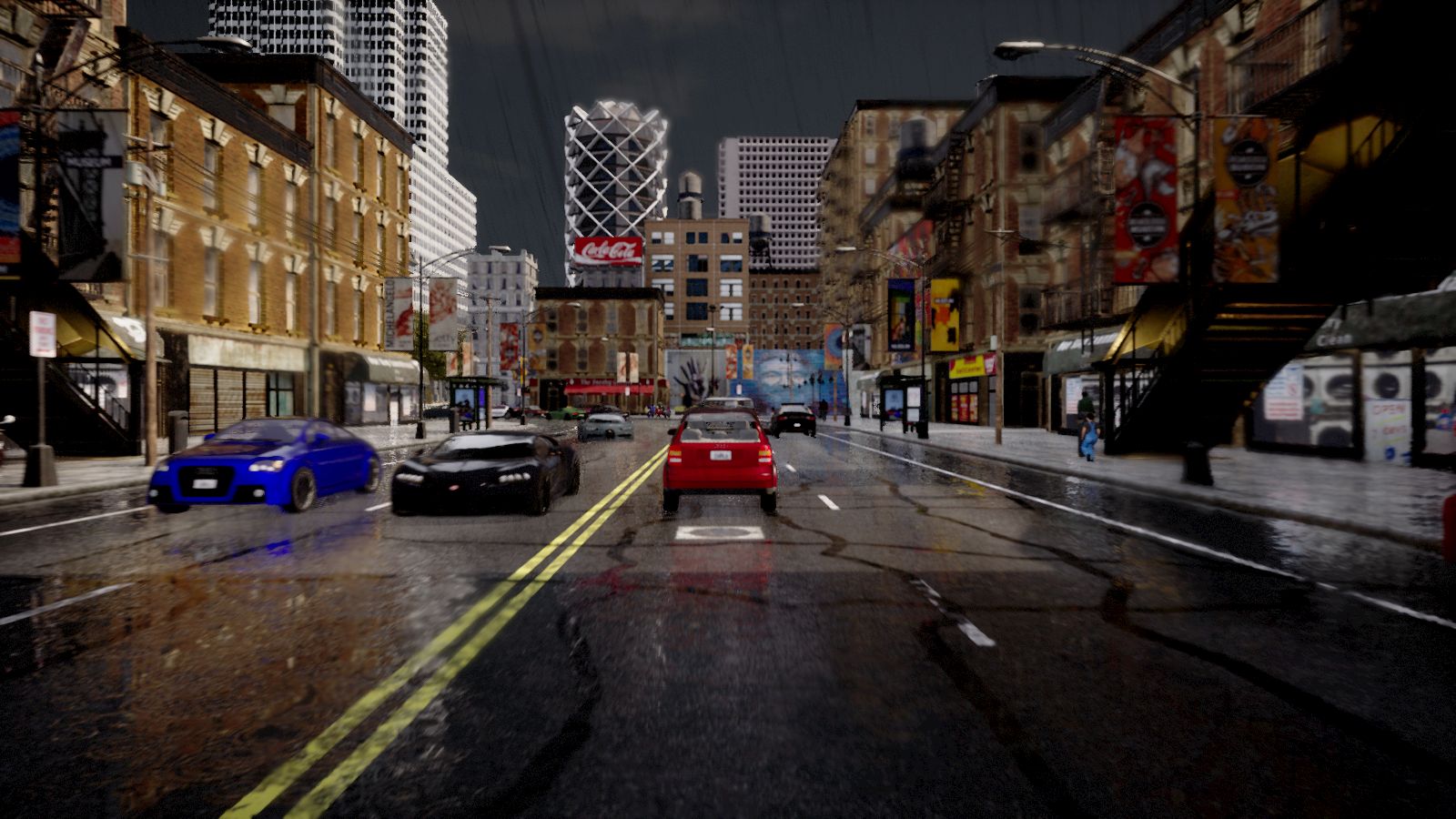}\hspace{-3.4pt}
        \includegraphics[width=0.2\textwidth]{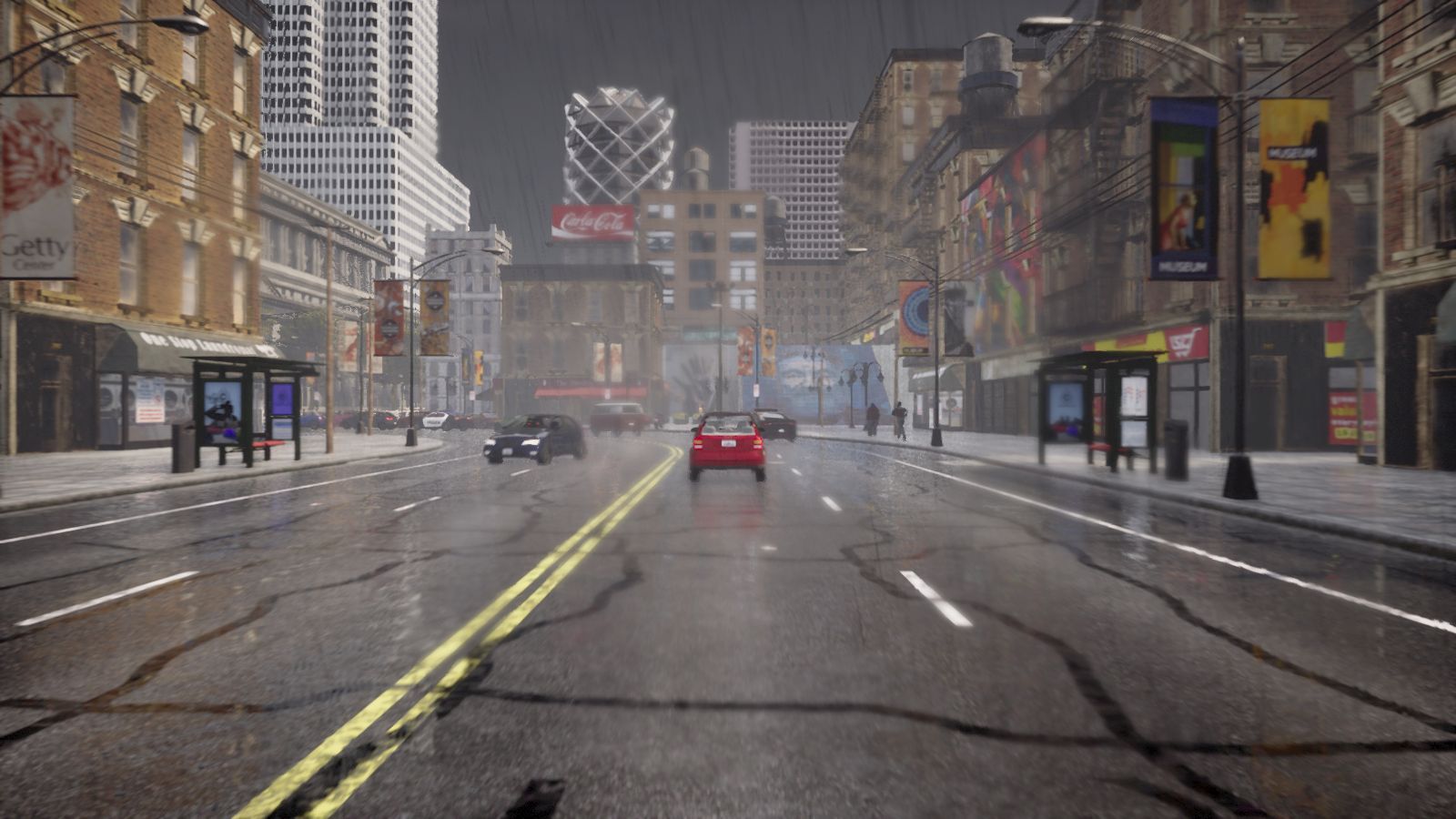}\hspace{-3.4pt}
        \includegraphics[width=0.2\textwidth]{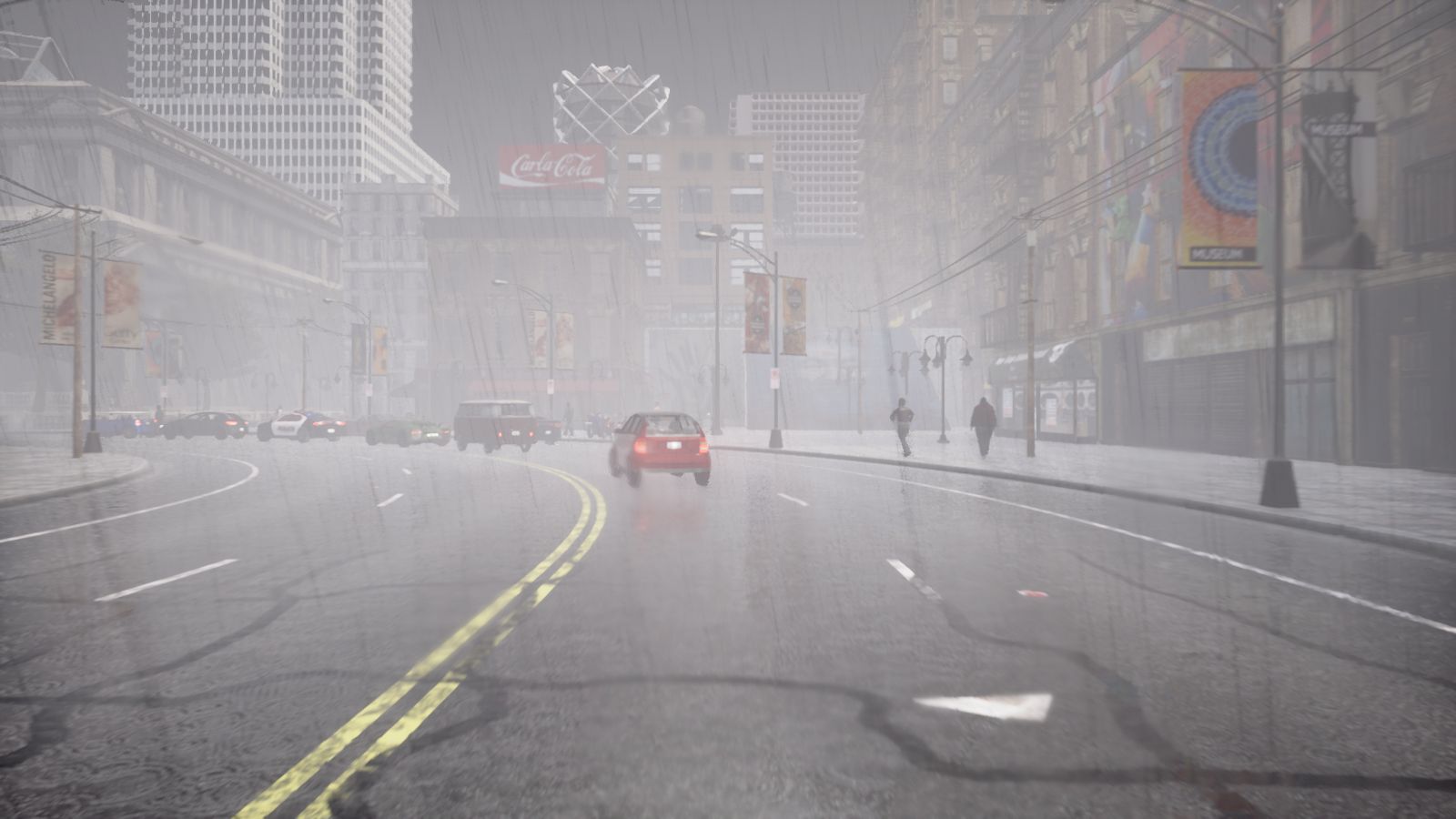}\hspace{-3.4pt}
        \includegraphics[width=0.2\textwidth]{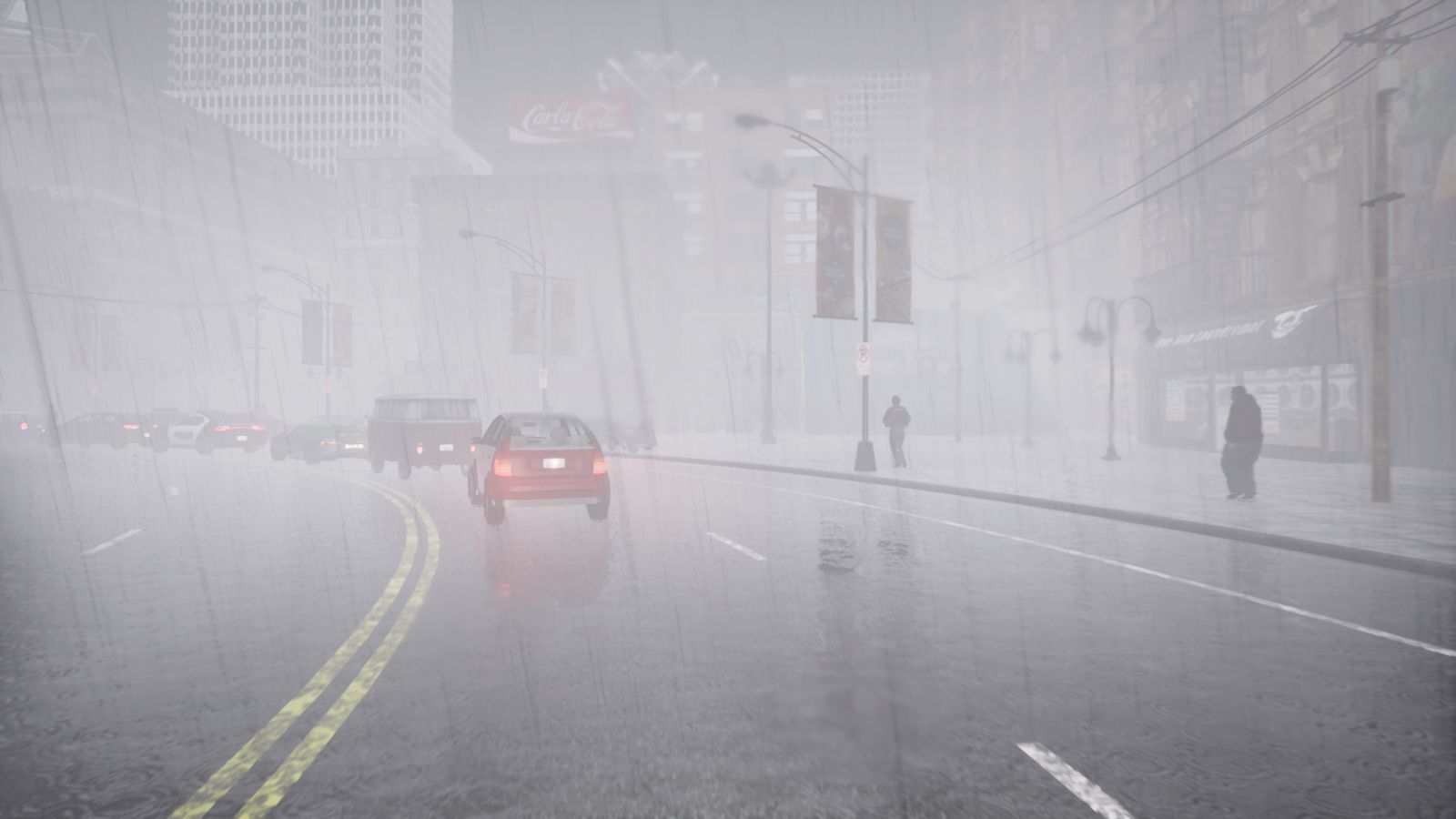}\hspace{-3.4pt}
        \includegraphics[width=0.2\textwidth]{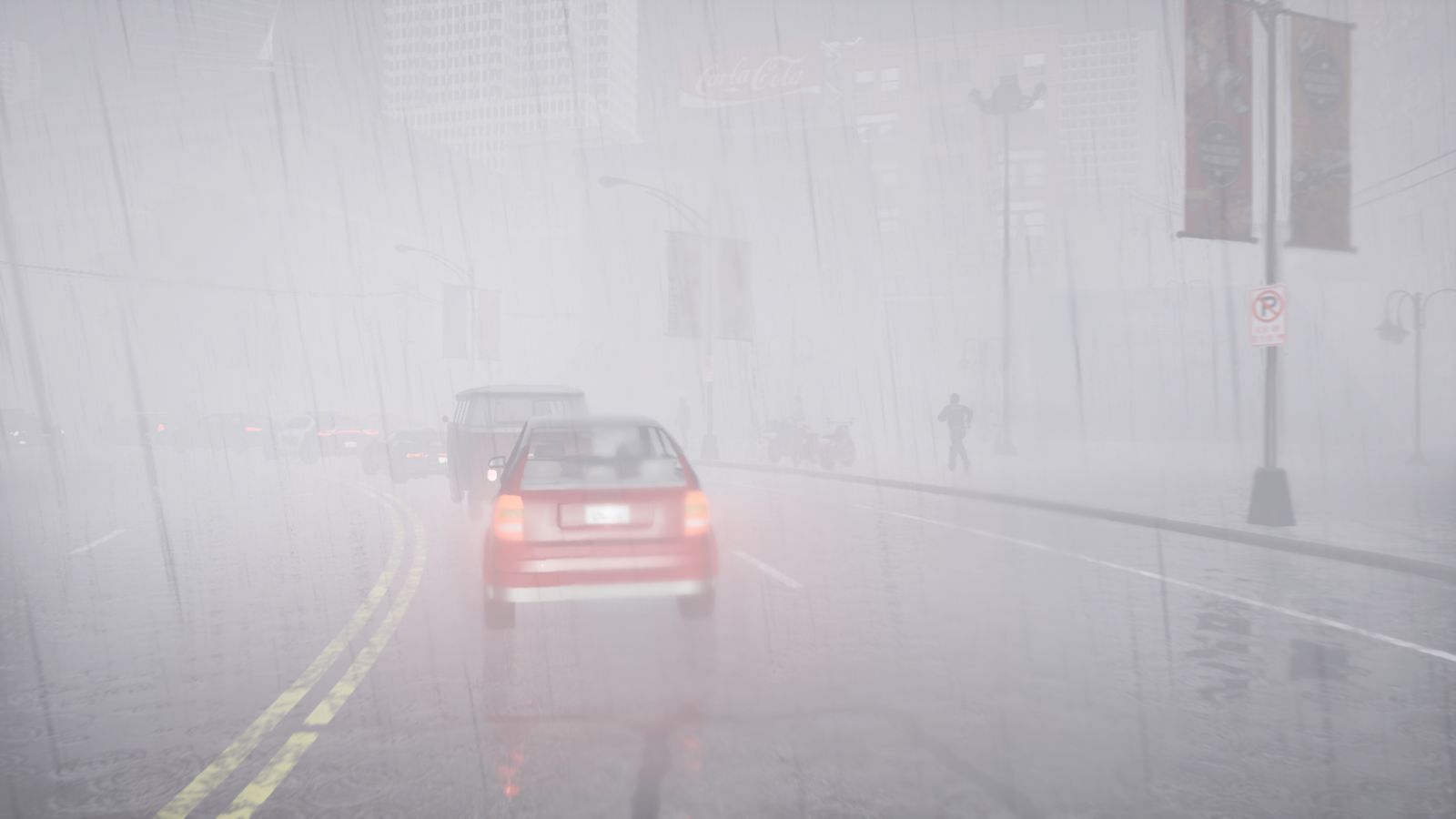}\hspace{-3.4pt}
        \setlength{\abovecaptionskip}{0 pt}
        \caption{Transition of weather from cloudy to rainy (top), day to night (middle), and clear to dense fog (bottom).}
        \label{fig:dynamic-weather}
        \vspace*{-8 pt}
    \end{figure*}

    \paragraph{Hazard areas} \label{par:hazard-areas}

\begin{figure}[!t]
        \centering
        \includegraphics[width=0.389\columnwidth]{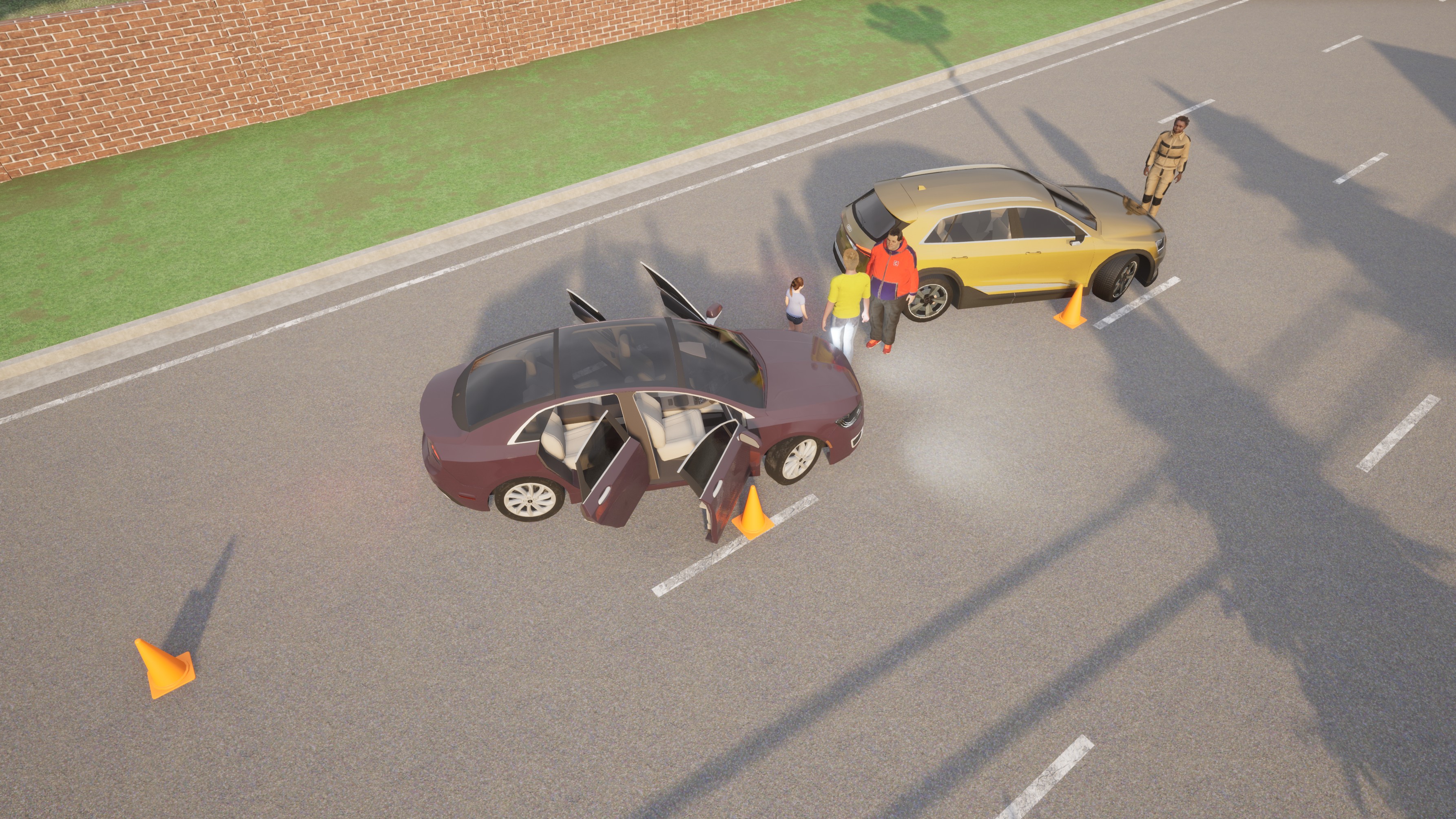}\hspace{-3.2 pt}
        \vspace*{2 pt}
        \includegraphics[width=0.389\columnwidth]{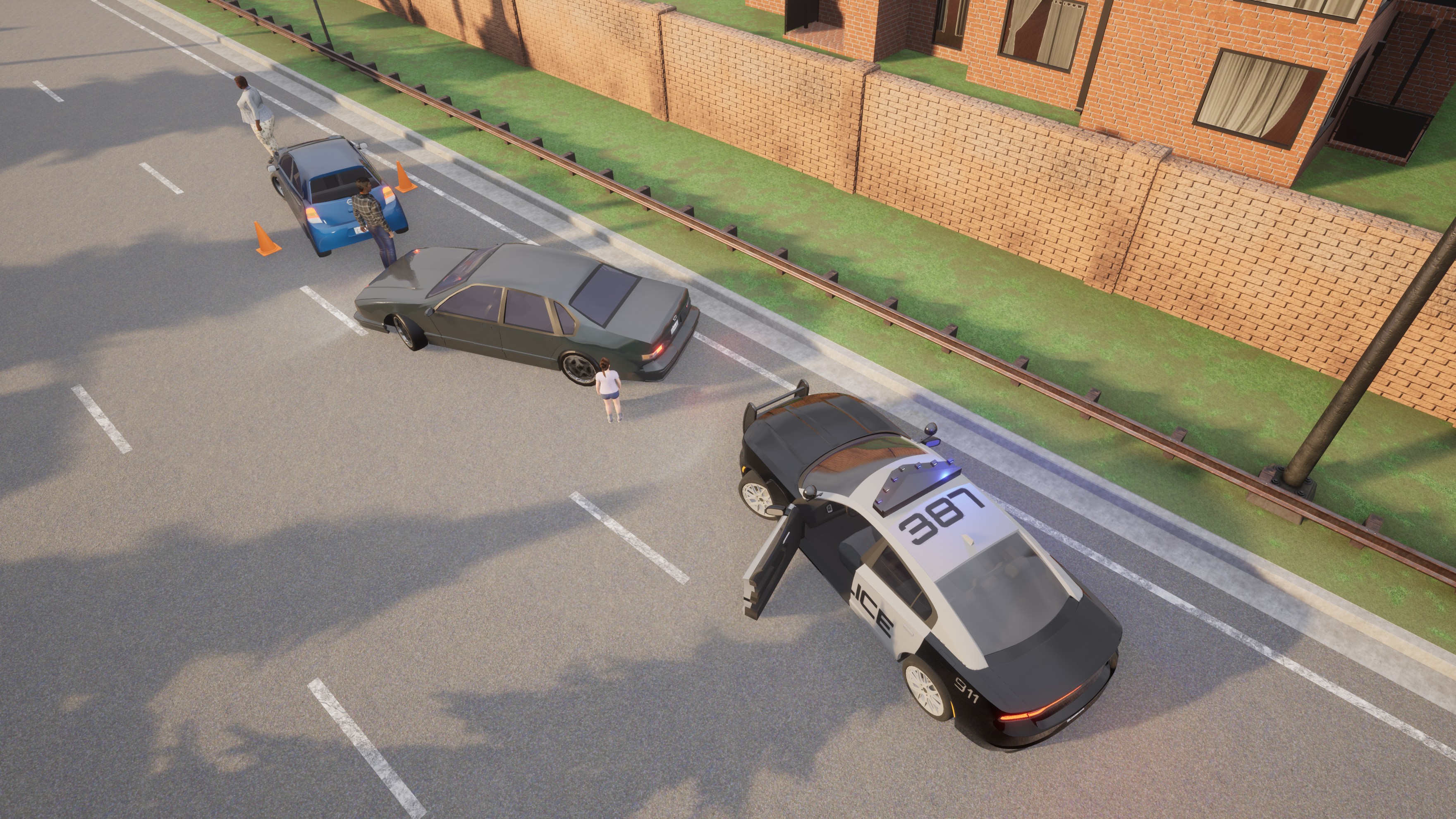}\hspace{-3.2 pt}
        \includegraphics[width=0.219\columnwidth]{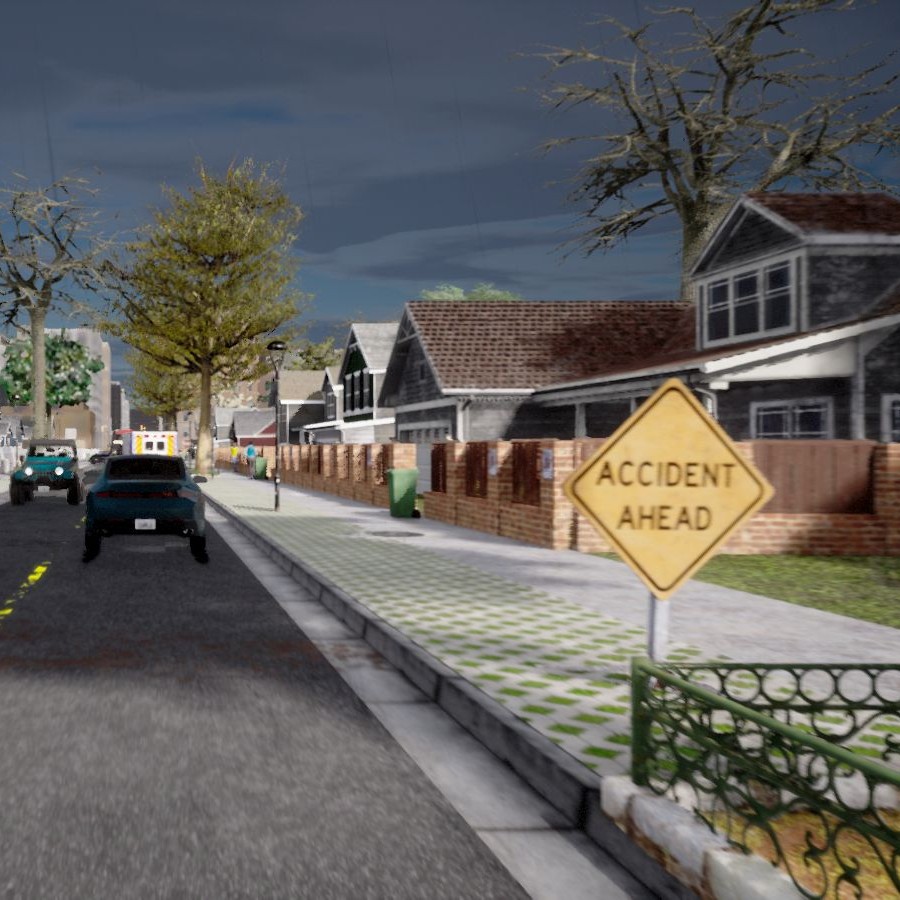}
        \includegraphics[width=0.389\columnwidth]{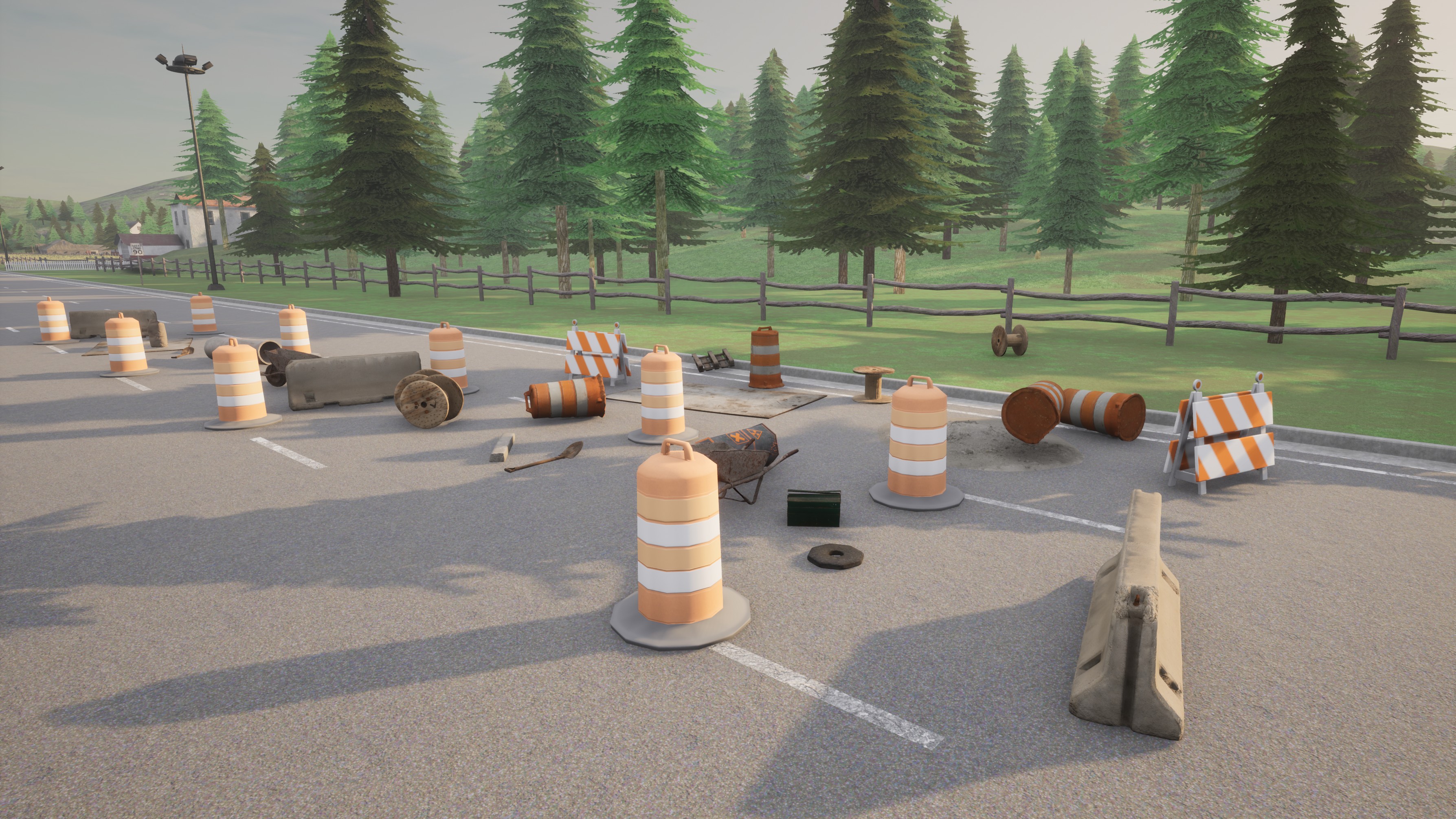}\hspace{-3.2 pt}
        \includegraphics[width=0.389\columnwidth]{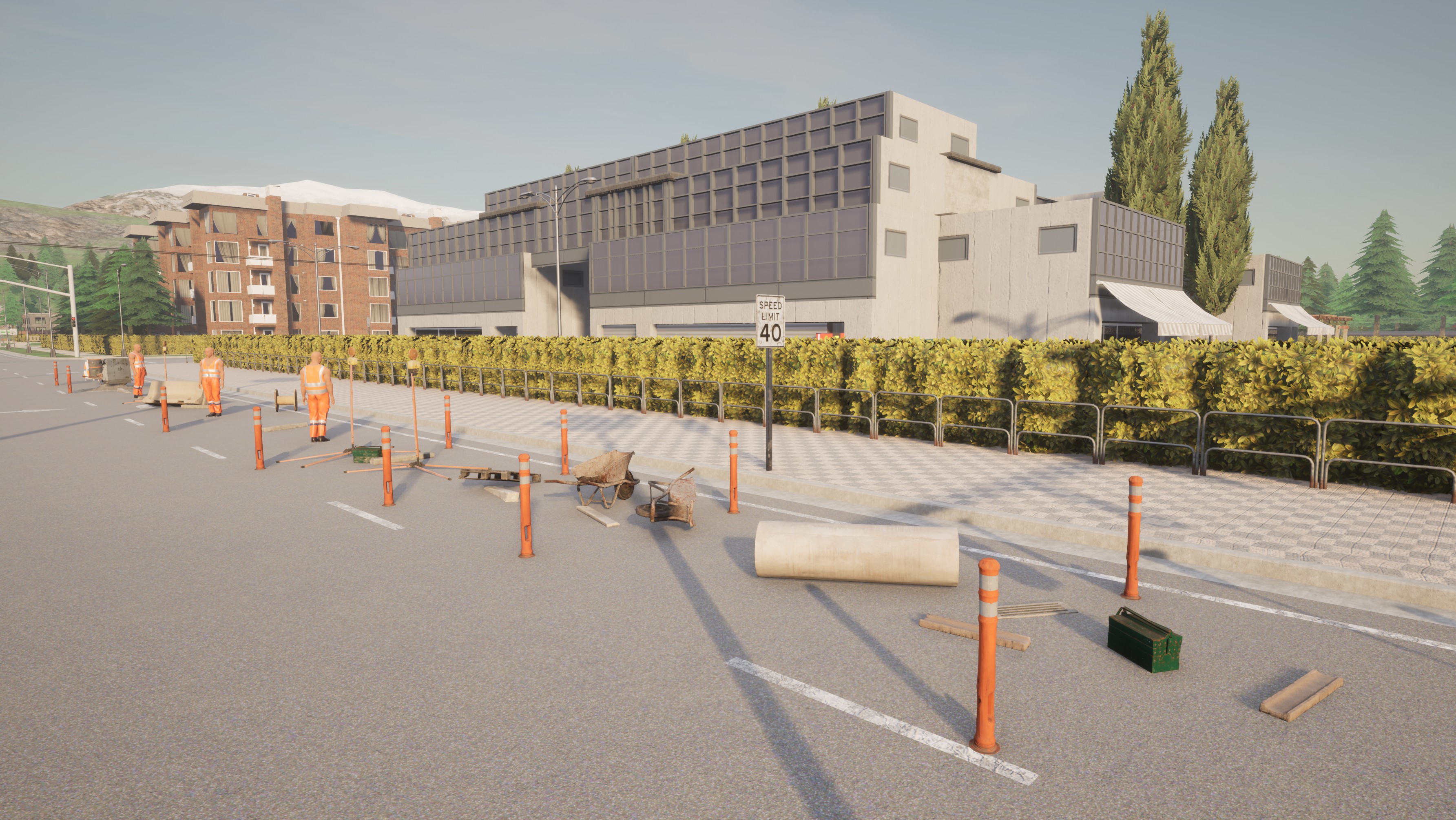}\hspace{-3.2 pt}
        \includegraphics[width=0.219\columnwidth]{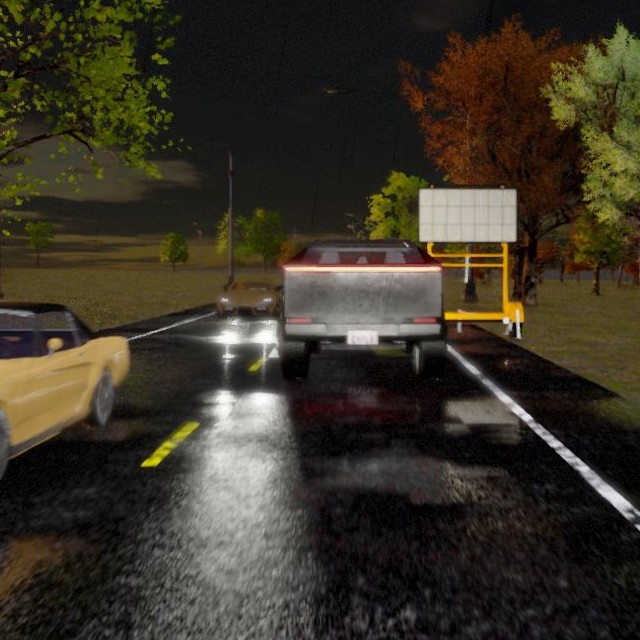}
        \setlength{\abovecaptionskip}{-12 pt}
        \caption{Randomly generated traffic accidents and the associated warning sign (top), and road work areas and an associated warning sign (bottom).}
        \label{fig:hazards}
        \vspace*{-4 pt}
    \end{figure}

    A novel feature of SimBEV2X is the random generation of traffic accidents and road work areas, collectively known as hazard areas and shown in \autoref{fig:hazards}. The user controls their overall prevalence and the proportion of each type.

    To create a traffic accident, SimBEV2X randomly selects an available spawn point and places two vehicles (which may also be motorcycles or bicycles) immediately in front of and behind it, as if they were involved in an accident. A traffic cone or an emergency vehicle (ambulance, fire truck, or police car with its emergency lights turned on) is then placed behind the rear vehicle according to a user-defined probability. If an emergency vehicle is present, a warning sign is placed at a random distance ahead of the accident (between 40 and 160 m) on the shoulder of the road to alert approaching traffic. Up to three pedestrians (which may be police officers if a police car is present) are also spawned in each of the spaces between the vehicles and ahead of the front vehicle. Traffic cones are placed on both sides of the vehicles involved in the accident, and some or all of the doors of the vehicles present at the scene are opened randomly.

    To create a road work area, SimBEV2X randomly selects an available spawn point and marks it as the beginning of the road work area by placing a road barrier there. Then, step by step, it moves forward in that lane by a random amount (within some bounds) and randomly chooses an item from the following list to place on the road: construction equipment, a worker in high-visibility gear, or a street barrier that marks the end of the road work area. Finally, the area is lined on both sides with construction cones or barrels, and a warning sign is placed at a random distance ahead of the road work area (between 40 and 160 m) on the shoulder of the road to alert approaching traffic.

    For both types of hazard areas, the location and orientation of all objects placed on the road are subject to random variations, sometimes within a limited interval depending on the type of object. The user also controls the percentage of missing traffic/construction cones. For example, a value of 50\% implies that half of the cones that would have been placed on the road are missing.
    
    \subsection{Data Annotation} \label{subsec:simbev-gt}
    
    SimBEV2X provides four primary annotations for each agent at each simulation time step (frame): 3D object bounding boxes, HD map information, and BEV and 3D semantic occupancy ground truths. The output of some perception sensors, such as segmentation, depth, and optical flow cameras and semantic lidar, can also serve as annotations but are not discussed here.
    
    \paragraph{3D object bounding boxes} \label{par:bev-bbox}
    
    At each time step, SimBEV2X collects 3D bounding boxes of objects that are within a user-specified radius of each agent (vehicle and RSU) for the following ten classes: \textit{car}, \textit{truck} (includes trucks, vans, etc., but not buses), \textit{bus}, \textit{motorcycle}, \textit{bicycle}, \textit{pedestrian}, \textit{traffic light}, \textit{traffic sign}, \textit{traffic cone} (also includes construction cones and barrels), and \textit{barrier}. Additional object attributes such as the object's ID, location, orientation, linear and angular velocity, and distance and angle to the agent are also collected. For vehicles, the make, model, and color are also collected. For traffic signs, the sign's type (stop, yield, speed limit, etc.) and for traffic lights, the current state and light timings are also collected.

    \paragraph{HD map information} \label{par:hd-map-gt}

    SimBEV2X collects the following information about the waypoint at each data collection vehicle's location at each time step: OpenDRIVE \cite{dupuis2010opendrive} IDs of the road, section, and lane that the waypoint belongs to; the type of the lane (driving, parking, shoulder, etc.) and its width; the possibility of lane change on either side; whether the waypoint is inside an intersection or a junction (and if so, the ID of the junction); type, width, and color of the lane markings on both sides; and similar information about the lanes immediately to the left and right of the waypoint. This information can be combined with the OpenDRIVE data to create a more expansive vehicle-centered map of the environment.
    
    \paragraph{BEV ground truth} \label{par:bev-gt}
    
    SimBEV2X provides the BEV ground truth for the following 14 classes: \textit{road}, \textit{hazard}, \textit{road line}, \textit{sidewalk}, \textit{crosswalk}, \textit{traffic cone}, \textit{barrier}, \textit{car}, \textit{truck}, \textit{bus}, \textit{motorcycle}, \textit{bicycle}, \textit{rider} (person on a \textit{motorcycle} or \textit{bicycle}), and \textit{pedestrian}. At each time step, the BEV ground truth for each agent is saved as a $C \times l \times l$ binary array, where $C$ is the number of classes (14 in this case) and $l$ is the dimension of the BEV grid that is centered on the agent.

\begin{figure}[t]
        \centering
        \includegraphics[width=\columnwidth]{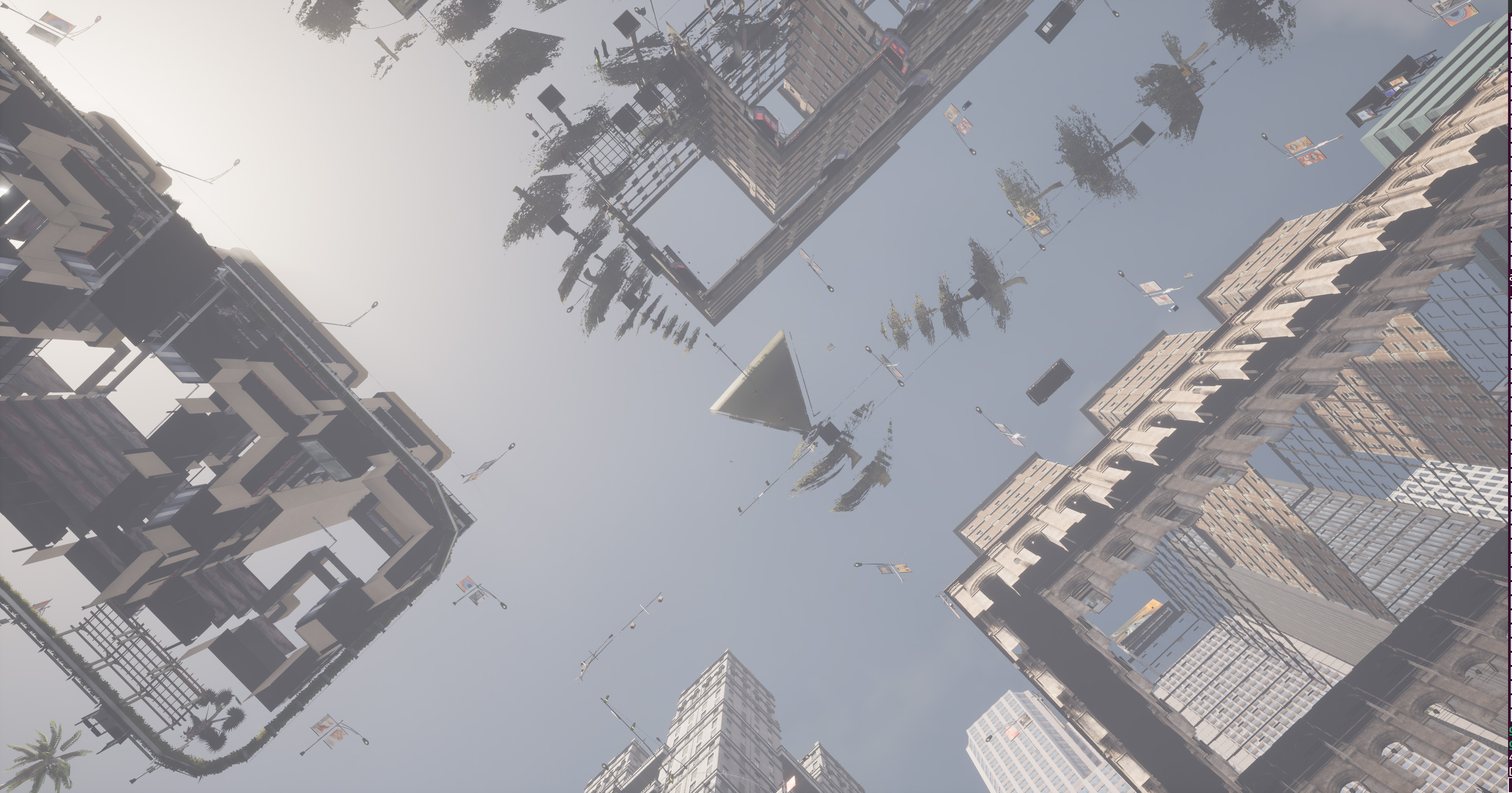}
        \setlength{\abovecaptionskip}{-12 pt}
        \caption{Ground elements (roads, sidewalks, etc.) in CARLA use one-way see-through materials, appearing invisible to a camera placed below them. We use this property to capture an accurate BEV ground truth by placing a camera below each agent facing up.}\label{fig:invis-road}
        \vspace*{-4 pt}
    \end{figure}

    To calculate the BEV ground truth for non-map classes (\textit{traffic cone}, \textit{barrier}, \textit{car}, \textit{truck}, \textit{bus}, \textit{motorcycle}, \textit{bicycle}, \textit{rider}, \textit{pedestrian}), we use the fact that ground elements in CARLA (roads, sidewalks, etc.) use one-way see-through materials, appearing solid from one direction but see-through from the other, as shown in \autoref{fig:invis-road}. SimBEV2X takes advantage of this by placing a semantic segmentation camera at a user-defined distance (by default 1 km, far enough to minimize perspective distortion) above the agent facing down, and another far below the agent facing up to capture what the overhead camera may not be able to due to obstructions (bridges, light poles, etc.) Both cameras have a $l \times l$ resolution, and their FoV is set in a way that each pixel represents a $d \times d$ area on the ground. The BEV ground truth for each non-map class is obtained by merging data from the two cameras using a \textit{logical or} operation. By default, $l$ is set to 400 and $d$ is set to 0.4 m, creating a 160 m $\times$ 160 m box around the agent. This area is larger than what is typically used for BEV segmentation (100 m $\times$ 100 m) and is intended to help with the augmentation (rotation, translation, scaling) of collected training data.
    
\begin{figure}[t]
        \centering
        \fbox{\includegraphics[width=0.32\columnwidth]{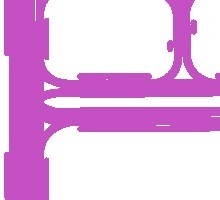}}\hspace{-2pt}
        \fbox{\includegraphics[width=0.32\columnwidth]{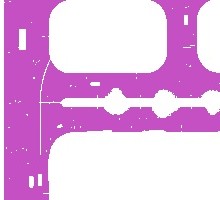}}\hspace{-2pt}
        \fbox{\includegraphics[width=0.32\columnwidth]{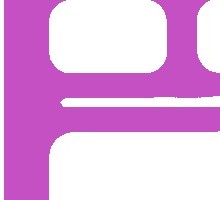}}
        \setlength{\abovecaptionskip}{0 pt}
        \caption{Left: BEV road data calculated using the OpenDRIVE map; there are clear gaps where lanes diverge. Middle: BEV road data obtained from the overhead camera; gaps are present due to the obstruction of the overhead view by vehicles and vegetation. Right: the BEV \textit{road} ground truth obtained from combining the two sources of data and performing \textit{binary closing}.}\label{fig:road-mask}
    \end{figure}

    To obtain the BEV ground truth for the \textit{road}, \textit{road line}, \textit{sidewalk}, and \textit{crosswalk} classes, SimBEV2X parses the OpenDRIVE data for the environment before the start of each scene and catalogs a list of road sections, sidewalks, and crosswalks within an area surrounding each agent (the radius of this area is determined based on the duration of each scene to reduce computation time). From this list, SimBEV2X calculates and stores the boundary points of each section. For the \textit{hazard} class, SimBEV2X compiles a list of hazard areas in a similar fashion and then calculates and stores the boundary points for each area. To further reduce computation time, a much smaller list is obtained from the master list every 0.5 simulation seconds that only contains those boundary points that are within a much smaller radius of each agent. Then, at each time step, the boundary points from this smaller list are used to calculate and label the corresponding area (or line in the case of the \textit{road line} class) for each class on the BEV grid. For the \textit{road} class, SimBEV2X combines the labels obtained through this method with data from the overhead camera and performs \textit{binary closing} to patch any potential gaps, obtaining a more accurate BEV ground truth. This process is illustrated in \autoref{fig:road-mask}.
    
    Overall, the approach discussed above reduces calculation time for each BEV ground truth array to an average of 20 ms (10 - 50 ms range depending on the environment). Another advantage of this approach is that multiple labels can be assigned to the same $d \times d$ cell using this approach. For example, a cell occupied by a cyclist will have a \textit{rider} (obtained from the overhead camera), a \textit{bicycle} (obtained from the below-ground camera), and a \textit{road} (calculated from the map data) label. The only instance where this approach yields inaccurate results is when roads with different elevations cross each other, e.g., when the agent is near an overpass. In those instances, SimBEV2X falls back to calculating the BEV ground truth for non-map classes using object bounding boxes and filters out the boundary points of map sections whose elevation difference with the agent is larger than 4.8 m. Although not as accurate as our primary approach, the resulting ground truth is still acceptable. SimBEV2X switches to this method automatically when it detects two waypoints within 48.0 m of each other that have an elevation difference of at least 6.4 m.

    \paragraph{3D semantic occupancy ground truth} \label{par:3d-semantic-occupancy-gt}

\begin{figure}[!t]
        \centering
        \includegraphics[width=\columnwidth]{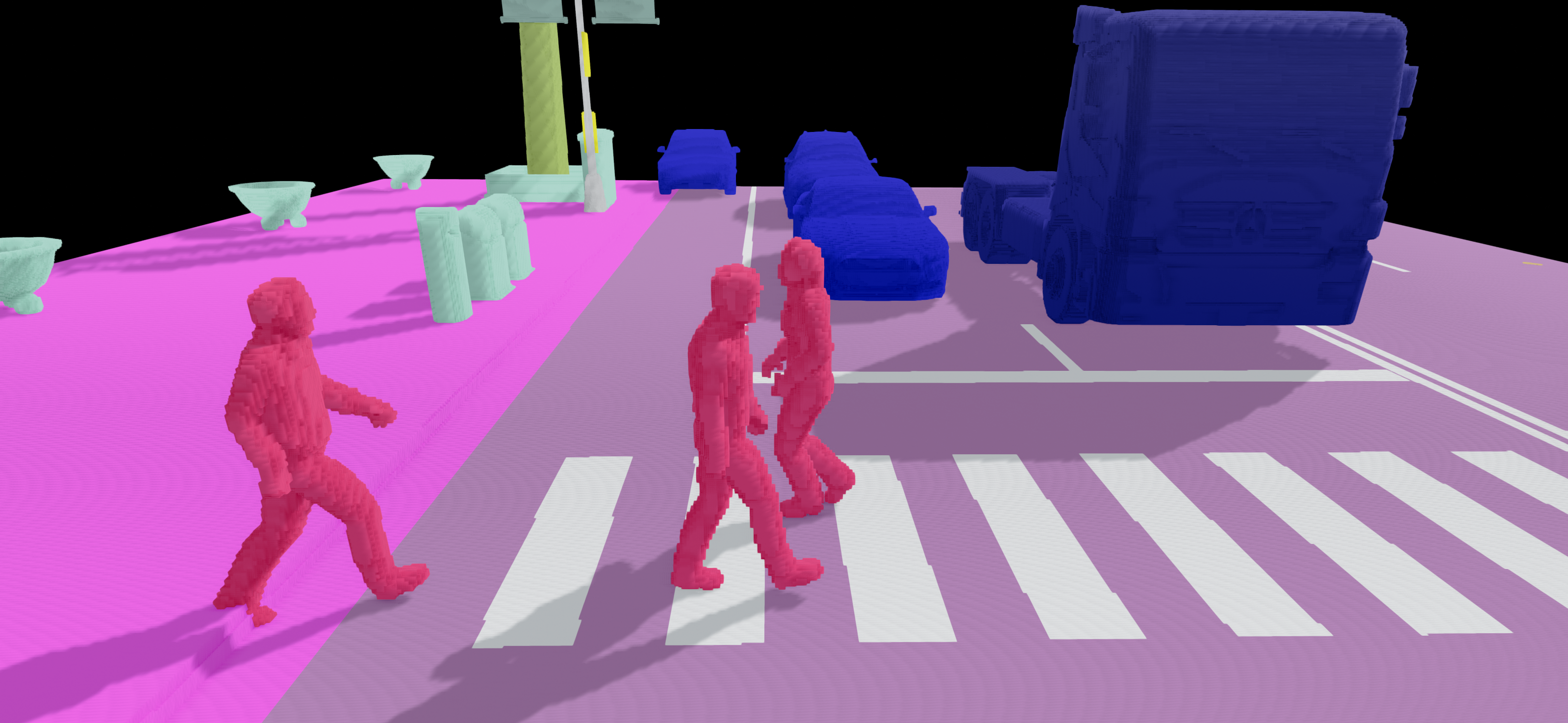}
        \setlength{\abovecaptionskip}{-12 pt}
        \caption{Our custom CARLA sensor can produce semantic voxel grids of arbitrary dimension and spatial resolution. This image portrays a semantic voxel grid generated using 2 cm voxels.}
        \label{fig:voxel-grid}
        \vspace*{-8 pt}
    \end{figure}

    Our approach to calculating the 3D semantic occupancy ground truth builds upon the work of \cite{wu2025synthetic}. Because CARLA lacks a native sensor for producing 3D semantic voxel grids, we developed a custom sensor utilizing Unreal Engine's collision and overlap detection functions, with the user having full control over the attributes of the sensor (including the dimensions and resolution of the grid), as shown in \autoref{fig:voxel-grid}.

    Unlike \cite{wu2025synthetic}, where semantic voxel grids are obtained by first detecting seed voxels for each object using a top-down box trace and then performing parallel Breadth-First Search (BFS) to label the remaining voxels, our method explicitly checks every voxel in the grid for overlap with any object and assigns the object's semantic tag to the voxel if one is found. In cases where a single voxel overlaps with multiple objects, we use the semantic label of one of said objects with the highest priority (as shown in \autoref{table:semantic-priority}). The key to our method is disabling the thread lock in Unreal Engine to perform these calculations in parallel. Doing so is thread-safe because the calculations are performed after the physics step, allowing us to achieve a performance that is on par with, if not superior to, \cite{wu2025synthetic}. 
    
    Our sensor also provides an alternative, faster but less accurate, calculation method. This second method performs parallel box traces along user-specified directions ($\pm x$, $\pm y$, and/or $\pm z$) and identifies and labels the initial collision point of the box with each object. If box traces are conducted in both $\pm z$ directions, the user has the option to automatically fill the intermediate space as a continuous scanline or explicitly check each intermediate voxel for object overlap and label it.

    \subsection{Post-processing} \label{subsec:post-processing}

    SimBEV2X post-processes 3D object bounding boxes collected by each agent to determine which objects are \textit{valid} (i.e., observed by at least one sensor of the agent; this labeling is useful for training object detection algorithms, as it can filter out fully occluded objects \cite{caesar2020nuscenes}), and semantic occupancy voxel grids to fill the interior of labeled hollow shells that represent each object. SimBEV2X does so using custom PyTorch \cite{paszke2019pytorch} CUDA extensions for increased efficiency.
    
    For 3D object bounding boxes, SimBEV2X first calculates the number of lidar and radar points that fall within each bounding box collected by each agent in each frame. Then, SimBEV2X combines this information with object IDs obtained from instance segmentation images (if collected by the agent) to determine which objects are \textit{valid}, and assigns a detection difficulty level (\textit{easy}, \textit{moderate}, or \textit{hard}) to each object based on that object's distance from the agent and the number of points located inside its bounding box. Next, for each frame, SimBEV2X uses object IDs to combine all bounding boxes collected by each agent into a unified list, which can be useful for training or evaluating motion tracking and prediction algorithms.
    
    For 3D semantic occupancy voxel grids, because object labels form hollow shells (i.e., the interior voxels are unlabeled because the labels are assigned when voxels overlap with object meshes, which have an empty interior), SimBEV2X assigns the appropriate label to the unlabeled interior voxels for each voxel grid. For each unlabeled voxel, SimBEV2X checks the first labeled voxel along each of the six principal directions of the voxel grid, starting from that unlabeled voxel. If all of those have the same label, then the unlabeled voxel is assigned that label as well.

    \begin{table}[t]
        \centering
        \setlength{\abovecaptionskip}{0 pt}
        \caption{Semantic Occupancy Class Priority} \label{table:semantic-priority}
        \begin{tabular}{c c}
            \toprule
            \textbf{Priority} & \textbf{Class} \\
            \toprule
            8 & road line \\
            7 & road, sidewalk \\
            6 & pedestrian, rider \\
            5 & motorcycle, bicycle \\
            4 & car \\
            3 & truck, bus \\
            2 & traffic light, traffic sign, traffic cone, barrier \\
            1 & all other classes \\
            0 & unlabeled \\
            \bottomrule
        \end{tabular}
    \end{table}

    \section{The SimBEV2X Dataset} \label{sec:simbev2x-dataset}

    To showcase SimBEV2X, we used it to create the SimBEV2X dataset, the largest V2X perception dataset to date and an order of magnitude larger than the existing V2X datasets. It is a collection of 258 scenes spread across all 11 CARLA maps according to \autoref{table:simbev2x-split}. We reserved Town13 for the \textit{validation} and \textit{test} sets, because it shares many common features with Town12 (both are CARLA's large maps) but uses different building styles, textures, and vegetation, allowing us to evaluate the generalization performance of the trained models and expose overfitting.

    \begin{table}[t]
        \centering
        \footnotesize
        \setlength{\abovecaptionskip}{0 pt}
        \caption{The SimBEV2X Dataset} \label{table:simbev2x-split}
        \begin{tabular}{c c c c}
            \toprule
            \textbf{Map} & \textbf{Train} & \textbf{Validation} & \textbf{Test} \\
            \toprule
            Town01 & 8 & 1 & 2 \\
            Town02 & 8 & 1 & 2 \\
            Town03 & 16 & 2 & 4 \\
            Town04 & 16 & 2 & 4 \\
            Town05 & 16 & 2 & 4 \\
            Town06 & 16 & 2 & 4 \\
            Town07 & 16 & 2 & 4 \\
            Town10HD & 16 & 2 & 4 \\
            Town12 & 32 & 4 & 8 \\
            Town13 & 0 & 8 & 16 \\
            Town15 & 24 & 4 & 8 \\
            \toprule
            \textbf{Total} & 168 & 30 & 60 \\
            \bottomrule
        \end{tabular}
    \end{table}
    \begin{table}[t]
        \centering
        \footnotesize
        \setlength{\abovecaptionskip}{0 pt}
        \setlength{\tabcolsep}{4.8 pt}
        \caption{Sensor Properties.} \label{table:sensor-properties}
        \begin{tabular}{l l}
            \toprule
            \textbf{Sensor type} & \textbf{Properties} \\
            \toprule
            RGB camera & 1600$\times$900 resolution, 80 deg FoV, $f/1.8$ \\
            \midrule
            All other cameras & 1600$\times$900 resolution, 80 deg FoV \\
            \midrule
            \multirow{4}*{Lidar} & 128 channels, 120.0 m range, 20.0 Hz rotation \\
             & frequency, 5,242,880 points per second, -30.67 deg\\
             & to 10.67 deg vertical FoV, 14\% general drop-off\\
             & rate, 3 cm radial noise std \\
            \midrule
            \multirow{4}*{RSU Lidar} & 128 channels, 120.0 m range, 20.0 Hz rotation \\
             & frequency, 2,621,440 points per second, -45.0 deg\\
             & to 0.0 deg vertical FoV, 180 deg horizontal FoV,\\
             & 14\% general drop-off rate, 3 cm radial noise std \\
            \midrule
            \multirow{2}*{Radar} & 120.0 m range, 100 deg horizontal FoV, 20 deg \\
             & vertical FoV, 40,000 points per second \\
            \midrule
            \multirow{2}*{RSU Radar} & 120.0 m range, 100 deg horizontal FoV, 20 deg \\
             & vertical FoV, 48,000 points per second \\
            \midrule
            \multirow{2}*{GNSS} & \{4e-2 m, 4e-7 deg, 4e-7 deg\} noise std for \\
             & \{altitude, latitude, longitude\} \\
            \midrule
            \multirow{3}*{IMU} & 1.7e-4 rad/s gyroscope bias, \\
             & \{1.7e-4 m/s$^ {2}$, 5.6e-6 rad/s\} noise std \\
             & for \{accelerometer, gyroscope\} \\
            \bottomrule
        \end{tabular}
    \end{table}
    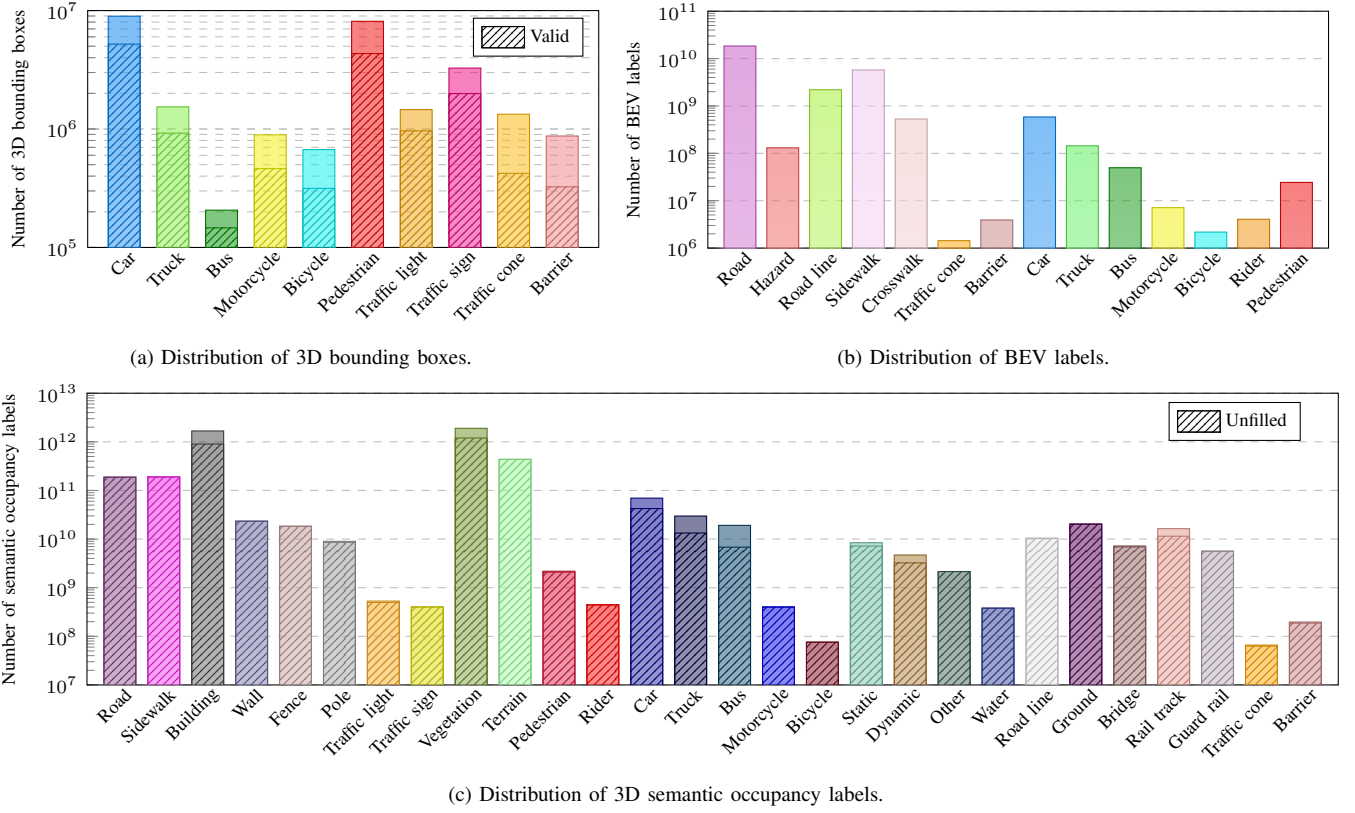
\begin{figure*}[!t]
    \centering
        \subfloat[Distribution of 3D bounding boxes.]{
        \begin{tikzpicture}
            \begin{axis}[
                title = {},
                ylabel = {Number of 3D bounding boxes}, ylabel near ticks,
                xmin = 0.25, xmax = 10.75,
                ymin = 100000, ymax = 10000000, ymode = log,
                xtick = {1, ..., 10}, xtick style = {draw = none},
                xticklabels = {Car, Truck, Bus, Motorcycle, Bicycle, Pedestrian, Traffic light, Traffic sign, Traffic cone, Barrier}, xticklabel style = {rotate = 45, anchor = east, xshift = 4 pt, yshift = -4 pt},
                label style = {font = \scriptsize},
                tick pos = left, tick label style = {font = \scriptsize},
                legend pos = north east, legend style = {font = \scriptsize},
                ymajorgrids = true, yminorgrids = true,
                grid style = dashed,
                width = 0.46\textwidth, height = 0.26\textwidth
            ]
                \addlegendimage{custom hatch legend}
                \addlegendentry{Valid}
                \addplot[glassybar = Car] coordinates {(1, 8969282)};
                \addplot[hatchbar = {north east lines}{Car}] coordinates {(1, 5209929)};
                \addplot[glassybar = Truck] coordinates {(2, 1539699)};
                \addplot[hatchbar = {north east lines}{Truck}] coordinates {(2, 923630)};
                \addplot[glassybar = Bus] coordinates {(3, 206571)};
                \addplot[hatchbar = {north east lines}{Bus}] coordinates {(3, 146310)};
                \addplot[glassybar = Motorcycle] coordinates {(4, 893555)};
                \addplot[hatchbar = {north east lines}{Motorcycle}] coordinates {(4, 462674)};
                \addplot[glassybar = Bicycle] coordinates {(5, 670171)};
                \addplot[hatchbar = {north east lines}{Bicycle}] coordinates {(5, 315425)};
                \addplot[glassybar = Pedestrian] coordinates {(6, 8123729)};
                \addplot[hatchbar = {north east lines}{Pedestrian}] coordinates {(6, 4337994)};
                \addplot[glassybar = TrafficLight] coordinates {(7, 1458201)};
                \addplot[hatchbar = {north east lines}{TrafficLight}] coordinates {(7, 964036)};
                \addplot[glassybar = TrafficSign] coordinates {(8, 3268554)};
                \addplot[hatchbar = {north east lines}{TrafficSign}] coordinates {(8, 1991774)};
                \addplot[glassybar = TrafficCone] coordinates {(9, 1333855)};
                \addplot[hatchbar = {north east lines}{TrafficCone}] coordinates {(9, 422427)};
                \addplot[glassybar = Barrier] coordinates {(10, 873449)};
                \addplot[hatchbar = {north east lines}{Barrier}] coordinates {(10, 324883)};
            \end{axis}
        \end{tikzpicture}
        \label{subfloat:bbox-count}
        }
        \subfloat[Distribution of BEV labels.]{
        \begin{tikzpicture}
            \begin{axis}[
                title = {},
                ylabel = {Number of BEV labels}, ylabel near ticks,
                xmin = 0.25, xmax = 14.75,
                ymin = 1000000, ymax = 100000000000, ymode = log,
                xtick = {1, ..., 14}, xtick style = {draw = none},
                xticklabels = {Road, Hazard, Road line, Sidewalk, Crosswalk, Traffic cone, Barrier, Car, Truck, Bus, Motorcycle, Bicycle, Rider, Pedestrian}, xticklabel style = {rotate = 45, anchor = east, xshift = 4 pt, yshift = -4 pt},
                ytick = {1000000, 10000000, 100000000, 1000000000, 10000000000, 100000000000},
                label style = {font = \scriptsize},
                tick pos = left, tick label style = {font = \scriptsize},
                ymajorgrids = true, grid style = dashed,
                width = 0.54\textwidth, height = 0.26\textwidth
            ]
                \addplot[glassybar = MapRoad] coordinates {(1, 18423410070)};
                \addplot[glassybar = MapHazard] coordinates {(2, 130789092)};
                \addplot[glassybar = MapRoadLine] coordinates {(3, 2203938665)};
                \addplot[glassybar = MapSidewalk] coordinates {(4, 5764613682)};
                \addplot[glassybar = MapCrosswalk] coordinates {(5, 530387859)};
                \addplot[glassybar = MapTrafficCone] coordinates {(6, 1435393)};
                \addplot[glassybar = MapBarrier] coordinates {(7, 3920504)};
                \addplot[glassybar = MapCar] coordinates {(8, 586268097)};
                \addplot[glassybar = MapTruck] coordinates {(9, 144110972)};
                \addplot[glassybar = MapBus] coordinates {(10, 49656143)};
                \addplot[glassybar = MapMotorcycle] coordinates {(11, 7145620)};
                \addplot[glassybar = MapBicycle] coordinates {(12, 2177515)};
                \addplot[glassybar = MapRider] coordinates {(13, 4059365)};
                \addplot[glassybar = MapPedestrian] coordinates {(14, 24345256)};
            \end{axis}
        \end{tikzpicture}
        \label{subfloat:bev-count}
        } \\\vspace*{-8 pt}
        \subfloat[Distribution of 3D semantic occupancy labels.]{
        \begin{tikzpicture}
            \begin{axis}[
                title = {},
                ylabel = {Number of semantic occupancy labels}, ylabel near ticks,
                xmin = 0.25, xmax = 28.75,
                ymin = 10000000, ymax = 10000000000000, ymode = log,
                xtick = {1, ..., 28}, xtick style = {draw = none},
                xticklabels = {Road, Sidewalk, Building, Wall, Fence, Pole, Traffic light, Traffic sign, Vegetation, Terrain, Pedestrian, Rider, Car, Truck, Bus, Motorcycle, Bicycle, Static, Dynamic, Other, Water, Road line, Ground, Bridge, Rail track, Guard rail, Traffic cone, Barrier},  xticklabel style = {rotate = 45, anchor = east, xshift = 4 pt, yshift = -4 pt},
                ytick = {10000000, 100000000, 1000000000, 10000000000, 100000000000, 1000000000000, 10000000000000},
                label style = {font = \scriptsize},
                tick pos = left, tick label style = {font = \scriptsize},
                legend pos = north east, legend style = {font = \scriptsize},
                ymajorgrids = true, grid style = dashed,
                width = \textwidth, height = 0.3\textwidth
            ]
                \addlegendimage{custom hatch legend}
                \addlegendentry{Unfilled}
                \addplot[glassybar = OccRoad] coordinates {(1, 187936233179)};
                \addplot[hatchbar = {north east lines}{OccRoad}] coordinates {(1, 187936233179)};
                \addplot[glassybar = OccSidewalk] coordinates {(2, 189922307774)};
                \addplot[hatchbar = {north east lines}{OccSidewalk}] coordinates {(2, 189922307774)};
                \addplot[glassybar = OccBuilding] coordinates {(3, 1674330702776)};
                \addplot[hatchbar = {north east lines}{OccBuilding}] coordinates {(3, 900817475055)};
                \addplot[glassybar = OccWall] coordinates {(4, 23378189499)};
                \addplot[hatchbar = {north east lines}{OccWall}] coordinates {(4, 23378189499)};
                \addplot[glassybar = OccFence] coordinates {(5, 18363831129)};
                \addplot[hatchbar = {north east lines}{OccFence}] coordinates {(5, 18363831129)};
                \addplot[glassybar = OccPole] coordinates {(6, 8989939222)};
                \addplot[hatchbar = {north east lines}{OccPole}] coordinates {(6, 8579662196)};
                \addplot[glassybar = OccTrafficLight] coordinates {(7, 531133151)};
                \addplot[hatchbar = {north east lines}{OccTrafficLight}] coordinates {(7, 498884251)};
                \addplot[glassybar = OccTrafficSign] coordinates {(8, 402163184)};
                \addplot[hatchbar = {north east lines}{OccTrafficSign}] coordinates {(8, 400975040)};
                \addplot[glassybar = OccVegetation] coordinates {(9, 1898647383174)};
                \addplot[hatchbar = {north east lines}{OccVegetation}] coordinates {(9, 1198452609539)};
                \addplot[glassybar = OccTerrain] coordinates {(10, 436551954377)};
                \addplot[hatchbar = {north east lines}{OccTerrain}] coordinates {(10, 436551954377)};
                \addplot[glassybar = OccPedestrian] coordinates {(11, 2183767111)};
                \addplot[hatchbar = {north east lines}{OccPedestrian}] coordinates {(11, 2094463408)};
                \addplot[glassybar = OccRider] coordinates {(12, 443567509)};
                \addplot[hatchbar = {north east lines}{OccRider}] coordinates {(12, 443130500)};
                \addplot[glassybar = OccCar] coordinates {(13, 69524669086)};
                \addplot[hatchbar = {north east lines}{OccCar}] coordinates {(13, 42413913362)};
                \addplot[glassybar = OccTruck] coordinates {(14, 29759074139)};
                \addplot[hatchbar = {north east lines}{OccTruck}] coordinates {(14, 13310215313)};
                \addplot[glassybar = OccBus] coordinates {(15, 19140621987)};
                \addplot[hatchbar = {north east lines}{OccBus}] coordinates {(15, 6826241522)};
                \addplot[glassybar = OccMotorcycle] coordinates {(16, 404215458)};
                \addplot[hatchbar = {north east lines}{OccMotorcycle}] coordinates {(16, 397784633)};
                \addplot[glassybar = OccBicycle] coordinates {(17, 75849624)};
                \addplot[hatchbar = {north east lines}{OccBicycle}] coordinates {(17, 75835868)};
                \addplot[glassybar = OccStatic] coordinates {(18, 8419060788)};
                \addplot[hatchbar = {north east lines}{OccStatic}] coordinates {(18, 7171988904)};
                \addplot[glassybar = OccDynamic] coordinates {(19, 4702667653)};
                \addplot[hatchbar = {north east lines}{OccDynamic}] coordinates {(19, 3235741524)};
                \addplot[glassybar = OccOther] coordinates {(20, 2150601495)};
                \addplot[hatchbar = {north east lines}{OccOther}] coordinates {(20, 2150601495)};
                \addplot[glassybar = OccWater] coordinates {(21, 379943469)};
                \addplot[hatchbar = {north east lines}{OccWater}] coordinates {(21, 379943469)};
                \addplot[glassybar = OccRoadLine] coordinates {(22, 10380085422)};
                \addplot[hatchbar = {north east lines}{OccRoadLine}] coordinates {(22, 10380085422)};
                \addplot[glassybar = OccGround] coordinates {(23, 20361839035)};
                \addplot[hatchbar = {north east lines}{OccGround}] coordinates {(23, 20361839035)};
                \addplot[glassybar = OccBridge] coordinates {(24, 7216751179)};
                \addplot[hatchbar = {north east lines}{OccBridge}] coordinates {(24, 6893309672)};
                \addplot[glassybar = OccRailTrack] coordinates {(25, 16379034314)};
                \addplot[hatchbar = {north east lines}{OccRailTrack}] coordinates {(25, 11447168465)};
                \addplot[glassybar = OccGuardRail] coordinates {(26, 5681437028)};
                \addplot[hatchbar = {north east lines}{OccGuardRail}] coordinates {(26, 5591909462)};
                \addplot[glassybar = OccTrafficCone] coordinates {(27, 65986418)};
                \addplot[hatchbar = {north east lines}{OccTrafficCone}] coordinates {(27, 62982599)};
                \addplot[glassybar = OccBarrier] coordinates {(28, 196681421)};
                \addplot[hatchbar = {north east lines}{OccBarrier}] coordinates {(28, 187344182)};
            \end{axis}
        \end{tikzpicture}
        \label{subfloat:occ-count}
        }
        \caption{Number of 3D object bounding boxes, BEV ground truth labels, and 3D semantic occupancy labels per class in the SimBEV2X dataset. The hatching indicates the number of valid object bounding boxes in (a) and unfilled voxels (before post-processing) in (c). Similar to \cite{zimmer2024tumtraf, xiang2024v2x}, cars and pedestrians are highly represented among the annotations. The overall distribution of the annotations is balanced and similar to real-world data.} \label{fig:gt-stats}
        \vspace*{-16 pt}
    \end{figure*}
    \begin{figure*}[!t]
        \centering
        \subfloat[Distribution of scene duration.]{
        \begin{tikzpicture}
            \begin{axis}[
                title = {},
                ylabel = {Scene}, ylabel near ticks, 
                xmin = 7, xmax = 33, 
                ymin = 0, ymax = 18, 
                xtick = {8, 10, ..., 32}, xtick style = {draw = none},
                ytick = {2, 4, ..., 18},
                label style = {font = \scriptsize}, 
                tick pos = left, tick label style = {font = \scriptsize}, 
                ymajorgrids = true, grid style = dashed, 
                width = 0.4\textwidth, height = 0.28\textwidth
            ]
                \addplot[glassybar = green, no marks, bar width = 4 pt] coordinates {
                    (8, 3)
                    (9, 15)
                    (10, 12)
                    (11, 9)
                    (12, 16)
                    (13, 8)
                    (14, 15)
                    (15, 9)
                    (16, 12)
                    (17, 11)
                    (18, 9)
                    (19, 7)
                    (20, 9)
                    (21, 10)
                    (22, 10)
                    (23, 5)
                    (24, 12)
                    (25, 17)
                    (26, 10)
                    (27, 15)
                    (28, 11)
                    (29, 8)
                    (30, 8)
                    (31, 10)
                    (32, 7)
                };
            \end{axis}
        \end{tikzpicture}
        \label{subfloat:scene-duration}
        } \hspace*{8 pt}
        \subfloat[Distribution of data collection vehicles and RSUs per scene.]{
        \begin{tikzpicture}
            \begin{axis}[
                title = {},
                ybar = {0 pt},
                ylabel = {Scene}, ylabel near ticks, 
                xmin = -0.25, xmax = 8.25, 
                ymin = 0, ymax = 180,
                xtick = {0, ..., 8}, xtick style = {draw = none},
                ytick = {20, 40, ..., 180},
                xticklabel style = {yshift = 4 pt},
                label style = {font = \scriptsize}, 
                tick pos = left, tick label style = {font = \scriptsize}, 
                legend style = {font = \scriptsize}, 
                ymajorgrids = true, grid style = dashed, 
                width = 0.26\textwidth, height = 0.28\textwidth
            ]
                \addplot[glassybar = blue, no marks, bar width = 4 pt, area legend] coordinates {
                    (2, 46)
                    (3, 32)
                    (4, 39)
                    (5, 40)
                    (6, 39)
                    (7, 26)
                    (8, 36)
                };
                \addplot[glassybar = red, no marks, bar width = 4 pt, area legend] coordinates {
                    (0, 162)
                    (1, 24)
                    (2, 25)
                    (3, 23)
                    (4, 24)
                };
                \legend{Vehicle, RSU}
            \end{axis}
        \end{tikzpicture}
        \label{subfloat:vehicles-rsus}
        } \hspace*{8 pt}
        \subfloat[Distribution of hazard areas per scene.]{
        \begin{tikzpicture}
            \begin{axis}[
                title = {},
                ybar = {0 pt},
                ylabel = {Scene}, ylabel near ticks, 
                xmin = -0.5, xmax = 10.5, 
                ymin = 0, ymax = 60, 
                xtick = {0, ..., 10}, xtick style = {draw = none},
                xticklabels = {0, 1, 2, 3, 4, 5, 6, 7, 8, 9, 10+}, xticklabel style = {yshift = 4 pt},
                ytick = {10, 20, ..., 60},
                label style = {font = \scriptsize}, 
                tick pos = left, tick label style = {font = \scriptsize}, 
                legend pos = north east, legend style = {font = \scriptsize}, 
                ymajorgrids = true, grid style = dashed, 
                width = 0.32\textwidth, height = 0.28\textwidth
            ]
                \addplot[glassybar = orange, no marks, bar width = 4 pt, area legend] coordinates {
                    (0, 36)
                    (1, 47)
                    (2, 48)
                    (3, 32)
                    (4, 41)
                    (5, 20)
                    (6, 15)
                    (7, 9)
                    (8, 4)
                    (9, 2)
                    (10, 4)
                };
                \addplot[glassybar = purple, no marks, bar width = 4 pt, area legend] coordinates {
                    (0, 14)
                    (1, 51)
                    (2, 51)
                    (3, 35)
                    (4, 38)
                    (5, 19)
                    (6, 14)
                    (7, 20)
                    (8, 7)
                    (9, 7)
                    (10, 2)
                };
                \legend{Accident, Road work}
            \end{axis}
        \end{tikzpicture}
        \label{subfloat:hazards}
        } \\\vspace*{-8 pt}
        \subfloat[Distribution of background vehicles and pedestrians per scene.]{
        \begin{tikzpicture}
            \begin{axis}[
                title = {},
                ybar = {0 pt},
                ylabel = {Scene}, ylabel near ticks, 
                xmin = 0, xmax = 420, 
                ymin = 0, ymax = 55, 
                xtick = {10, 50, ..., 410}, xtick style = {draw = none},
                xticklabels = {[0{,} 20), [40{,} 60), [80{,} 100), [120{,} 140), [160{,} 180), [200{,} 220), [240{,} 260), [280{,} 300), [320{,} 340), [360{,} 380), 400+}, xticklabel style = {rotate = 30, anchor = east, xshift = 6 pt, yshift = -2 pt},
                ytick = {10, 20, ..., 50},
                label style = {font = \scriptsize}, 
                tick pos = left, tick label style = {font = \scriptsize}, 
                legend pos = north east, legend style = {font = \scriptsize}, 
                ymajorgrids = true, grid style = dashed, 
                width = 0.48\textwidth, height = 0.28\textwidth
            ]
                \addplot[glassybar = Car, no marks, bar width = 4 pt, area legend] coordinates {
                    (10, 22)
                    (30, 24)
                    (50, 17)
                    (70, 23)
                    (90, 14)
                    (110, 17)
                    (130, 18)
                    (150, 14)
                    (170, 6)
                    (190, 11)
                    (210, 5)
                    (230, 11)
                    (250, 4)
                    (270, 5)
                    (290, 8)
                    (310, 7)
                    (330, 6)
                    (350, 3)
                    (370, 4)
                    (390, 3)
                    (410, 36)
                };
                \addplot[glassybar = Pedestrian, no marks, bar width = 4 pt, area legend] coordinates {
                    (10, 50)
                    (30, 13)
                    (50, 16)
                    (70, 19)
                    (90, 19)
                    (110, 15)
                    (130, 18)
                    (150, 19)
                    (170, 12)
                    (190, 16)
                    (210, 11)
                    (230, 14)
                    (250, 9)
                    (270, 6)
                    (290, 7)
                    (310, 5)
                    (330, 2)
                    (350, 2)
                    (370, 0)
                    (390, 2)
                    (410, 3)
                };
                \legend{Vehicles, Pedestrians}
            \end{axis}
        \end{tikzpicture}
        \label{subfloat:spawns}
        }
        \subfloat[Distribution of \textit{valid} objects per frame.]{
        \begin{tikzpicture}
            \begin{axis}[
                title = {},
                ylabel = {Frame}, ylabel near ticks,
                xmin = 0, xmax = 105,
                ymin = 1, ymax = 100000, ymode = log,
                xtick = {0, 10, ..., 100}, xtick style = {draw = none},
                xticklabels = {0, 10, ..., 90, 100+},
                ytick = {10, 100, 1000, 10000, 100000},
                label style = {font = \scriptsize},
                tick pos = left, tick label style = {font = \scriptsize},
                legend style = {font = \scriptsize, at = {(1.01, 0.536)}, anchor = west},
                ymajorgrids = true, grid style = dashed,
                width = 0.42\textwidth, height = 0.31\textwidth,
            ]
                \addplot[glassyarea = Car] coordinates {
                    (2.5, 2081)
                    (7.5, 2452)
                    (12.5, 6053)
                    (17.5, 8915)
                    (22.5, 8164)
                    (27.5, 9277)
                    (32.5, 8896)
                    (37.5, 8298)
                    (42.5, 6510)
                    (47.5, 4860)
                    (52.5, 3400)
                    (57.5, 3021)
                    (62.5, 3898)
                    (67.5, 4345)
                    (72.5, 2038)
                    (77.5, 2052)
                    (82.5, 1991)
                    (87.5, 1442)
                    (92.5, 1059)
                    (97.5, 1692)
                    (102.5, 11756)
                } \closedcycle;
                \addplot[glassyarea = Truck] coordinates {
                    (2.5, 35552)
                    (7.5, 30664)
                    (12.5, 15710)
                    (17.5, 9098)
                    (22.5, 4922)
                    (27.5, 3156)
                    (32.5, 1822)
                    (37.5, 704)
                    (42.5, 369)
                    (47.5, 142)
                    (52.5, 51)
                    (57.5, 10)
                } \closedcycle;
                \addplot[glassyarea = Bus] coordinates {
                    (2.5, 96743)
                    (7.5, 5318)
                    (12.5, 139)
                } \closedcycle;
                \addplot[glassyarea = Motorcycle] coordinates {
                    (2.5, 63009)
                    (7.5, 24832)
                    (12.5, 11378)
                    (17.5, 2884)
                    (22.5, 97)
                } \closedcycle;
                \addplot[glassyarea = Bicycle] coordinates {
                    (2.5, 77425)
                    (7.5, 19425)
                    (12.5, 5144)
                    (17.5, 206)
                } \closedcycle;
                \addplot[glassyarea = Pedestrian] coordinates {
                    (2.5, 25278)
                    (7.5, 8899)
                    (12.5, 8211)
                    (17.5, 6407)
                    (22.5, 3784)
                    (27.5, 4392)
                    (32.5, 4194)
                    (37.5, 3918)
                    (42.5, 3378)
                    (47.5, 2748)
                    (52.5, 2217)
                    (57.5, 2085)
                    (62.5, 1627)
                    (67.5, 2602)
                    (72.5, 2773)
                    (77.5, 1844)
                    (82.5, 1771)
                    (87.5, 1095)
                    (92.5, 1261)
                    (97.5, 2062)
                    (102.5, 11654)
                } \closedcycle;
                \addplot[glassyarea = TrafficLight] coordinates {
                    (2.5, 42142)
                    (7.5, 18139)
                    (12.5, 15963)
                    (17.5, 12176)
                    (22.5, 6265)
                    (27.5, 2370)
                    (32.5, 1938)
                    (37.5, 1659)
                    (42.5, 776)
                    (47.5, 768)
                    (52.5, 4)
                } \closedcycle;
                \addplot[glassyarea = TrafficSign] coordinates {
                    (2.5, 20909)
                    (7.5, 19224)
                    (12.5, 17588)
                    (17.5, 7908)
                    (22.5, 6713)
                    (27.5, 4743)
                    (32.5, 3332)
                    (37.5, 3497)
                    (42.5, 3282)
                    (47.5, 4129)
                    (52.5, 3994)
                    (57.5, 3011)
                    (62.5, 2237)
                    (67.5, 1053)
                    (72.5, 343)
                    (77.5, 234)
                    (82.5, 3)
                } \closedcycle;
                \addplot[glassyarea = TrafficCone] coordinates {
                    (2.5, 72460)
                    (7.5, 16394)
                    (12.5, 6448)
                    (17.5, 3630)
                    (22.5, 889)
                    (27.5, 1176)
                    (32.5, 956)
                    (37.5, 165)
                    (42.5, 82)
                } \closedcycle;
                \addplot[glassyarea = Barrier] coordinates {
                    (2.5, 78417)
                    (7.5, 12543)
                    (12.5, 6701)
                    (17.5, 3935)
                    (22.5, 451)
                    (27.5, 134)
                    (32.5, 19)
                } \closedcycle;
                \legend{Car, Truck, Bus, Motorcycle, Bicycle, Pedestrian, Traffic light, Traffic sign, Traffic cone, Barrier}
            \end{axis}
        \end{tikzpicture}
        \label{subfloat:valid-per-frame}
        }
        \caption{Distribution of various parameters across the scenes and frames of the SimBEV2X dataset. While scene duration is chosen randomly from a uniform distribution, it is rounded to the nearest integer, which could be the cause of the uneven distribution seen in (a). Because RSUs are only placed at signalized intersections and in many scenes (e.g., those involving highways) RSUs cannot be spawned, a large number of scenes have no RSUs, as shown in (b). As shown in (e), cars and pedestrians are highly represented across the dataset.} \label{fig:scene-stats}
    \end{figure*}
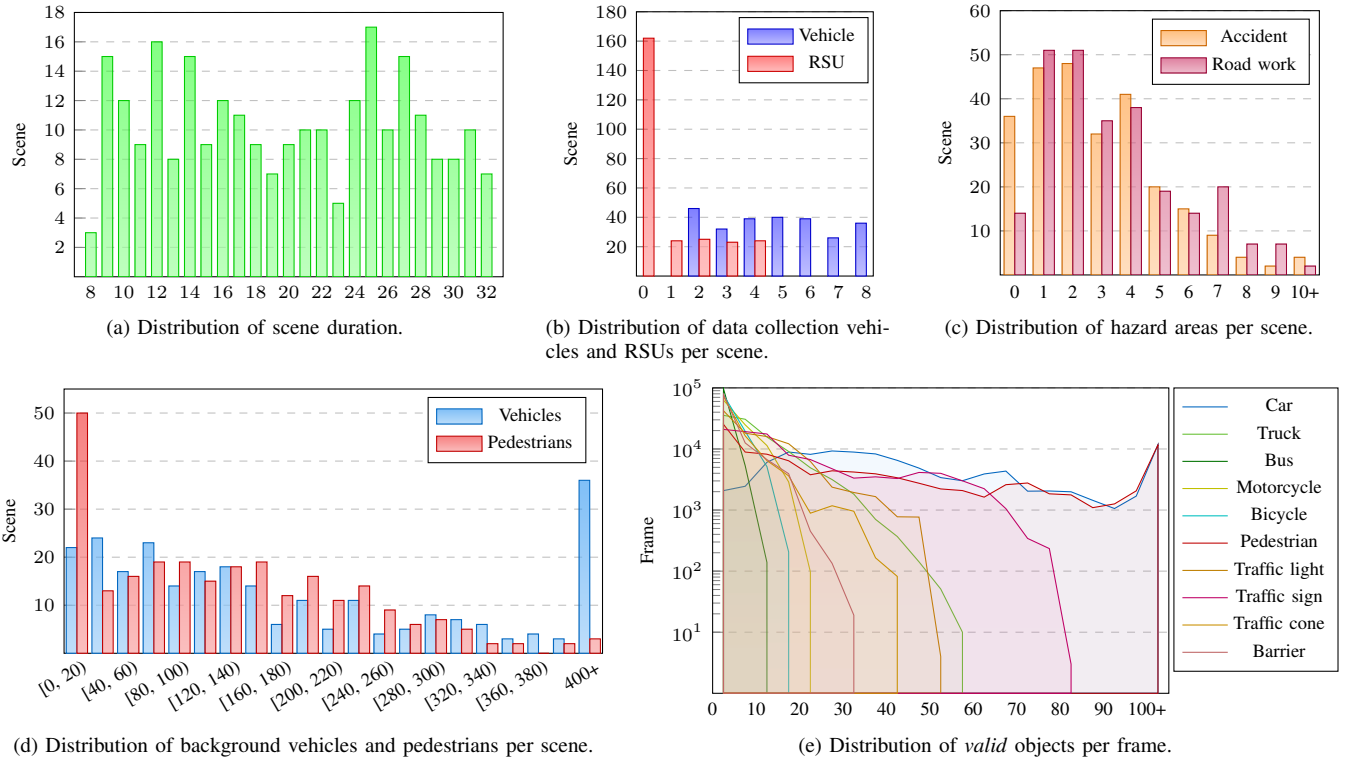
    \begin{figure*}[!t]
        \centering
        \subfloat[Road]{\includegraphics[width=0.13\textwidth]{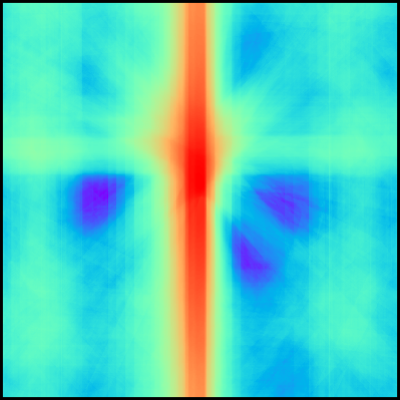} \label{subfloat:road}}
        \subfloat[Hazard]{\includegraphics[width=0.13\textwidth]{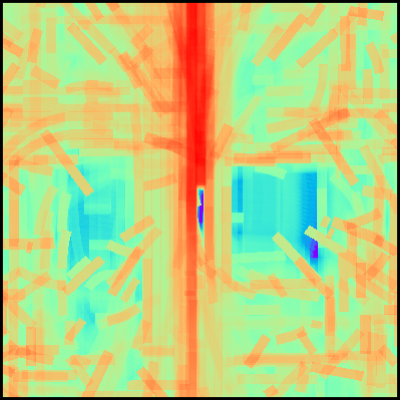} \label{subfloat:hazard}}
        \subfloat[Road line]{\includegraphics[width=0.13\textwidth]{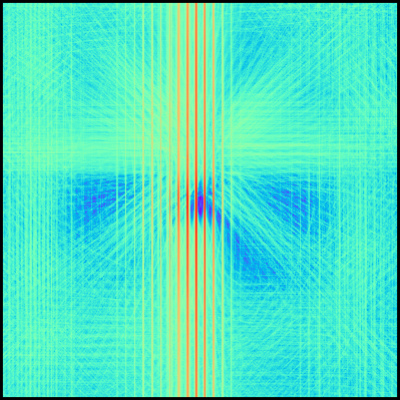} \label{subfloat:road-line}}
        \subfloat[Sidewalk]{\includegraphics[width=0.13\textwidth]{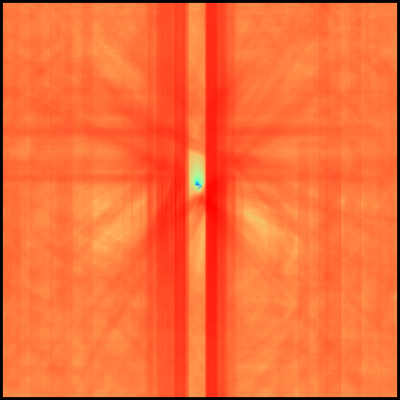} \label{subfloat:sidewalk}}
        \subfloat[Crosswalk]{\includegraphics[width=0.13\textwidth]{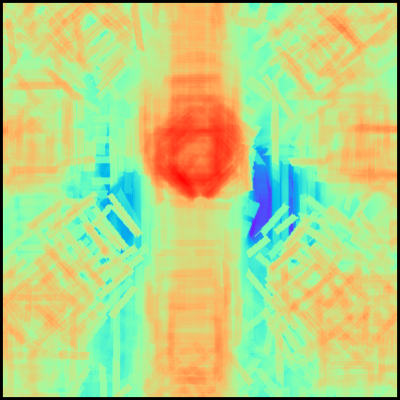} \label{subfloat:crosswalk}}
        \subfloat[Traffic cone]{\includegraphics[width=0.13\textwidth]{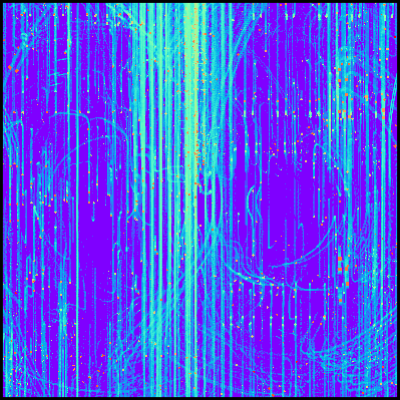} \label{subfloat:traffic-cone}}
        \subfloat[Barrier]{\includegraphics[width=0.13\textwidth]{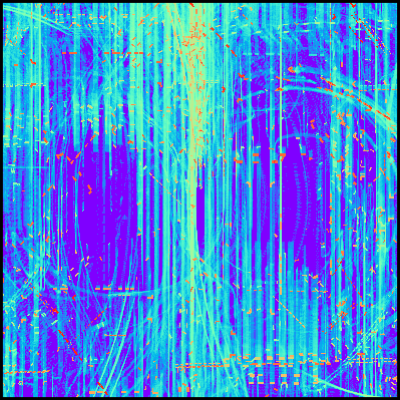} \label{subfloat:barrier}} \\\vspace*{-8 pt}
        \subfloat[Car]{\includegraphics[width=0.13\textwidth]{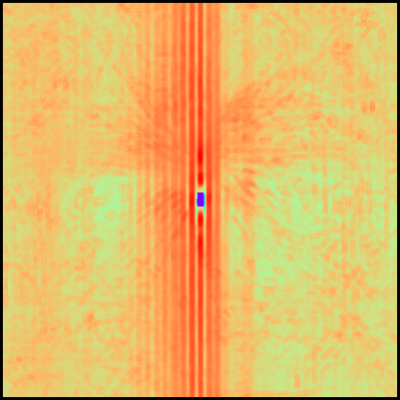} \label{subfloat:car}}
        \subfloat[Truck]{\includegraphics[width=0.13\textwidth]{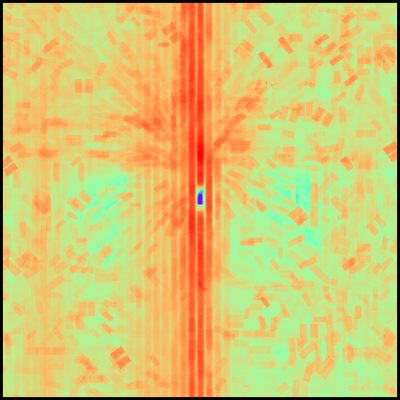} \label{subfloat:truck}}
        \subfloat[Bus]{\includegraphics[width=0.13\textwidth]{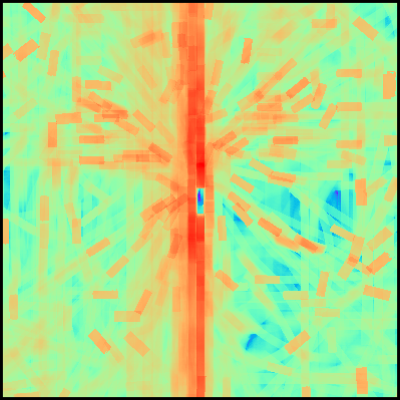} \label{subfloat:bus}}
        \subfloat[Motorcycle]{\includegraphics[width=0.13\textwidth]{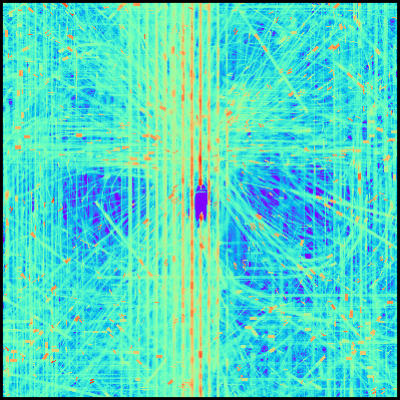} \label{subfloat:motorcycle}}
        \subfloat[Bicycle]{\includegraphics[width=0.13\textwidth]{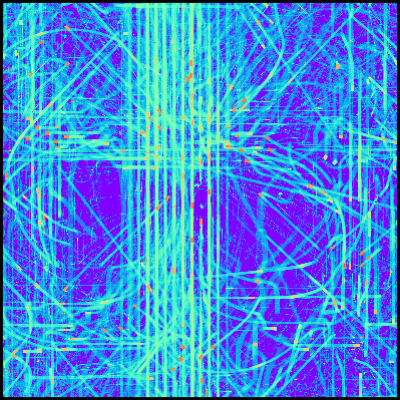} \label{subfloat:bicycle}}
        \subfloat[Rider]{\includegraphics[width=0.13\textwidth]{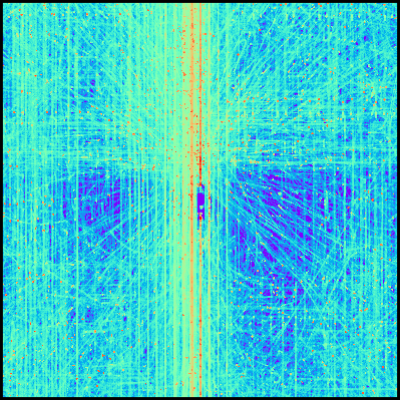} \label{subfloat:rider}}
        \subfloat[Pedestrian]{\includegraphics[width=0.13\textwidth]{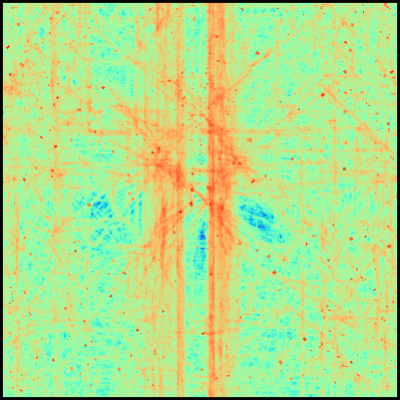} \label{subfloat:pedestrian}}
        \caption{Logarithmic heat map of BEV labels for each class aggregated across the SimBEV2X dataset. The biggest concentration of labels is along the path of travel of data collection vehicles.} \label{fig:bev-heatmap}
        \vspace*{-16 pt}
    \end{figure*}
    \begin{figure*}[!t]
        \centering
        \subfloat[Cloud cover]{
        \begin{tikzpicture}
            \begin{axis}[
                confusion matrix,
                xlabel = {Final weather},
                ylabel = {Initial weather},
                xtick = {0, 1, 2, 3},
                ytick = {0, 1, 2, 3},
                xticklabels = {none, low, moderate, heavy},
                yticklabels = {none, low, moderate, heavy},
                width = 0.248\textwidth, height = 0.248\textwidth,
            ]
                \addplot[matrix plot, mesh/cols = 4, point meta = explicit] coordinates {
                    (0,0) [0.0233] (1,0) [0.0581] (2,0) [0.0426] (3,0) [0.0581]
                    (0,1) [0.0504] (1,1) [0.0930] (2,1) [0.0853] (3,1) [0.0465]
                    (0,2) [0.0426] (1,2) [0.1008] (2,2) [0.0698] (3,2) [0.0736]
                    (0,3) [0.0310] (1,3) [0.0543] (2,3) [0.0853] (3,3) [0.0853]
                };
            \end{axis}
        \end{tikzpicture}
        \label{subfloat:cloudiness}
        }
        \subfloat[Rain intensity]{
        \begin{tikzpicture}
            \begin{axis}[
                confusion matrix,
                xlabel = {Final weather},
                xtick = {0, 1, 2, 3},
                ytick = {0, 1, 2, 3},
                xticklabels = {none, low, moderate, heavy},
                yticklabels = {none, low, moderate, heavy},
                width = 0.248\textwidth, height = 0.248\textwidth,
            ]
                \addplot[matrix plot, mesh/cols = 4, point meta = explicit] coordinates {
                    (0,0) [0.2481] (1,0) [0.0233] (2,0) [0.1318] (3,0) [0.0775]
                    (0,1) [0.0388] (1,1) [0.0155] (2,1) [0.0271] (3,1) [0.0194]
                    (0,2) [0.1279] (1,2) [0.0388] (2,2) [0.0581] (3,2) [0.0543]
                    (0,3) [0.0543] (1,3) [0.0116] (2,3) [0.0465] (3,3) [0.0271]
                };
            \end{axis}
        \end{tikzpicture}
        \label{subfloat:precipitation}
        }
        \subfloat[Fog density]{
        \begin{tikzpicture}
            \begin{axis}[
                confusion matrix,
                xlabel = {Final weather},
                xtick = {0, 1, 2, 3},
                ytick = {0, 1, 2, 3},
                xticklabels = {none, low, moderate, heavy},
                yticklabels = {none, low, moderate, heavy},
                width = 0.248\textwidth, height = 0.248\textwidth,
            ]
                \addplot[matrix plot, mesh/cols = 4, point meta = explicit] coordinates {
                    (0,0) [0.2209] (1,0) [0.1047] (2,0) [0.0736] (3,0) [0.0543]
                    (0,1) [0.1279] (1,1) [0.0620] (2,1) [0.0504] (3,1) [0.0271]
                    (0,2) [0.0659] (1,2) [0.0349] (2,2) [0.0233] (3,2) [0.0310]
                    (0,3) [0.0465] (1,3) [0.0388] (2,3) [0.0233] (3,3) [0.0155]
                };
            \end{axis}
        \end{tikzpicture}
        \label{subfloat:fog}
        }
        \subfloat[Sun altitude angle]{
        \begin{tikzpicture}
            \begin{axis}[
                confusion matrix,
                xlabel = {Final weather},
                xtick = {0, 1, 2},
                ytick = {0, 1, 2},
                xticklabels = {night, dawn/dusk, day},
                yticklabels = {night, dawn/dusk, day},
                width = 0.248\textwidth, height = 0.248\textwidth,
            ]
                \addplot[matrix plot, mesh/cols = 3, point meta = explicit] coordinates {
                    (0,0) [0.1008] (1,0) [0.0116] (2,0) [0.1822]
                    (0,1) [0.0194] (1,1) [0.0000] (2,1) [0.0349]
                    (0,2) [0.1938] (1,2) [0.0465] (2,2) [0.4109]
                };
            \end{axis}
        \end{tikzpicture}
        \label{subfloat:sun-angle}
        }
        \caption{Transition matrices over the SimBEV2X dataset for select weather parameters. For (a) through (c), \textit{none} is defined as $<$10\%, \textit{low} as 10-40\%, \textit{moderate} as 40-70\%, and \textit{heavy} as $>$70\%. For (d), \textit{night} is defined as $<$0$^ {\circ}$, \textit{dawn/dusk} as 0-6$^ {\circ}$, and \textit{day} as $>$6$^ {\circ}$. As (a) shows, scenes with the highest prevalence are those that start or end with \textit{low}, \textit{moderate}, or \textit{heavy} cloud cover (less so those transitioning from \textit{low} to \textit{heavy} or vice versa). (b) shows that around a quarter of the scenes do not involve any rain, while a majority of the rest start or end with a moderate to heavy amount of rain. (c) shows that scenes become less prevalent as the starting/ending fog density increases. Finally, (d) shows that around half of the scenes start and end in the same lighting condition (\textit{day} to \textit{day} or \textit{night} to \textit{night}) while the other half transition from one lighting condition to another.}
        \label{fig:weather-transition}
        \vspace*{-8 pt}
    \end{figure*}
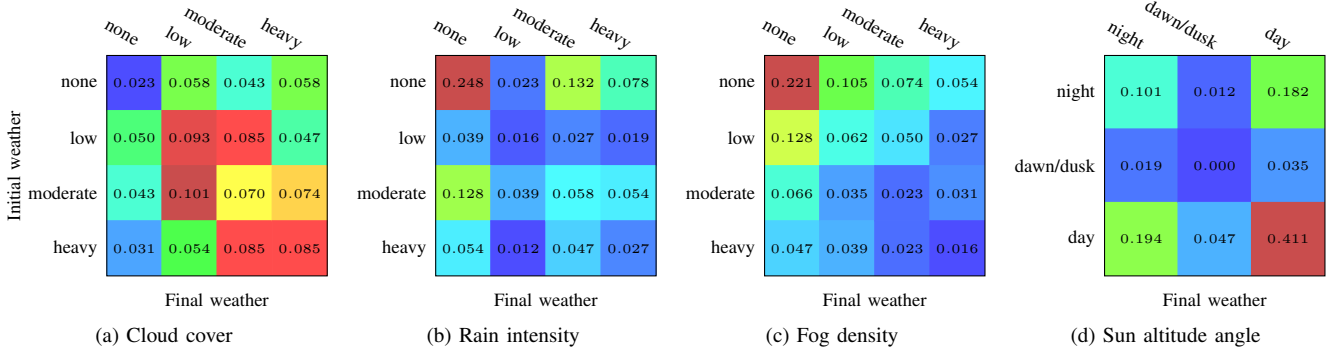

    We utilized SimBEV2X to its fullest extent to create the SimBEV2X dataset. We configured it to generate a diverse set of unique scenarios and collected data from various sensors, with the sensor configuration used listed in \autoref{table:sensor-properties}. Our sensor setup was mostly inspired by \cite{caesar2020nuscenes} (for example, the 1600$\times$900 image resolution, the arrangement of the cameras, and the lidar vertical FoV), though there are a few differences. Both lidars in our setup have 128 channels instead of 32 to collect a much denser point cloud. For GNSS and IMU, we used the bias and noise standard deviation values of the GNSS/INS module that is used in X-CAR, our experimental connected autonomous driving platform \cite{mehr2022x}. Alongside 3D object bounding boxes, the dataset includes $400\times400$ BEV ground truths for each agent in each frame, where each cell corresponds to a 0.4 m$\times$0.4 m square on the ground (160 m$\times$160 m grid). The dataset also includes $1280 \times 1280 \times 80$ semantic occupancy voxel grids for each agent in each frame, where each voxel corresponds to a cube with 10 cm sides (128 m$\times$128 m$\times$8 m (7 m above the origin of the grid, 1 m below).
    
    The SimBEV2X dataset was collected using a single Nvidia GeForce RTX 4090 graphics card over a 44-day period and is approximately 4.0 TB in size. Data were collected at a rate of 20 frames per simulation second, with each scene lasting between 8 and 32 seconds and containing between 2 and 8 data collection vehicles and 0 to 4 RSUs. In total, the SimBEV2X dataset contains 102,200 annotated frames, 27,337,066 3D object bounding boxes (15,099,082 of which are \textit{valid}), 27,866,258,233 BEV ground truth labels, and 4,636,519,690,601 3D semantic occupancy ground truth labels. Further statistics from the SimBEV2X dataset are available in \autoref{fig:gt-stats} to \autoref{fig:weather-transition} and in Appendix A.II.

    \section{Evaluation and Analysis} \label{sec:eval}

    We benchmark the SimBEV2X dataset using the two most common cooperative perception tasks: cooperative 3D object detection and cooperative BEV segmentation.
    
    \subsection{Tasks and Metrics} \label{subsec:eval-metrics}
    
    BEV segmentation is evaluated using the intersection over union (IoU) metric, where for each class, a prediction is considered positive if its probability (score) is above a certain threshold (commonly 0.5). For 3D object detection, our metrics are inspired by \cite{caesar2020nuscenes}, with average precision (AP) as the primary metric. Similar to \cite{caesar2020nuscenes}, a prediction and a ground truth bounding box are matched if the distance between their centers is below a certain threshold. As \cite{caesar2020nuscenes} points out, if bounding boxes were to be matched based on IoU, small translation errors for small objects (such as pedestrians) would result in low or even zero IoU, making performance comparison of camera-only models that tend to have large localization errors difficult.
    To prevent models from generating excessive low-confidence predictions that degrade AP, we impose class-specific limits on the number of predicted bounding boxes. Specifically, we retain only the highest-confidence predictions, capped at 20 for \textit{truck}, \textit{bus}, \textit{motorcycle}, \textit{bicycle}, and \textit{barrier}; 40 for \textit{traffic light} and \textit{traffic cone}; 60 for \textit{traffic sign}; and 100 for \textit{pedestrian} and \textit{car}.
    
    We calculate AP as the area under the precision-recall curve for precision over 10\%. Points on the curve where precision is less than 10\% are removed to minimize the impact of noise. Similar to \cite{caesar2020nuscenes}, we use the $\mathbb{T} = \{0.5, 1, 2, 4\}$ m distance thresholds to match bounding boxes, and define mAP as the mean AP over all classes and all matching thresholds:
    \begin{equation} \label{eq:map}
        \mathrm{mAP} = \frac{1}{|\mathbb{T}||\mathbb{C}|}\sum_{t \in \mathbb{T}, c \in \mathbb{C}}\mathrm{AP}_{t, c}.
    \end{equation}
    
    Similar to \cite{caesar2020nuscenes}, we measure a set of True Positive (TP) metrics for each pair of matched bounding boxes: Average Translation Error (ATE), which is the Euclidean distance (in m) between bounding box centers; Average Orientation Error (AOE), which is the smallest yaw angle difference (in rad) between the two bounding boxes; Average Scale Error (ASE), which is equal to one minus the 3D IoU of the two bounding boxes after aligning their orientation and translation; and Average Velocity Error (AVE), which is the L2 norm of the difference in bounding box velocities (in m/s). The mean TP value (mTP) for each metric is computed by averaging that metric over all classes and thresholds:
    \begin{equation} \label{eq:mtp}
        \mathrm{mTP} = \frac{1}{|\mathbb{T}||\mathbb{C}|}\sum_{t \in \mathbb{T}, c \in \mathbb{C}}\mathrm{TP}_{t, c}.
    \end{equation}
    
    Finally, similar to \cite{caesar2020nuscenes}, we define the SimBEV2X Detection Score (SDS) as:
    \begin{equation} \label{eq:sds}
        \mathrm{SDS} = \frac{1}{8}\Big(4\,\mathrm{mAP} + \sum_{\mathrm{mTP} \in \mathbb{TP}}(1-\min(1, \mathrm{mTP}))\Big).
    \end{equation}

    \subsection{Models} \label{subsec:models}

    \begin{figure*}[!t]
        \centering
        \includegraphics[width=\textwidth]{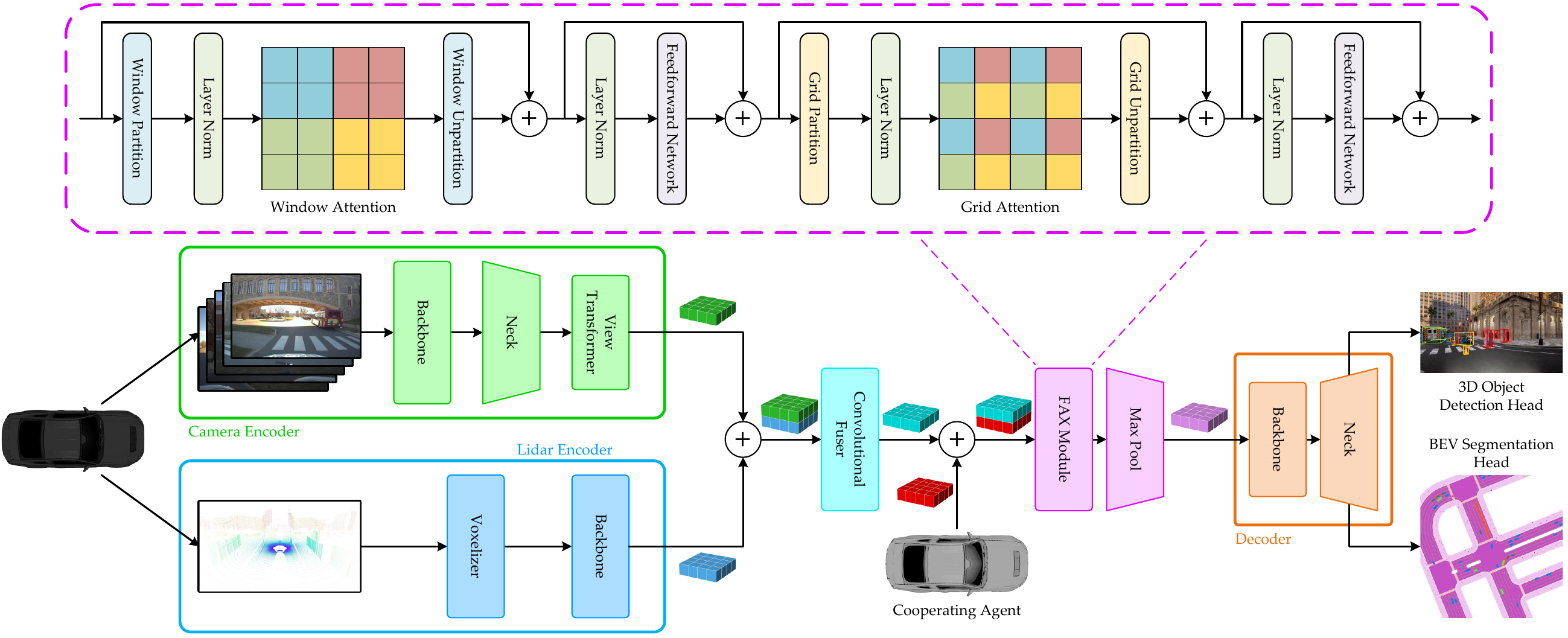}
        \setlength{\abovecaptionskip}{-12 pt}
        \caption{CoBEVFusion's architecture builds on CoopDet3D's by augmenting multi-agent BEV features using a fused axial attention (FAX) module that performs local window-based attention followed by sparse global attention. After element-wise max-pooling is performed, BEV features are processed by the decoder and passed on to the appropriate head.}
        \label{fig:cobevfusion-overview}
        \vspace*{-8 pt}
    \end{figure*}

    We use BEVFusion \cite{liu2022bevfusion} and its camera-only and lidar-only variants (BEVFusion-C and BEVFusion-L) as the single-agent (no communication) models, and CoopDet3D \cite{zimmer2024tumtraf} and its camera-only and lidar-only variants (CoopDet3D-C and CoopDet3D-L) as the cooperative perception models for benchmarking the SimBEV2X dataset. CoopDet3D uses BEVFusion's backbone to extract and fuse BEV camera and lidar features for each agent, and then performs element-wise max-pooling (inspired by PillarGrid \cite{bai2022pillargrid}) to fuse multi-agent features before passing them on to the detection head. 
    
    Our implementation of CoopDet3D differs from the original in a few ways. First, unlike in \cite{zimmer2024tumtraf} where the model only fuses data from one vehicle and one RSU, our implementation accommodates an arbitrary number of vehicles and RSUs. Second, we use BEVFusion's original backbones (VoxelNet \cite{zhou2018voxelnet} and Swin Transformer \cite{liu2021swin}) and utilize its BEV segmentation head in addition to its object detection head. Finally, in \cite{zimmer2024tumtraf} vehicle data have already been transformed into the RSU's coordinate frame. This can be unrealistic, especially for V2V communication where the exact location of cooperating vehicles is not known beforehand. Therefore, in our implementation, BEV features are extracted in each agent's own coordinate frame first and then transformed into the receiving agent's coordinate frame using bilinear grid sampling.

    We also introduce CoBEVFusion, shown in \autoref{fig:cobevfusion-overview}, and its camera- and lidar-only variants CoBEVFusion-C and CoBEVFusion-L, and benchmark them on the SimBEV2X dataset. CoBEVFusion's architecture builds on CoopDet3D's by augmenting multi-agent BEV features using CoBEVT's fused axial attention (FAX) module \cite{xu2023cobevt} (inspired by MaxViT \cite{tu2022maxvit}) that performs local window-based attention followed by sparse global attention, before element-wise max-pooling takes place. We only use one FAX module (instead of the three used in \cite{xu2021opencda}), both to manage GPU memory usage and because our BEV features have a much higher resolution (128$\times$128 or 180$\times$180, depending on the variant, compared to 32$\times$32 used by CoBEVT). The attention operation uses an 8$\times$8 window with 4 heads, and the number of hidden layer nodes of the feedforward neural network is twice the number of BEV features (i.e., 160 for CoBEVFusion-C and 512 for the other two variants).

    To train these models, we used the same hyperparameters that were used to train BEVFusion, with a few exceptions. We reduced the learning rate for most model variants by 30\% to ensure convergence and made changes to the detection and segmentation heads to match them to the classes used by the SimBEV2X dataset. All models were trained for 20 epochs, except for the fused (camera+lidar) models used for 3D object detection. Those models were initialized using the weights from their lidar-only variants and trained for only 6 epochs.
    
    Similar to how BEVFusion was trained on nuScenes, we sampled the SimBEV2X dataset at 2 frames per second and used class-balanced sampling \cite{zhu2019class} to obtain 209,696 samples. Each sample contained data from one vehicle designated as ego, alongside data from cooperating agents (vehicles and RSUs) that were within the 80-meter communication radius and the $\pm$4.8-meter elevation threshold of the ego vehicle. Consistent with prior work \cite{caesar2020nuscenes, liu2022bevfusion}, we aggregated each sample's lidar point cloud with a subset of point clouds from the previous nine time steps, each decorated with the time differential, to improve detection accuracy and enhance object velocity measurement. Because the lidar feature encoder was the computational bottleneck, we randomly selected four of the previous nine point clouds for aggregation during training but used all nine during evaluation. To further manage training time, we down-sampled the original 128-channel lidar data to simulate a 32-channel lidar, similar to nuScenes \cite{caesar2020nuscenes}. Finally, while our implementation supports V2V, V2I, and V2X evaluation by filtering out unwanted agents, the results presented in the following section only cover V2X performance.

    \subsection{Results} \label{subsec:results}

    \begin{table*}[t!]
        \setlength{\tabcolsep}{3.4 pt}
        \newcommand{\rot}[1]{\multirow{2}{*}{\rotatebox[origin=c]{-90}{\textbf{#1}}}}
        \renewcommand{\arraystretch}{1.2}
        \centering
        \scriptsize
        \setlength{\abovecaptionskip}{0 pt}
        \caption{BEV Segmentation Benchmark} \label{table:seg-results}
        \begin{tabular}{@{} l c c c c c c c c c c c c c c c c c c c c c c @{}}
            \toprule
             & \rot{Road} & \rot{Hazard} & \rot{Road Line} & \rot{Sidewalk} & 
            \rot{Crosswalk} & \rot{Traffic Cone} & \rot{Barrier} & \rot{Car} & 
            \rot{Truck} & \rot{Bus} & \rot{Motorcycle} & \rot{Bicycle} & 
            \rot{Rider} & \rot{Pedestrian} & 
            \multicolumn{6}{c}{\textbf{\vspace*{12 pt}\raisebox{-8 pt}{mIoU (\%)}}} &  & \textbf{\vspace*{-12 pt}\raisebox{-20 pt}{VRAM}} \\
            \cmidrule(l){16-21}
            \textbf{Model} & & & & & & & & & & & & & & & \textbf{@0.5} & \textbf{@0.7} & \textbf{@0.9} & \textbf{$<$20 m} & \textbf{20-40 m} & \textbf{$>$40 m} & \textbf{FPS} & \textbf{(GB)} \\
            \toprule
            BEVFusion-C & 84.3 & 8.2 & 43.4 & 61.9 & 41.9 & 0.0 & 0.0 & 27.4 & 13.0 & 13.4 & 0.1 & 0.0 & 0.0 & 0.0 & 21.0 & 16.7 & 10.5 & 32.2 & 21.2 & 12.6 & \rb{37.5} & \rb{3.4} \\
            CoopDet3D-C & 85.3 & 9.5 & 39.8 & 62.6 & 42.3 & 0.0 & 0.0 & 33.7 & 7.3 & 4.6 & 0.2 & 0.0 & 0.0 & 0.0 & 20.4 & 15.9 & 9.9 & 28.6 & 20.3 & 14.5 & \gb{32.7} & \gb{3.9} \\
            CoBEVFusion-C & 85.2 & 9.7 & 39.7 & 60.1 & 41.7 & 0.0 & 0.0 & 34.9 & 9.7 & 11.4 & 0.1 & 0.0 & 0.0 & 0.0 & 21.0 & 16.5 & 10.3 & 30.5 & 21.0 & 14.1 & \bb{27.1} & \bb{3.9} \\
            \midrule
            BEVFusion-L & 84.4 & 39.0 & 51.8 & 70.8 & 48.0 & \gb{5.7} & 3.0 & 66.1 & 61.0 & 70.0 & 24.6 & 6.0 & 17.8 & 11.9 & 40.0 & 29.1 & 15.1 & \bb{52.5} & 43.5 & 27.2 & 7.0 & 13.6 \\
            CoopDet3D-L & 87.5 & \gb{46.2} & 52.6 & \bb{76.9} & 50.9 & \bb{4.2} & 3.1 & 69.8 & 63.0 & 70.0 & 26.2 & 9.1 & 16.6 & 9.3 & 41.8 & 29.0 & 13.6 & 50.0 & 44.6 & 33.0 & 6.9 & 14.2 \\
            CoBEVFusion-L & \gb{90.3} & 39.3 & \gb{55.1} & \gb{80.2} & \bb{51.9} & 2.9 & \gb{4.7} & \gb{73.3} & \gb{67.6} & \gb{74.0} & \gb{30.8} & \gb{17.6} & \gb{23.7} & \gb{19.1} & \gb{45.0} & \gb{34.2} & \gb{20.6} & 52.0 & \gb{47.9} & \gb{36.9} & 6.2 & 14.2 \\
            \midrule
            BEVFusion & 86.3 & \bb{44.4} & \bb{55.0} & 75.0 & \gb{52.3} & \rb{8.1} & 3.0 & 67.7 & 63.2 & \bb{72.2} & 27.7 & \bb{11.9} & \bb{22.1} & \bb{15.0} & \bb{43.1} & \bb{32.1} & \bb{18.0} & \gb{55.5} & \bb{46.8} & 30.5 & 6.2 & 15.4 \\
            CoopDet3D & \bb{88.9} & \rb{47.1} & 53.1 & 76.5 & 49.6 & 3.0 & \bb{4.3} & \bb{70.9} & \bb{65.5} & 71.6 & \bb{27.8} & 7.1 & 18.9 & 12.2 & 42.6 & 30.2 & 15.5 & 50.6 & 45.2 & \bb{34.1} & 5.9 & 16.5 \\
            CoBEVFusion & \rb{90.7} & 42.6 & \rb{55.8} & \rb{81.0} & \rb{54.2} & 3.4 & \rb{5.3} & \rb{74.8} & \rb{69.2} & \rb{75.0} & \rb{34.3} & \rb{23.2} & \rb{27.4} & \rb{27.4} & \rb{47.4} & \rb{36.2} & \rb{22.5} & \rb{56.3} & \rb{50.0} & \rb{38.5} & 5.5 & 16.4 \\
            \bottomrule
        \end{tabular}
        \vspace*{-8 pt}
    \end{table*}
    \begin{table*}[t!]
        \renewcommand{\arraystretch}{1.2}
        \centering
        \footnotesize
        \setlength{\abovecaptionskip}{0 pt}
        \caption{3D Object Detection Benchmark} \label{table:det-results}
        \begin{tabular}{l c c c c c c c c c c c}
            \toprule
            \multirow{2}*{\textbf{Model}} & \multicolumn{4}{c}{\textbf{mAP} (\%)} & \textbf{mATE} & \textbf{mAOE} & \textbf{mASE} & \textbf{mAVE} & \textbf{SDS} & \textbf{FPS} & \textbf{VRAM} \\
            \cline{2-5}
             & \textbf{All $\uparrow$} & \textbf{Easy $\uparrow$} & \textbf{Mod. $\uparrow$} & \textbf{Hard $\uparrow$} & \textbf{(m) $\downarrow$} & \textbf{(rad) $\downarrow$} & \textbf{$\downarrow$} & \textbf{(m/s) $\downarrow$} & \textbf{(\%) $\uparrow$} & \textbf{$\uparrow$} & \textbf{(GB) $\downarrow$} \\
            \toprule
            BEVFusion-C & 25.8 & 40.6 & 12.6 & 5.8 & 0.797 & 0.481 & 0.250 & 1.210 & 31.3 & \rb{30.1} & \rb{3.4} \\
            CoopDet3D-C & 24.2 & 35.5 & 10.9 & 9.1 & 0.801 & 0.577 & 0.261 & 1.248 & 29.1 & \gb{27.2} & \gb{3.9} \\
            CoBEVFusion-C & 11.1 & 14.0 & 3.7 & 5.4 & 0.848 & 0.650 & 0.282 & 1.352 & 20.8 & \bb{23.2} & \bb{3.9} \\
            \midrule
            BEVFusion-L & 63.9 & \bb{89.1} & 73.3 & 25.6 & \bb{0.272} & 0.353 & 0.205 & 0.373 & 66.9 & 6.7 & 13.8 \\
            CoopDet3D-L & 74.1 & 85.8 & 75.8 & \bb{48.5} & 0.288 & 0.339 & \gb{0.194} & 0.373 & 72.1 & 6.4 & 14.8 \\
            CoBEVFusion-L & \gb{76.1} & \gb{89.2} & \gb{78.6} & \rb{51.7} & \gb{0.255} & \rb{0.303} & 0.198 & \gb{0.349} & \gb{74.2} & 5.2 & 15.2 \\
            \midrule
            BEVFusion & 65.4 & 86.2 & 73.3 & 28.8 & 0.300 & 0.384 & 0.236 & 0.403 & 66.2 & 5.9 & 15.7 \\
            CoopDet3D & \bb{74.2} & 85.5 & \bb{76.0} & 47.8 & 0.289 & \bb{0.323} & \rb{0.191} & \bb{0.364} & \bb{72.5} & 5.4 & 16.2 \\
            CoBEVFusion & \rb{76.2} & \rb{90.1} & \rb{79.7} & \gb{51.3} & \rb{0.247} & \gb{0.306} & \bb{0.196} & \rb{0.333} & \rb{74.6} & 4.6 & 16.8 \\
            \bottomrule
        \end{tabular}
        \vspace*{-8 pt}
    \end{table*}
    \begin{table*}[t!]
        \renewcommand{\arraystretch}{1.2}
        \centering
        \footnotesize
        \setlength{\abovecaptionskip}{0 pt}
        \caption{3D Object Detection AP} \label{table:det-map-by-class}
        \begin{tabular}{l c c c c c c c c c c}
            \toprule
            \multirow{2}*{\textbf{Model}} & \textbf{Traffic} & \textbf{Traffic} & \textbf{Pedes-} & \multirow{2}*{\textbf{Car}} & \multirow{2}*{\textbf{Truck}} & \multirow{2}*{\textbf{Bus}} & \textbf{Motor-} & \multirow{2}*{\textbf{Bicycle}} & \textbf{Traffic} & \multirow{2}*{\textbf{Barrier}} \\
             & \textbf{Light} & \textbf{Sign} & \textbf{trian} &  &  &  & \textbf{cycle} &  & \textbf{Cone} &  \\
            \toprule
            BEVFusion-C & 25.6 & 15.4 & 24.4 & 49.3 & 43.3 & 42.6 & 26.7 & 11.5 & 13.5 & 5.6 \\
            CoopDet3D-C & 20.7 & 9.7 & 26.6 & 49.4 & 35.4 & 34.7 & 30.3 & 15.7 & 12.9 & 6.7 \\
            CoBEVFusion-C & 10.1 & 1.6 & 10.7 & 44.9 & 11.0 & 16.8 & 6.3 & 1.7 & 6.4 & 1.6 \\
            \midrule
            BEVFusion-L & 51.5 & 67.2 & 72.1 & 75.9 & 77.4 & 83.3 & 68.3 & 62.0 & 44.4 & 36.6 \\
            CoopDet3D-L & 67.6 & 75.8 & \bb{88.6} & \bb{90.1} & \bb{85.0} & \bb{86.1} & \bb{80.7} & \gb{78.3} & 48.3 & 40.9 \\
            CoBEVFusion-L & \gb{72.2} & \rb{82.2} & \rb{89.4} & \rb{91.4} & \gb{86.4} & \gb{87.1} & \rb{82.9} & \rb{79.4} & 48.7 & \gb{41.0} \\
            \midrule
            BEVFusion & 48.0 & 67.4 & 72.1 & 83.2 & 78.2 & 84.2 & 67.9 & 64.6 & \bb{48.7} & 40.2 \\
            CoopDet3D & \bb{68.4} & \bb{76.2} & 87.9 & 89.9 & 84.7 & 86.0 & 79.5 & 76.0 & \rb{50.8} & \rb{42.4} \\
            CoBEVFusion & \rb{72.8} & \gb{81.9} & \gb{89.1} & \gb{91.4} & \rb{86.8} & \rb{88.0} & \gb{82.3} & \bb{78.2} & \gb{50.6} & \bb{41.0} \\
            \bottomrule
        \end{tabular}
        \vspace*{-4 pt}
    \end{table*}

    \autoref{table:seg-results} presents the BEV segmentation benchmark results across all variants of BEVFusion, CoopDet3D, and CoBEVFusion. Overall, CoBEVFusion performs the best, reaching a mIoU of 47.4\% at a 0.5 IoU threshold, followed by CoBEVFusion-L at 45.0\% and BEVFusion at 43.1\%. The fused and lidar-only variants substantially outperform their camera-only counterparts, underscoring the importance of accurate depth information for BEV segmentation. Camera-only models score 0.0 IoU across all small object classes (\textit{traffic cone}, \textit{barrier}, \textit{bicycle}, \textit{rider}, \textit{pedestrian}) because the image-to-BEV view transformer (Lift, Splat, Shoot \cite{philion2020lift}) cannot reliably recover the metric position of small objects. These models perform reasonably well on static map classes, including \textit{road} ($\smallsim$85\%), \textit{road line} ($\smallsim$40\%), \textit{sidewalk} ($\smallsim$60\%), and \textit{crosswalk} ($\smallsim$42\%), where coarse spatial structure and the ground plane provide sufficient cues.
    
    Comparing the lidar-only and fused variants reveals a nuanced interaction between fusion strategy and camera features, where CoopDet3D gains much less from the added camera features than BEVFusion or CoBEVFusion, even regressing for the \textit{bicycle} class by 2.0\%. CoopDet3D also performs worse than both BEVFusion and CoBEVFusion, with the gap growing as the confidence threshold is increased. This pattern suggests that naive max-pooling of the fused BEV features is sensitive to the spatial misalignment inherent in image-based BEV projections: erroneous features introduced by imprecise depth estimates can suppress the correct lidar-derived activations, which can be compounded by the bilinear grid transformation of the communicated features, degrading performance. CoBEVFusion's FAX module mitigates this by learning to selectively weight multi-agent contributions, allowing the model to exploit complementary cues where they are reliable while suppressing noise where they are not. This is particularly effective for smaller objects (e.g. \textit{bicycle}, \textit{motorcycle}, \textit{pedestrian}) where camera features become relatively more informative, resulting in CoBEVFusion significantly outperforming CoopDet3D. These gains come at a modest computational cost: CoBEVFusion is only 11.3\% slower than BEVFusion and 6.8\% slower than CoopDet3D, and uses roughly the same amount of GPU memory.

    We observe a similar pattern in the 3D object detection benchmarks shown in \autoref{table:det-results} and \autoref{table:det-map-by-class}, though the performance gap between CoBEVFusion and CoopDet3D is not as large. This comes down to the head used for each task. The BEV segmentation head reads every single BEV cell and assigns it a class. Any local feature error, e.g., a noisy or misaligned cell produced by CoopDet3D's max-pooling, propagates directly into a wrong cell prediction. On the other hand, CoBEVFusion's features are spatially smoother and more contextually consistent, which is why it performs noticeably better. For 3D object detection, the Transfusion \cite{bai2022transfusion} head that these models use generates a small set of learned object queries, initialized from heatmap peaks, and uses cross-attention over the entire BEV feature map to aggregate evidence for each query. That pooling operation is inherently robust to local feature noise, i.e., a single misaligned cell is diluted across the whole receptive field of a query. As a result, the detection head can recover accurate bounding boxes even from CoopDet3D's rougher features, leaving little room for the FAX module to add value. The evaluation metrics further reinforce this. Bounding box mAP (using center-distance matching) tolerates small spatial offsets, but segmentation mIoU has no such tolerance.

    Overall, \autoref{table:det-results} indicates that both CoBEVFusion and CoopDet3D significantly outperform BEVFusion. This improvement is particularly pronounced for \textit{hard} objects (which are either distant or have few lidar points inside their bounding box), where CoBEVFusion leverages cooperative features to nearly double the baseline performance. \autoref{table:det-map-by-class} shows that CoBEVFusion achieves its most substantial gains over BEVFusion in the \textit{traffic light}, \textit{traffic sign}, \textit{pedestrian}, \textit{motorcycle}, and \textit{bicycle} classes.

    \section{Conclusion and Future Work} \label{sec:conclusion}

    In this paper, we addressed the lack of comprehensive, large-scale, multi-agent perception datasets, a fundamental challenge facing vehicle-to-everything (V2X) cooperative perception. To that end, we introduced SimBEV2X, a modular, configurable, and scalable synthetic data generation tool built on the CARLA simulator, which creates highly randomized driving scenarios to collect rich multi-agent multi-modal sensor data alongside a multitude of annotations. We also presented the SimBEV2X dataset, the largest V2X perception dataset to date, comprised of 258 scenes and an order of magnitude larger than existing V2X datasets. Finally, we established a strong baseline on the SimBEV2X dataset using CoopDet3D and proposed CoBEVFusion, a novel architecture that combines CoopDet3D with fused axial attention (FAX) for context-aware multi-agent feature aggregation, resulting in superior performance.

    Future work will focus on multi-agent domain transfer, where we investigate and develop methods that transform the multi-agent data from the SimBEV2X dataset into real-world data. We will also focus on developing novel cooperative perception algorithms, in particular algorithms that are sensor- and model-agnostic, and will benchmark them on the SimBEV2X dataset.

    \section*{Acknowledgment}

    This work was supported by the computational resources of Virginia Tech's Advanced Research Computing (ARC), the Commonwealth Cyber Initiative Central Virginia Node (CCI-CVN) AI Testbed, and Virginia Commonwealth University's High Performance Research Computing (HPRC).
    
    \bibliography{simbev2x}
    \bibliographystyle{IEEEtran}

    \vspace*{-16 pt}
    \begin{IEEEbiography}[{\includegraphics[width = 1 in, height = 1.25 in, clip, keepaspectratio]{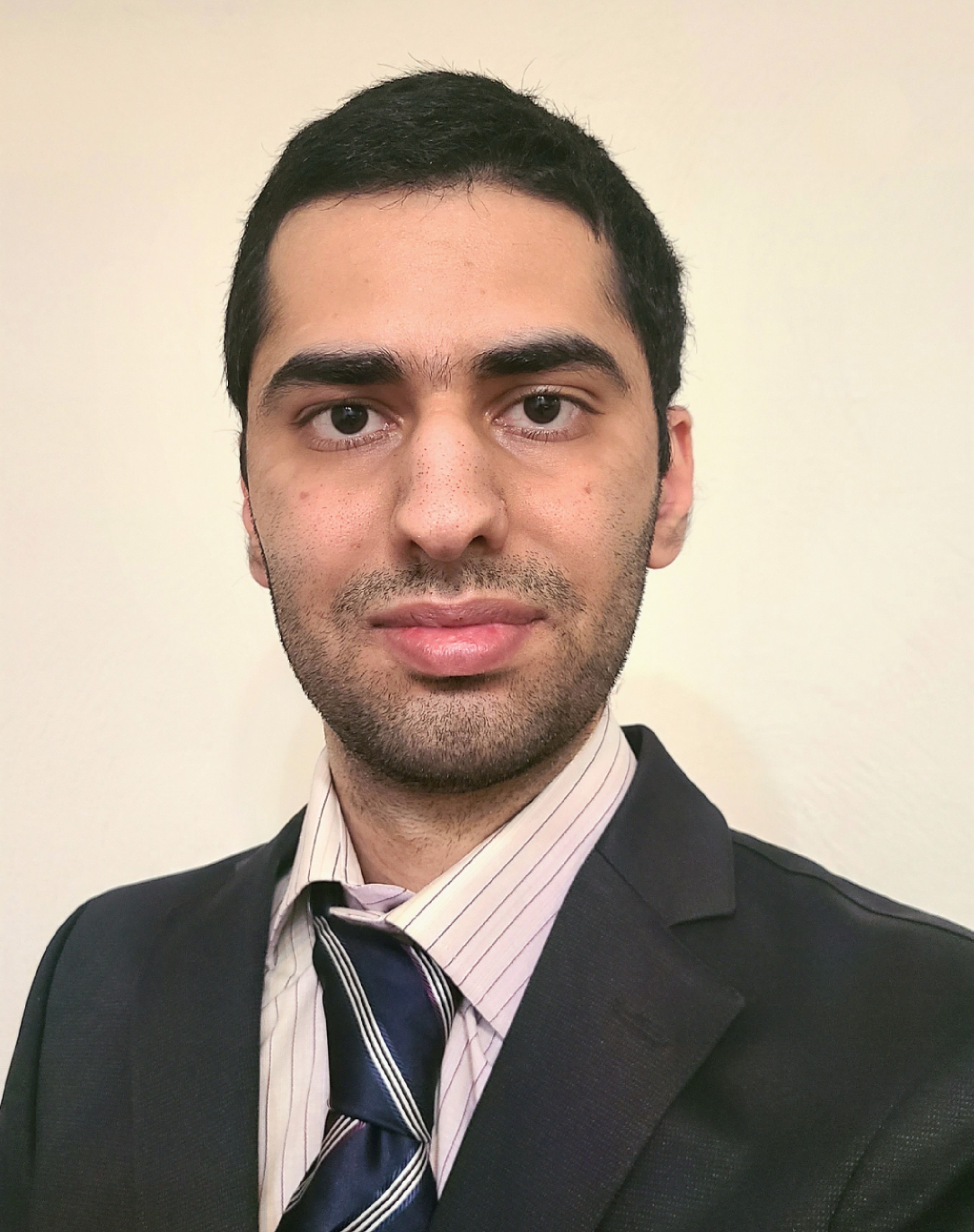}}]{Goodarz Mehr}
		received the B.Sc. degree in mechanical engineering from Sharif University of Technology, Tehran, Iran, in 2016 and the M.Sc. and Ph.D. degrees in mechanical engineering from Virginia Tech, Blacksburg, VA, USA, in 2023 and 2024, respectively. He is currently a postdoctoral research associate at Autonomous Robots and Vehicles Lab (ARVL) at Virginia Commonwealth University (VCU), Richmond, VA, USA. His research interests include multi-agent robotics, stochastic planning models, and cooperative perception.
	\end{IEEEbiography}
    \vspace*{-16 pt}
    \begin{IEEEbiography}[{\includegraphics[width = 1 in, height = 1.25 in, clip, keepaspectratio]{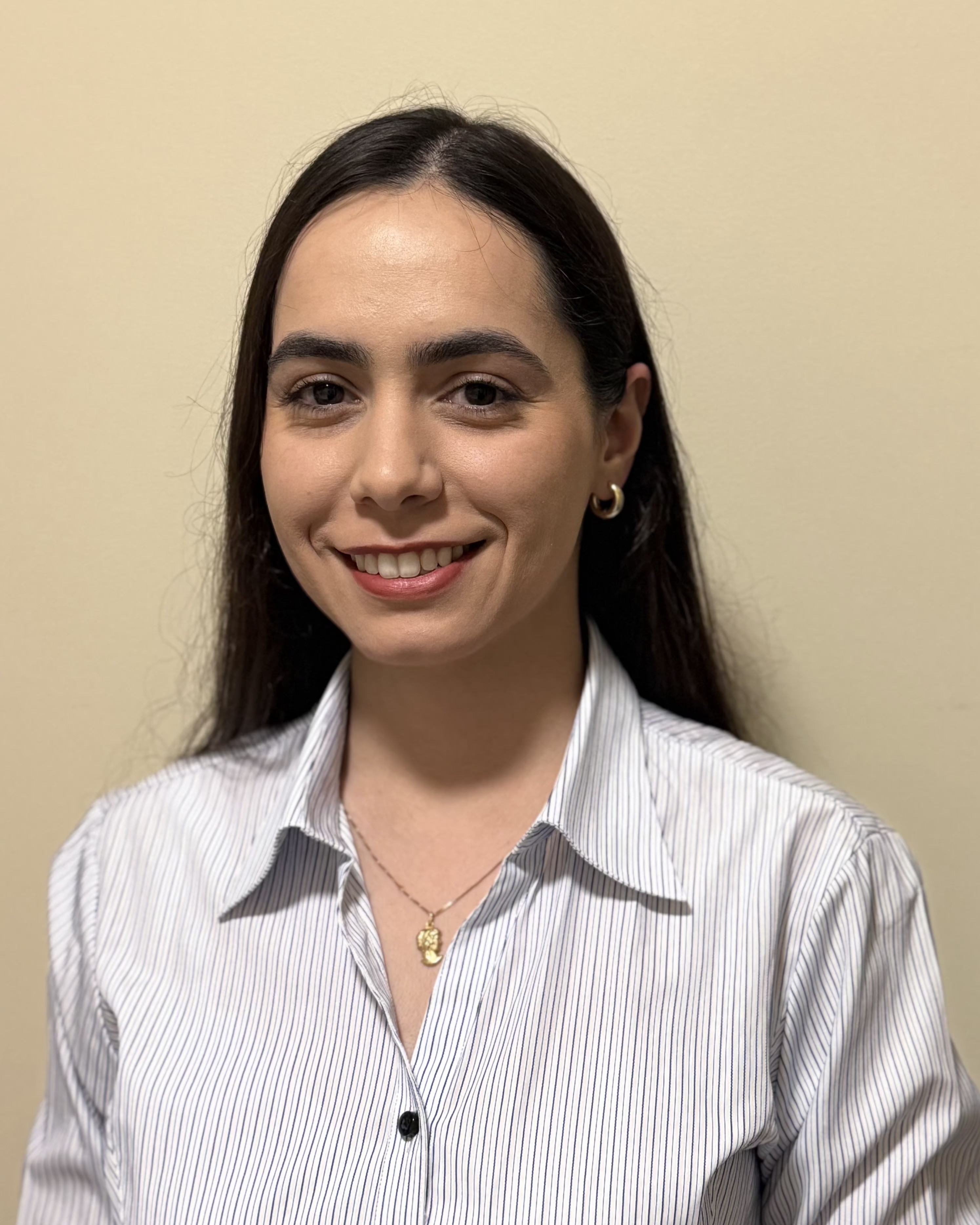}}]{Sepideh Gohari}
		received the B.S. degree in electrical engineering and the M.S. degree in artificial intelligence and robotics from Ferdowsi University of Mashhad, Mashhad, Iran. She is currently pursuing the Ph.D. degree in electrical and computer engineering at the Autonomous Robots and Vehicles Lab (ARVL), at Virginia Commonwealth University (VCU), Richmond, VA, USA. Her research interests include computer vision, autonomous driving, robotics, and artificial intelligence.
	\end{IEEEbiography}
    \vspace*{-16 pt}
    \begin{IEEEbiography}[{\includegraphics[width = 1 in, height = 1.25 in, clip, keepaspectratio]{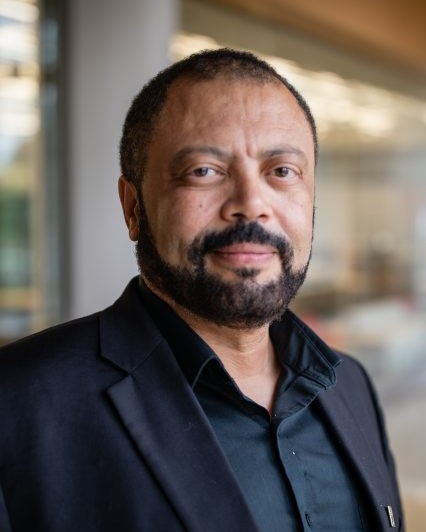}}]{Montasir Abbas}
		is a Professor of Civil Engineering at Virginia Tech, Blacksburg, VA, USA. He holds a B.Sc. degree in Civil Engineering from the University of Khartoum, Khartoum, Sudan, an M.Sc. in Civil Engineering from the University of Nebraska-Lincoln, Lincoln, NE, USA, and a Ph.D. in Civil Engineering from Purdue University, West Lafayette, IN, USA. His research interests include simulation, traffic operations and control, traffic flow theory, and the application of artificial intelligence in transportation engineering.
	\end{IEEEbiography}
    \vspace*{-16 pt}
    \begin{IEEEbiography}[{\includegraphics[width = 1 in, height = 1.25 in, clip, keepaspectratio]{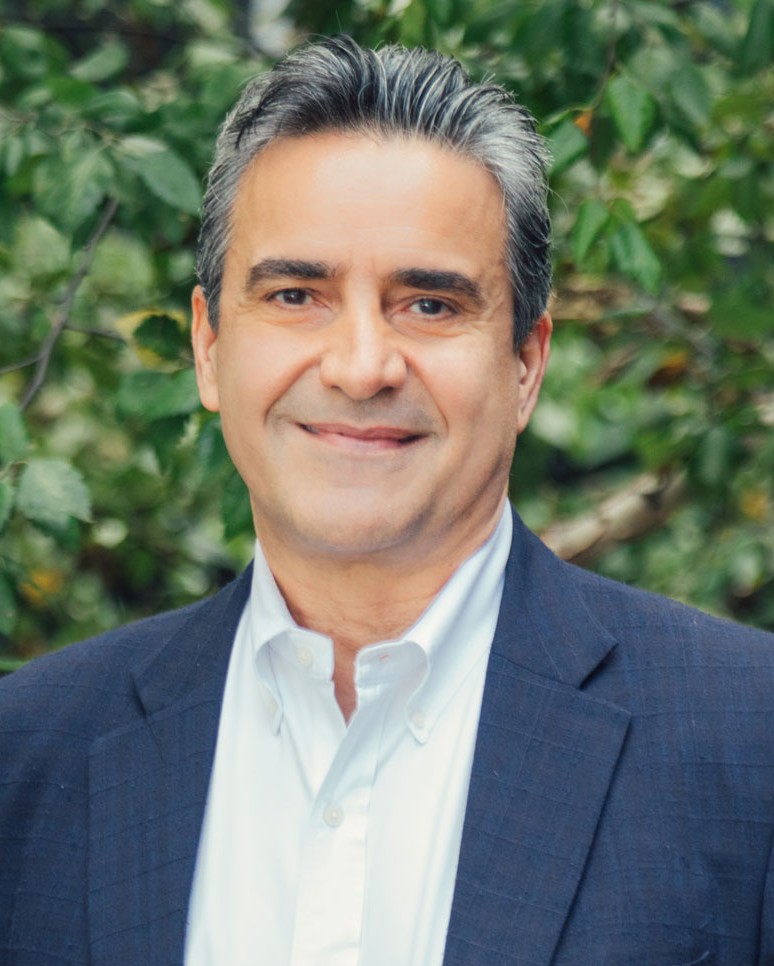}}]{Azim Eskandarian}
		(Fellow, IEEE) received the B.S. and D.Sc. degrees in mechanical engineering from George Washington University (GWU), Washington, D.C., USA and the M.S. degree in mechanical engineering from Virginia Tech, Blacksburg, VA, USA.
        
        He has been the Dean of Virginia Commonwealth University (VCU), College of Engineering, Richmond, USA since 2023. Before that, he was a Professor and the Head of the Department of Mechanical Engineering at Virginia Tech since 2015, where he became the Nicholas and Rebecca Des Champs Chair Professor in 2018, and where he established the Autonomous Systems and the Intelligent Machines Laboratory, to conduct research in intelligent and autonomous vehicles and mobile robotics. He was the Co-Founder of the National Crash Analysis Center in 1992. He was an Engineer/Project Manager in industry from 1983 to 1989. He was an Assistant Professor with The Pennsylvania State University, York, PA, USA, from 1989 to 1992. He was a Professor of engineering and applied science with GWU and the Founding Director of the Center for Intelligent Systems Research from 1996 to 2015. He was the Director of the National Crash Analysis Center from 1998 to 2002 and from 2013 to 2015. He was the Director of the Transportation Safety and Security University Area of Excellence from 2002 to 2015.
        
        Dr. Eskandarian is a fellow of ASME and a member of SAE professional societies. He received the IEEE Intelligent Transportation Society’s Outstanding Researcher Award in 2017 and the GWU School of Engineering Outstanding Researcher Award in 2013. He was the Editor-in-Chief of the IEEE Transactions on Intelligent Transportation Systems from 2019 to 2023.
	\end{IEEEbiography}

\end{document}